\pdfoutput=1  
\PassOptionsToPackage{table,dvipsnames}{xcolor}
\RequirePackage{float}

\documentclass{bmvc2k}

\usepackage{amsmath}
\usepackage{amssymb}
\usepackage{amsfonts}
\usepackage{booktabs}
\usepackage{comment}
\usepackage{cleveref}  
\hypersetup{pdftitle={Uncertainty-Aware RL-Controlled Adaptive 3D Mapping},
            pdfauthor={Alpay Ozkan, Tunc Ozan Aydin, Marc Pollefeys, Jelena Trisovic, Daniel Barath}}
\crefname{appendix}{Appendix}{Appendices}
\Crefname{appendix}{Appendix}{Appendices}
\crefname{subappendix}{Appendix}{Appendices}
\Crefname{subappendix}{Appendix}{Appendices}

\title{Uncertainty-Aware RL-Controlled Adaptive 3D Mapping}

\addauthor{Alpay Ozkan}{aoezkan@inf.ethz.ch}{1}
\addauthor{Tunc Ozan Aydin}{tunc.aydin@inf.ethz.ch}{1}
\addauthor{Marc Pollefeys}{marc.pollefeys@inf.ethz.ch}{1,2}
\addauthor{Jelena Trisovic}{jelena.trisovic@inf.ethz.ch}{$^{\star}$ 1,3}
\addauthor{Daniel Barath}{danielbela.barath@inf.ethz.ch}{$^{\star}$ 1}

\addinstitution{%
 ETH Zurich\\
 Zurich, Switzerland
}
\addinstitution{%
 Microsoft\\
 Zurich, Switzerland
}
\addinstitution{%
 ETH AI Center\\
 Zurich, Switzerland
}

\runninghead{Ozkan et al.}{UnRL: Adaptive 3D Mapping}

\def\eg{\emph{e.g}\bmvaOneDot}

\def\ie{\emph{i.e}\bmvaOneDot}

\newcommand{\ablLarge}{\textsc{Voxblox (8 cm)}}
\newcommand{\ablMiddle}{\textsc{Voxblox (4 cm)}}
\newcommand{\ablSmall}{\textsc{Voxblox (1 cm)}}
\newcommand{\ablMapadaptSplitAvg}{\textsc{Map-Adapt-Split-Avg}}
\newcommand{\ablMapadaptGeo}{\textsc{Map-Adapt}}
\newcommand{\ablUncertaintyGeoTexture}{\textsc{Un-Map-Adapt (Ours)}\xspace}
\newcommand{\ablUncertaintyGeoTextureRL}{\textsc{UnRL (Ours)}\xspace}
\newcommand{\equalcontrib}[1]{%
  \begin{NoHyper}%
  \renewcommand\thefootnote{}%
  \footnote{\null\hspace{-1.9em}$^{\star}$#1}%
  \addtocounter{footnote}{-1}%
  \end{NoHyper}}

\begin{document}

\maketitle
\equalcontrib{Equal supervision}

\begin{abstract}
    Voxel-based volumetric mapping is fundamental to 3D reconstruction, yet fixed-resolution grids remain inherently inefficient -- wasting memory in uniform regions and losing detail in complex ones. Existing adaptive methods, such as MAP-ADAPT, partially address this by varying resolution based on geometry and user-defined semantic class lists, but these heuristics require expert tuning, lack generalization to unseen objects, and provide no explicit mechanism to control memory usage. We propose an adaptive framework that refines voxels based on semantic entropy, which captures label uncertainty, together with geometric curvature and texture richness as scene complexity cues, yielding principled resolution allocation without reliance on semantic taxonomies. To make the accuracy-memory trade-off explicit and user-controlled, we further introduce a reinforcement learning agent that learns voxel subdivision policies under a user-specified target memory budget, replacing hand-tuned thresholds with a single intuitive control parameter. The resulting multi-resolution TSDF achieves higher geometric accuracy, better semantic consistency, and improved memory-accuracy trade-offs compared to MAP-ADAPT and fixed-resolution baselines on both synthetic and real-world datasets.
    Our code and models are available at \url{https://github.com/alpayozkan/UnRL}.
\end{abstract}

\section{Introduction}
\label{sec:intro}

Autonomous systems rely on 3D maps that capture geometry, color, and semantics to perceive, navigate, and interact with their environment.
Volumetric representations based on Truncated Signed Distance Fields (TSDFs)~\cite{newcombe2011kinectfusion,niessner2013voxel,voxblox} have become the standard for dense reconstruction, offering accurate geometry and consistent integration over time.
However, conventional TSDF pipelines employ a \emph{fixed voxel resolution}, which leads to redundant storage in large homogeneous regions (\eg, walls or floors) and loss of detail in complex areas.

\begin{figure}[tb]
    \centering
    \includegraphics[width=0.85\linewidth]{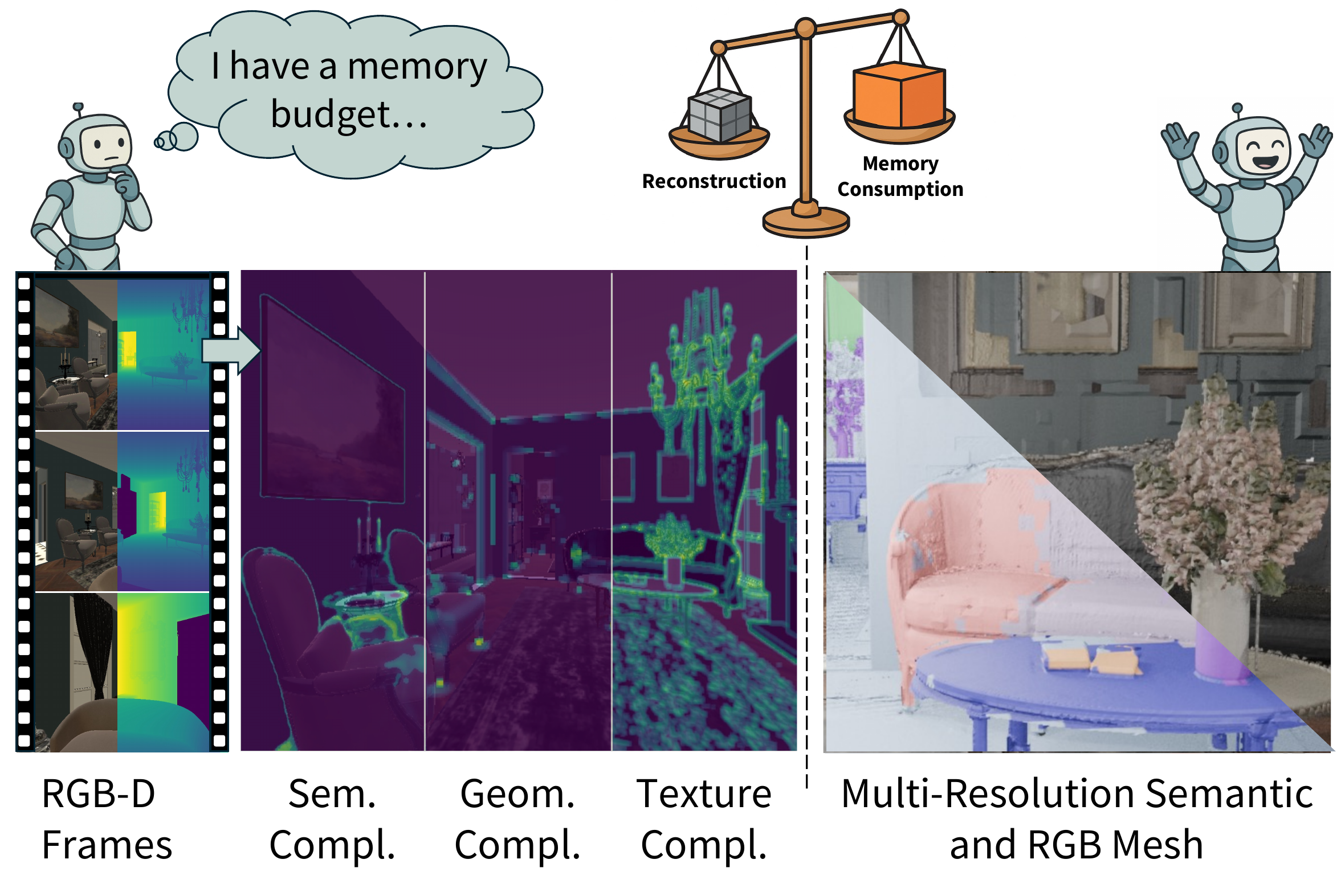}
    \caption{Given RGB-D input frames, our method computes semantic entropy, 
    geometric complexity, and texture complexity as voxel-level subdivision signals.
    Semantic entropy highlights object boundaries where multiple labels compete; 
    geometric complexity responds to corners and curved surfaces (\eg, sofa armrests); 
    and texture complexity activates on visually detailed regions such as carpets.
    These complexity cues drive selective voxel refinement, while a reinforcement 
    learning agent uses a user-specified target memory budget to balance reconstruction 
    fidelity and storage cost.
    The resulting multi-resolution TSDF mesh preserves high-quality geometry, 
    semantics, and appearance where it matters most.}
    \label{fig:teaser}
\end{figure}

Recent work has therefore shifted toward \emph{adaptive mapping}, where voxel resolution is adjusted according to scene complexity.
MAP-ADAPT~\cite{zheng2024mapadapt} represents an important step in this direction, combining geometric cues and manually defined semantic classes to control voxel subdivision.
By allocating higher resolution to semantically meaningful regions, it achieves compact multi-resolution 3D voxel maps while maintaining fidelity.
However, its refinement policy remains tied to user-specified semantic classes and fixed geometric rules, requiring dataset- or task-specific design choices and providing no class-agnostic measure of semantic ambiguity or texture complexity.
Moreover, these deterministic rules provide no principled mechanism to manage the trade-off between reconstruction quality and limited storage capacity.

We address these limitations by reformulating adaptive mapping as a \emph{data-driven refinement problem} guided by semantic uncertainty and geometric/texture complexity.
Each voxel maintains a semantic class distribution whose entropy quantifies label uncertainty, complemented by geometric curvature and texture richness scores that capture local scene complexity.
Subdivision is triggered when semantic uncertainty is high or when scene complexity indicates that coarse resolution cannot faithfully represent the local geometry or appearance, ensuring that memory is devoted to informative regions rather than predefined object categories.
This data-driven refinement produces adaptive maps that respond dynamically to scene complexity, semantic ambiguity, and visual richness.

While these signals provide a principled refinement criterion, the deterministic variant relies on threshold selection and provides no direct mechanism to target a desired memory operating point.
To overcome this, we introduce a \emph{reinforcement learning agent} that learns to control subdivision under a user-defined \emph{target storage budget}.
The agent observes local and global voxel statistics and learns subdivision policies that maximize reconstruction accuracy while penalizing budget overuse (see \Cref{fig:teaser}).
The target budget corresponds directly to a physical memory amount (\eg, in megabytes), providing an intuitive interface for resource-aware mapping.

\noindent
Our main contributions are as follows:
\begin{itemize}
    \item \textbf{Unified adaptive mapping framework:} We integrate geometric, semantic, and texture complexity signals into a multi-resolution TSDF representation that generalizes beyond class-based heuristics and scales with available memory.
    \item \textbf{Uncertainty- and complexity-driven adaptivity:} We trigger voxel refinement using semantic entropy as a measure of label uncertainty, complemented by geometric curvature and texture richness as scene complexity signals, providing a unified and data-driven subdivision criterion.
    \item \textbf{Reinforcement-learned subdivision policy:} We train an RL agent using Proximal Policy Optimization (PPO)~\cite{schulman2017proximal} to decide when and where to subdivide, learning budget-aware policies that trade map fidelity against storage cost.
\end{itemize}

\section{Related Work}
\label{sec:relatedwork}

\noindent \textbf{Volumetric Mapping.}
Volumetric 3D reconstruction based on Truncated Signed Distance Fields (TSDFs) is widely used for dense scene mapping~\cite{newcombe2011kinectfusion, niessner2013voxel, voxblox}.
These methods integrate RGB-D observations into voxel grids, providing continuous surface representations and free-space reasoning.
To mitigate the memory footprint of dense grids, hierarchical data structures such as voxel hashing~\cite{niessner2013voxel, kahler2016hierarchical} and octrees~\cite{hornung2013octomap, vespa2019adaptive} have been proposed to allocate voxels only near surfaces.
While these systems achieve real-time performance and scalable mapping, they typically rely on a \emph{fixed voxel resolution}, which limits their ability to represent both fine local details and large-scale environments efficiently.
\emph{Neural implicit methods} such as NeRF~\cite{mildenhall2021nerf} and 3D Gaussian Splatting~\cite{kerbl2023gaussians} target image rendering rather than explicit volumetric mapping, and do not produce TSDF grids, semantic labels, or free-space representations required for planning and navigation.

\noindent \textbf{Semantic Mapping.}
Integrating semantic understanding into 3D mapping has been central to advancing scene perception.
Early works like SemanticFusion~\cite{mccormac2017semanticfusion} fused per-frame semantic predictions into surfel-based reconstructions, while Kimera~\cite{rosinol2020kimera} combined real-time metric reconstruction with semantic fusion into a unified volumetric mesh.
Instance- and panoptic-level methods~\cite{mccormac2018fusion++, grinvald2019volumetric, narita2019panopticfusion, wu2021scenegraphfusion} further extended these ideas by maintaining per-object volumes or combining instance labels.
However, these pipelines generally operate at uniform spatial resolution, distributing computational and memory resources equally across all regions, independent of their semantic or geometric complexity.

\noindent \textbf{Adaptive Mapping.}
To improve efficiency, adaptive techniques dynamically adjust voxel resolution based on scene content or task relevance.
\emph{Geometry-based adaptivity} refines regions of high curvature, surface density, or structural complexity~\cite{steinbrucker2014voxel, stuckler2014multiresolution, vespa2019adaptive, zienkiewicz2016real}.
These approaches improve geometric fidelity near surfaces but lack semantic or visual awareness.
\emph{Semantic-based adaptivity} extends this idea by incorporating object information to guide subdivision.
Panoptic Multi-TSDFs~\cite{schmid2022panoptic} reconstruct separate TSDF volumes per object instance, while MAP-ADAPT~\cite{zheng2024mapadapt} combines geometric and semantic cues into a single multi-resolution TSDF map.
MAP-ADAPT demonstrates real-time adaptive reconstruction by refining user-specified semantic classes, but its reliance on manually defined rules limits generalization to unseen categories and neglects uncertainty in semantic or textural predictions that could indicate ambiguous or informative regions.

\noindent \textbf{Uncertainty and Learning-Based Adaptivity.}
Several works have incorporated uncertainty into mapping, active exploration, and reconstruction.
SemanticFusion~\cite{mccormac2017semanticfusion} and Kimera~\cite{rosinol2020kimera} maintain probabilistic class distributions per voxel, though uncertainty is not used to influence map structure.
Information-theoretic formulations~\cite{charrow2015information, Asgharivaskasi2021SemanticOM} employ entropy and mutual information to guide exploration; notably, semantic entropy within a Bayesian Octree has been used to drive active exploration via mutual information maximization~\cite{Asgharivaskasi2021SemanticOM}.
Reinforcement learning has similarly been applied to active exploration and mapping~\cite{zhu2017target,chen2019self,placed2020deep}, demonstrating that learned policies can balance performance and efficiency.
However, these exploration-focused methods determine \emph{where to move next} rather than \emph{how to allocate reconstruction resolution} along a fixed trajectory.
In contrast, we formulate voxel subdivision as an \emph{uncertainty-driven} refinement problem and leverage reinforcement learning to learn adaptive, budget-aware subdivision policies that directly regulate the trade-off between reconstruction fidelity and storage capacity.

\begin{figure}[tb]
    \centering
    \includegraphics[width=0.98\linewidth]{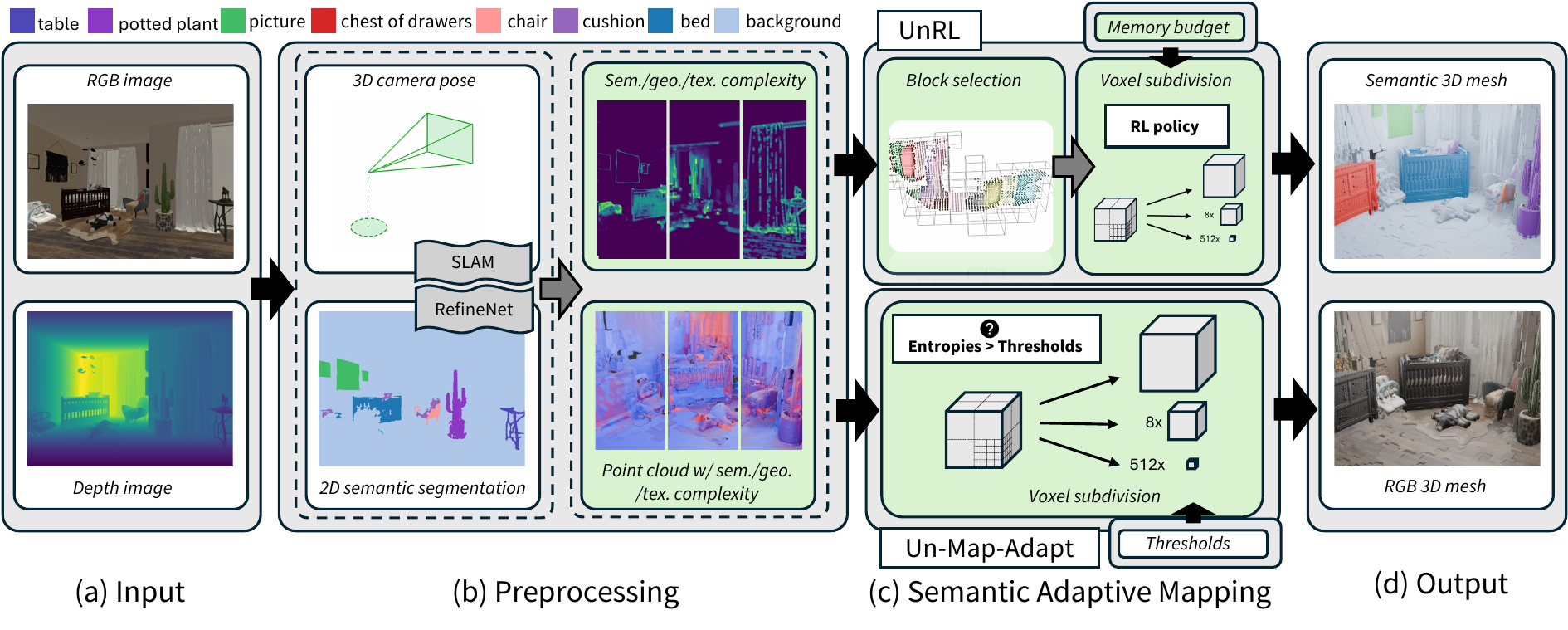}
    \caption{
        \textbf{Overview.}
        (a) RGB-D frames are fed into (b) a preprocessing stage that estimates camera poses via SLAM (similar to~\cite{zheng2024mapadapt}), performs 2D semantic segmentation~\cite{lin2017refinenet}, and computes per-voxel complexity cues (semantic, geometric, and texture) that highlight object boundaries, geometric curvature, and textured regions.
        (c) Our adaptive mapping refines map resolution either through deterministic complexity thresholds (\emph{Un-Map-Adapt}) or through a reinforcement-learning policy (\emph{UnRL}) allocating resolution under a user-specified target memory budget.
        The RL agent predicts subdivision thresholds for budget-aware refinement.
        (d) The final multi-resolution TSDF map is converted into a high-fidelity mesh.}
    \label{fig:method_pipeline}
\end{figure}

\section{Proposed Method}
\label{sec:method}

We propose an adaptive 3D semantic mapping framework that refines voxel resolution by explicitly targeting high-uncertainty regions.
We build upon MAP-ADAPT~\cite{zheng2024mapadapt}, which operates on an adaptively subdivided voxel grid and controls subdivision via manually defined semantic classes and geometric complexity.
In contrast, we (i) reformulate voxel subdivision as an uncertainty-driven refinement problem, using semantic, geometric, and textural complexity as proxies to trigger adaptive resolution allocation,
and (ii) learn a budget-aware, soft-constrained subdivision policy with reinforcement learning (RL) that replaces hand-tuned thresholds with a single intuitive parameter: the target storage budget.

\noindent
\textbf{Preliminaries.}
We consider a sequence of RGB-D frames $\{(I_k, D_k)\}_{k=1}^K$ with camera poses $\{C_k\}$ and per-frame semantic predictions $\{P_k\}$, where $I_k$ and $D_k$ are the input image and depth in the $k$-th frame.
The goal is to reconstruct an adaptive-resolution TSDF map $\mathcal{M}$ that allocates fine resolution only where it meaningfully reduces uncertainty.
We adopt a hierarchical voxel hashing structure in the spirit of MAP-ADAPT~\cite{zheng2024mapadapt}, with three resolution levels (\textit{coarse}: $8$\,cm, \textit{medium}: $4$\,cm, \textit{fine}: $1$\,cm voxels).
Each voxel $v_i$ stores 
$v_i = \{ d_i, S_i, g_i, t_i \}$,
where $d_i \in \mathbb{R}$ is the TSDF value, $S_i \in \mathbb{R}^C$ denotes 
the semantic log-scores over $C$ classes, $g_i \in \mathbb{R}$ a 
geometric complexity scalar, and $t_i \in \mathbb{R}$ a texture complexity 
scalar. Semantic entropy $H_i^{\text{sem}}$ is derived on-the-fly from the normalized semantic distribution $\hat{P}_{i}^{\text{sem}}$ obtained from $S_i$,
while the combined uncertainty proxy $h_i$ used by the RL agent is computed from 
all three modalities as defined in \Cref{subsec:rl}.
The pipeline is shown in \Cref{fig:method_pipeline}.

\subsection{Uncertainty-Driven Adaptive Mapping}
\label{subsec:uncertainty}

We propose our uncertainty-aware adaptive mapping pipeline Un-Map-Adapt.

\noindent\textbf{Voxel uncertainty and complexity measures.}
Each voxel $v_i$ maintains a semantic class distribution; geometric and texture
complexity are fused as scalar attributes computed in the observation domain.
Semantic entropy captures label uncertainty across observations, motivating 
refinement at class boundaries; geometric curvature identifies edges and corners 
where coarse voxels lose structural detail; and texture richness flags visually 
dense regions where low resolution causes blurring.

\noindent
\emph{Semantic entropy.}
In each voxel, we fuse per-frame class probabilities into semantic log-scores $S_i$ using a Bayesian update similar to SemanticFusion~\cite{mccormac2017semanticfusion} and Kimera~\cite{rosinol2020kimera}. After fusion, the normalized class probabilities are as follows:
\begin{equation}
    \hat{P}_{i}^{\text{sem}}(c) =
    \frac{\exp(S_i(c))}{\sum_{c'=1}^{C} \exp(S_i(c'))},
    \quad c = 1,\dots,C.
\end{equation}
The semantic entropy is then as follows:
\begin{equation}
    H_i^{\text{sem}}
    = - \sum_{c=1}^{C} \hat{P}_{i}^\text{sem}(c) \log \hat{P}_{i}^\text{sem}(c).
\end{equation}

We use entropy as a class-agnostic measure of predictive ambiguity: it increases when probability mass is distributed across competing labels and does not depend on which semantic class is most likely, thereby highlighting regions that may benefit from finer resolution.

\noindent
\emph{Geometric complexity.}
We estimate geometric complexity using the change-of-curvature (CoC) 
descriptor~\cite{pauly2002efficient}, similar to~\cite{zheng2024mapadapt}.
For each depth frame, pixels are back-projected to 3D at stride~2 for efficiency,
producing a subsampled point cloud.
For each 3D point $p$, we find all neighbors within radius $r{=}0.1$\,m using 
a KD-tree and compute the local covariance matrix as
\begin{equation}
    \mathbf{C}(p) = \frac{1}{n}\sum_{i=1}^{n}(p_i - \bar{p})(p_i - \bar{p})^\top,
\end{equation}
where $\bar{p}$ is the local centroid and $n$ is the number of neighbors.
Let $\lambda_1 \ge \lambda_2 \ge \lambda_3 \ge 0$ be the eigenvalues of $\mathbf{C}(p)$;
the CoC is defined as:
    $\text{CoC}(p) = \lambda_3 / (\lambda_1 + \lambda_2 + \lambda_3) \in [0, 0.33]$,
where $\text{CoC}{\approx}0$ indicates a locally planar surface and 
$\text{CoC}{\approx}0.33$ indicates an isotropic, maximally complex neighborhood.
The per-point CoC values are stored at their original pixel coordinates $(u,v)$, 
producing a per-frame 2D complexity map.
These per-observation values are then fused into each voxel $v_i$ as a 
\emph{single scalar} $g_i$ via confidence-weighted accumulation 
(weight $w_j = 1/z_j^2$, where $z_j$ is the observation depth, downweighting distant measurements)
 as:
\begin{equation}
    g_i \leftarrow \frac{w_i\, g_i + w_j\, \text{CoC}(p_j)}{w_i + w_j}.
\end{equation}
High $g_i$ indicates geometrically complex or curved regions, such as edges and 
corners, that benefit from finer voxel resolution.

\noindent
\emph{Texture complexity.}
We compute a per-pixel texture complexity map for each RGB frame using 
local gradient entropy.
We apply $3{\times}3$ Sobel filters to obtain horizontal and vertical 
gradients $G_x$, $G_y$, and compute per-pixel gradient magnitudes $m(u,v) = \sqrt{G_x(u,v)^2 + G_y(u,v)^2}$.
For each pixel $(u,v)$, we extract a $5{\times}5$ neighborhood (with reflect 
padding at borders), yielding 25 gradient magnitude samples.
These are binned into a 20-bin histogram spanning the local 
$[m_{\min}, m_{\max}]$ range of the window, normalized to a probability 
distribution $\{p_j\}_{j=1}^{20}$, and Shannon entropy is computed as:
\begin{equation}
    H^{\text{tex}}(u,v) = -\sum_{j=1}^{20} p_j \log(p_j + \varepsilon),
    \quad \varepsilon = 10^{-6}.
\end{equation}
Intuitively, flat textureless regions concentrate gradients in a single bin 
(low entropy), while richly textured regions spread mass across bins 
(high entropy).
The resulting entropy map is z-normalized per image over valid pixels as:
\begin{equation}
    t(u,v) = \frac{H^{\text{tex}}(u,v) - \mu_H}{\sigma_H},
\end{equation}
where $\mu_H$ and $\sigma_H$ are the mean and standard deviation computed 
over all valid pixels in the frame, making complexity values comparable 
across images with different overall texture levels.
Positive per-pixel values are then projected into 3D and fused into each voxel $v_i$ 
as a \emph{single scalar} $t_i$ via confidence-weighted accumulation 
(weight $w_j = 1/z_j^2$) as 
    $t_i \leftarrow (w_i\, t_i + w_j\, t(u_j, v_j))/(w_i + w_j)$.
High $t_i$ indicates visually rich regions like boundaries and 
textured surfaces that are prioritized for subdivision even when geometric 
complexity is low.

\noindent\textbf{Deterministic complexity-threshold subdivision.}
MAP-ADAPT subdivides voxels according to manually specified semantic classes and geometric complexity thresholds.
We replace these class-based rules with per-modality complexity thresholds.
For a voxel at level $\ell \in \{\text{coarse}, \text{medium}\}$, subdivision is triggered if \emph{any} of the following holds:
    $H_i^{\text{sem}} > \tau_{\ell}^{\text{sem}}$,
    $g_i > \tau_{\ell}^{\text{geo}}$,
    $t_i > \tau_{\ell}^{\text{tex}}$.
If subdivision is triggered, the parent $v_i$ is replaced by $n_\ell$ of its children $\{v_i^{(k)}\}_{k=1}^{n_\ell}$, whose initial TSDF and semantic fields are interpolated from the parent and neighboring voxels.
This can be interpreted as refining voxels in which at least one modality exhibits high complexity.
The semantic threshold is shared across both levels, while the geometry and texture thresholds are level-specific for the coarse-to-medium and medium-to-fine transitions.
Defaults are $\tau^{\text{sem}}{=}0.10$, 
$\tau^{\text{geo}}_{\text{lo}}{=}0.05$, $\tau^{\text{geo}}_{\text{hi}}{=}0.10$, 
$\tau^{\text{tex}}_{\text{lo}}{=}1.0$, $\tau^{\text{tex}}_{\text{hi}}{=}2.0$,
calibrated by inspecting complexity distributions on a held-out training scene.
This deterministic variant removes dependence on predefined class lists from~\cite{zheng2024mapadapt}, but it still requires hand-tuning thresholds and is not budget-aware. 
\Cref{tab:threshold_sensitivity} analyzes sensitivity to these thresholds, showing that their selection jointly determines the accuracy--memory operating point.

\subsection{TSDF Integration and Field Updates}

For each frame $k$, we lift pixels $(u,v)$ with depth $z$ into 3D points $p = \pi^{-1}(u,v,z)$ and cast rays from the camera center to $p$ into the voxel grid. Voxels whose centers fall within a truncation band $[z - \mu, z + \mu]$ are updated.
Each observation $j$ contributes with weight $w_j = 1 / z_j^2$ (downweighting distant measurements).
For a voxel $v_i$, the TSDF value $d_i$ and its weight $w_i$ are updated as
\begin{equation}
    d_i \leftarrow \frac{w_i d_i + w_j d_j}{w_i + w_j},
    \qquad
    w_i \leftarrow w_i + w_j.
\end{equation}

\paragraph{Semantic fusion (logit accumulation).}
Instead of directly accumulating probabilities, each voxel maintains a vector of \emph{log-scores} (logits) $S_i \in \mathbb{R}^C$.
Given a per-frame class probability vector $P_j \in [0,1]^C$ at the projected point (with $\sum_c P_j(c)=1$), we form an updated logit proposal as
\begin{equation}
    S_i^{\text{prop}}(c) \;=\; S_i(c) + w_j \,\log\!\big(P_j(c) + \varepsilon\big),
\end{equation}
where $w_j$ is the observation weight and $\varepsilon$ is a small constant to avoid $\log 0$.
We apply an exponential moving average:
\begin{equation}
    S_i(c) \;\leftarrow\; (1-\lambda)\, S_i(c) \;+\; \lambda\, S_i^{\text{prop}}(c),
    \label{eq:logit_update}
\end{equation}
with fusion rate $\lambda \in (0,1]$.
Setting $\lambda=1$ recovers simple additive accumulation $S_i \leftarrow S_i + w_j \log P_j$.
To obtain a normalized class distribution for entropy or rendering, we compute the stable softmax as:
\begin{equation}
    \hat{P}_{i}^{\text{sem}}(c) \;=\; \frac{\exp\!\big(S_i(c) - m_i\big)}{\sum_{c'} \exp\!\big(S_i(c') - m_i\big)},
    \qquad
    m_i \;=\; \max_{c} S_i(c).
    \label{eq:softmax_stable}
\end{equation}
Finally, semantic entropy is obtained by:
\begin{equation}
    H_i^{\text{sem}} \;=\; - \sum_{c=1}^{C} \hat{P}_{i}^{\text{sem}}(c)\,\log \hat{P}_{i}^{\text{sem}}(c).
\end{equation}

\subsection{Budget-Aware RL Subdivision Policy (UnRL)}
\label{subsec:rl}

Although the uncertainty formulation above is principled, it still requires manual threshold selection and provides no direct mechanism to target a desired global storage budget.
Thus, we cast adaptive subdivision as a sequential decision problem and train an RL agent to choose complexity thresholds per region while penalizing budget overuse.

\noindent\textbf{Spatial blocking and uncertainty proxy.}
For each frame, we cluster voxels into spatial blocks of fixed size 
(\eg, $0.5\,\text{m}$) using a grid index. Let $\mathcal{B} = \{1,\dots,B\}$ 
be the set of blocks and $\mathcal{V}_b$ the voxels assigned to block $b$.
For each voxel $v_i \in \mathcal{V}_b$, we compute a normalized uncertainty proxy as
    $h_i = \tilde{w}_{\text{sem}} \tilde{H}_i^{\text{sem}} +
          \tilde{w}_{\text{geo}} \tilde{g}_i +
          \tilde{w}_{\text{tex}} \tilde{t}_i$,
where $\tilde{H}_i^{\text{sem}} = H_i^{\text{sem}} / \log C$, with $C$ the number of semantic classes,
$\tilde{g}_i = \min(g_i,\,0.333)/0.333$, and
$\tilde{t}_i = \min(t_i,\,3.0)/3.0$ denote the per-voxel semantic entropy, 
geometric complexity, and texture complexity normalized to $[0,1]$, 
and $\tilde{w}_{\text{sem}} = \tilde{w}_{\text{geo}} = \tilde{w}_{\text{tex}} = 0.33$ 
are fixed equal weights used across all datasets.

For each block $b$, we compute its mean and standard deviation as:
\begin{equation}
    \mu_b = \frac{1}{|\mathcal{V}_b|} \sum_{i \in \mathcal{V}_b} h_i,
    \quad
    \sigma_b = \sqrt{\frac{1}{|\mathcal{V}_b|} \sum_{i \in \mathcal{V}_b} (h_i - \mu_b)^2}.
\end{equation}
We then importance-sample $N$ blocks with probabilities proportional to $\mu_b$, focusing the agent on informative regions over large uniform areas.
\Cref{tab:ablation_block_size,tab:ablation_num_blocks} analyze the spatial block size and number of sampled blocks, respectively, showing stable performance around the default configuration.

\noindent\textbf{Observation space.}
The agent observes the statistics of $N$ sampled blocks and global summary variables.
The observation state is as follows:
\begin{equation}
    \mathbf{s}
    =
    [\mu_1, \sigma_1, \dots, \mu_N, \sigma_N, \bar{h}, \sigma(h), \rho_{\text{used}}, \rho_{\text{budget}}],
\end{equation}
where $\bar{h}$ and $\sigma(h)$ are the mean and standard deviation of $h_i$ over all voxels in the frame, $\rho_{\text{used}} \in [0,1]$ is the fraction of the voxel budget currently consumed, and $\rho_{\text{budget}}$ is the fixed target budget ratio for the current agent.

\noindent\textbf{Action space and subdivision rule.}
For each sampled block $b$, the agent predicts a pair of complexity thresholds $(\tau_{0,b}, \tau_{1,b})$ from a discrete set
$\mathcal{T} = \{ 0.0, 0.025, 0.05, 0.1, 0.25, 0.5 \}$.
The environment encodes this as a MultiDiscrete action over $2N$ indices into $\mathcal{T}$. For voxels in block $b$, the predicted thresholds induce a three-level decision:
\begin{equation}
    \text{level}(v_i) =
    \begin{cases}
        \text{coarse}, & h_i \leq \tau_{0,b}, \\
        \text{medium}, & \tau_{0,b} < h_i \leq \tau_{1,b}, \\
        \text{fine},   & h_i > \tau_{1,b}.
    \end{cases}
\end{equation}
Each subdivision decision updates a \emph{voxel budget counter}, with coarse, medium, and fine levels incurring different costs; the medium level contributes a fraction $\alpha_{\text{mid}} \in (0,1)$ of the fine-level budget cost, serving as a cost-effective intermediate.

\noindent\textbf{Supervision losses and reward.}
To provide a training signal, we precompute high-resolution reference 
reconstructions at $1\,\text{cm}$ voxel size using Voxblox~\cite{voxblox}.
Each coarse ($8$\,cm) parent voxel maintains up to $8^3$ child slots at the 
finest ($1$\,cm) resolution, populated from this reference; only children with nonzero observation weight contribute to 
the loss computation.
For each lower-resolution voxel, we derive per-voxel losses by comparing its 
semantic label, RGB color, and TSDF value to these aggregated child voxels.

Concretely, $\mathcal{L}_{\text{sem}}$ is the fraction of child voxels whose 
semantic label differs from the parent (indicator-based, range $[0,1]$); 
$\mathcal{L}_{\text{tex}}$ is the mean L2 RGB distance between parent and 
children normalized by $\sqrt{3}$ (range $[0,1]$); and $\mathcal{L}_{\text{geo}}$ 
is the mean L1 TSDF difference normalized by the truncation distance $2\mu$ 
(range $[0,1]$). We denote these normalized losses by 
$\mathcal{L}_{\text{sem}}, \mathcal{L}_{\text{tex}}, \mathcal{L}_{\text{geo}} \in [0,1]$.
They are combined into a single reconstruction loss as follows:
\begin{equation}
    \mathcal{L} =
    w_{\text{sem}} \mathcal{L}_{\text{sem}} +
    w_{\text{tex}} \mathcal{L}_{\text{tex}} +
    w_{\text{geo}} \mathcal{L}_{\text{geo}},
\end{equation}
with equal weights $w_{\text{sem}} = w_{\text{tex}} = w_{\text{geo}} = 0.33$,
fixed across all datasets and shared with the proxy weights used in $h_i$. 
\Cref{tab:ablation_weights} evaluates sensitivity to the reward-weight configuration and shows that balanced settings yield similar performance.
For voxel $v_i$, we denote the resulting combined loss by $\mathcal{L}_i$.

At each step, we compute a per-voxel reward that encourages subdivision where it reduces reconstruction loss while penalizing unnecessary refinement.
For a voxel assigned to level $\ell \in \{\text{coarse}, \text{medium}, \text{fine}\}$, we use:
\begin{equation}
    r_i^{(\ell)} =
    \begin{cases}
        - \mathcal{L}_i, & \ell = \text{coarse}, \\
        \alpha_{\text{mid}} (\mathcal{L}_i - \lambda_{\text{sub}}), & \ell = \text{medium}, \\
        \mathcal{L}_i - \lambda_{\text{sub}}, & \ell = \text{fine},
    \end{cases}
\end{equation}
where $\lambda_{\text{sub}} > 0$ is a subdivision penalty and $\alpha_{\text{mid}} \in (0,1)$ downweights the effect of medium resolution relative to fine.
Intuitively, subdividing is only beneficial when the expected improvement in reconstruction outweighs the subdivision cost.
The step reward is obtained by averaging over all updated voxels and computing a temporal difference relative to the previous step to stabilize learning.

\noindent\textbf{Budget penalty.}
Let $N_{\text{large}}$ be the number of $8\,\text{cm}$ voxels required to 
fill the bounding box of the scene, computed once at episode 
initialization from the known scene extent.
We define a per-scene budget as
$N_{\text{budget}} = \rho_{\text{budget}} N_{\text{large}}$,
where $\rho_{\text{budget}} \in (0,1]$ is the fixed budget ratio for the current agent.
Given the fixed storage footprint $s_v$ (in bytes) per voxel, this corresponds to a target memory $M_{\text{target}} = N_{\text{budget}} \, s_v$, \ie, a target storage budget in MB or GB.
Let $N_{\text{sub}}$ denote the current effective number of subdivided voxels (medium and fine levels, accounting for their relative cost).
We compute the used fraction as 
    $\rho_{\text{used}} = \min\left(N_{\text{sub}}/N_{\text{budget}},\,1.0\right)$.
A penalty is applied when $\rho_{\text{used}}$ exceeds an offset $\rho_0$ (\eg, $0.7$ of the allocated memory) as:
\begin{equation}
    r_{\text{budget}} =
    \begin{cases}
        0, & \rho_{\text{used}} \leq \rho_0, \\
        \beta_{\text{budget}}\, (\rho_{\text{used}} - \rho_0), & \rho_{\text{used}} > \rho_0,
    \end{cases}
\end{equation}
where $\beta_{\text{budget}} < 0$ controls the severity of overuse.
The total reward becomes
    $r_t = \overline{r}_t + r_{\text{budget}}$,
with $\overline{r}_t$ denoting the mean voxel-level reward.
This formulation applies a \emph{budget-aware soft constraint}: rather than a formal hard constraint with satisfaction guarantees, the penalty makes sustained budget violations suboptimal during training, empirically yielding tight budget  adherence at inference.

\noindent\textbf{Training.}
We train the policy using PPO~\cite{schulman2017proximal}, following a reinforcement-learning formulation in which each episode corresponds to a complete scene and the agent sequentially selects thresholds for blocks across frames. We train separate agents for different fixed budget ratios $\rho_{\text{budget}}$, each targeting a different memory regime.

Inspired by RL-VO~\cite{messikommer2024reinforcement}, which uses cross-attention for variable-length RL observations, we adopt a PerceiverIO architecture~\cite{perceiverio} for the variable number of sampled blocks $N$. Each block's $(\mu_b, \sigma_b)$ pair is a key--value input; four learned latent queries compress all block features into a fixed 256-dim representation processed by an MLP to produce the action distribution over $2N$ threshold indices. The value network additionally receives the global critic features $(\bar{h},\,\sigma(h),\,\rho_{\text{used}},\,\rho_{\text{budget}})$, focusing the critic on budget context.

We train our reinforcement-learning agent on 50 scenes randomly sampled from the
ScanNet training split~\cite{dai2017scannet}, disjoint from all evaluation scenes.
Each scene provides a sequence of RGB-D frames with corresponding camera poses and
precomputed $1$\,cm Voxblox reference reconstructions used for reward computation.
Training is performed for a total of 10 million environment steps. Each policy is
trained separately for approximately 24 hours on a single consumer-grade RTX 3080
(10 GB), using the precomputed 1 cm reference reconstructions for reward supervision.
At inference, these references are no longer required: the user simply selects a
desired storage-budget regime, and the corresponding policy predicts subdivision
thresholds directly from the observed block uncertainty statistics
$\{\mu_b, \sigma_b\}$. Additional training details, including hardware and total
training cost, are provided in \Cref{sec:hardware}.

\section{Experiments}
\label{sec:experiments}

We evaluate geometric reconstruction, semantic accuracy, texture fidelity, and efficiency on both synthetic and real-world benchmarks, comparing fixed-resolution baselines, MAP-ADAPT, and our uncertainty- and RL-driven variants.

\begin{table}[tb]
\caption{\textbf{Accuracy and efficiency on ScanNet.}
We report geometric reconstruction quality (completion error and completion ratio within 5\,cm), semantic accuracy, and texture fidelity (MS-SSIM, LPIPS), all obtained from mesh renderings in held-out test views.
Efficiency metrics include total voxel count and peak memory usage in gigabytes.
Our complexity- and uncertainty-driven method (\ablUncertaintyGeoTexture) outperforms MAP-ADAPT in \textit{all} accuracy metrics while avoiding reliance on manually specified semantic class lists.
The RL-based variants enable explicit control of memory through voxel-budget constraints, offering competitive accuracy with reduced memory.}
\label{tab:accuracy_scannet_full}
\centering
\setlength{\tabcolsep}{4pt}
\resizebox{\textwidth}{!}{
\begin{tabular}{lccccccc}
\toprule
\textbf{Method}  &  \textbf{Compl. Err$\downarrow$}  &  \textbf{Compl.$<$5cm (\%)$\uparrow$}  &  \textbf{Sem. Acc.$\uparrow$}  &  \textbf{MS-SSIM$\uparrow$}  &  \textbf{LPIPS$\downarrow$}  &  \textbf{Total Voxels (M)}  &  \textbf{Memory (GB)$\downarrow$} \\
\midrule
\multicolumn{8}{l}{\textbf{Fixed-Resolution Baselines}} \\
\ablLarge~\cite{voxblox}  &  3.09  &  81.05  &  0.605  &  0.61  &  0.52  &  \textbf{0.64}  &  \textbf{0.15}  \\
\ablMiddle~\cite{voxblox}  &  2.17  &  91.43  &  0.625  &  0.64  &  0.46  &  1.12  &  0.26  \\
\ablSmall~\cite{voxblox}  &  \textbf{1.44}  &  \textbf{99.24}  &  \textbf{0.628}  &  \textbf{0.65}  &  \textbf{0.40}  &  35.5  &  8.46  \\
\midrule
\multicolumn{8}{l}{\textbf{Adaptive Methods (w/o Budget Constraint)}} \\
\ablMapadaptGeo~\cite{zheng2024mapadapt}  &  1.57  &  98.05  &   0.623  &  0.64  &  0.42  &  \textbf{26.6}  &  \textbf{6.35} \\
\ablMapadaptSplitAvg  & 1.57 $\pm$ 0.038 & 98.16 $\pm$  0.332 & 0.622 $\pm$ 0.001 & 0.64 $\pm$ 0.001 & 0.43 $\pm$ 0.001 & 28.20 $\pm$ 0.283 & 6.724 $\pm$ 0.672 \\
\ablUncertaintyGeoTexture  &  \textbf{1.48}  &  \textbf{98.87}   &  \textbf{0.624}  &  \textbf{0.65}  &  \textbf{0.41}  &  33.7  &  8.05 \\
\midrule
\multicolumn{8}{l}{\textbf{Adaptive Methods (w/ Budget Constraint)}} \\
\ablUncertaintyGeoTextureRL (75\%)  &  \textbf{1.72}  &  \textbf{95.54}  &  \textbf{0.610}  &  \textbf{0.61}  &  \textbf{0.45}  &  12.1  &  2.88 \\
\ablUncertaintyGeoTextureRL (25\%)  &  2.48  &  88.18  &  0.599  &  0.59  &  0.50  &  \textbf{3.8}  &  \textbf{0.92} \\
\bottomrule
\end{tabular}
}
\end{table}
\begin{table}[tb]
\caption{\textbf{Accuracy and efficiency on HSSD.}
Evaluation on synthetic HSSD scenes, where the RL models are \emph{not} trained on HSSD but transferred directly from ScanNet.
We measure geometric (completion error and ratio $<$5\,cm), semantic, and texture (MS-SSIM, LPIPS) accuracy, together with efficiency (voxel count, memory).
Our complexity- and uncertainty-driven method surpasses MAP-ADAPT without requiring any dataset-specific semantic class list.
The RL-based variants maintain strong performance under strict memory budgets and generalize robustly to unseen synthetic environments.}
\label{tab:accuracy_hssd_full}
\centering
\setlength{\tabcolsep}{4pt}
\resizebox{\textwidth}{!}{
\begin{tabular}{lccccccc}
\toprule
\textbf{Method}  &  \textbf{Compl. Err$\downarrow$}  &  \textbf{Compl.$<$5cm (\%)$\uparrow$}  &  \textbf{Sem. Acc.$\uparrow$}  &  \textbf{MS-SSIM$\uparrow$}  &  \textbf{LPIPS$\downarrow$}  &  \textbf{Total Voxels (M)}  &  \textbf{Memory (GB)$\downarrow$} \\
\midrule
\multicolumn{8}{l}{\textbf{Fixed-Resolution Baselines}} \\
\ablLarge~\cite{voxblox}   &  3.88  &  76.90  &  0.61  &  0.69  &  0.45  &  \textbf{0.7}  &  \phantom{1}\textbf{0.16} \\
\ablSmall~\cite{voxblox}  &  \textbf{3.18}  &  \textbf{86.55}  &  \textbf{0.62}  &  \textbf{0.72}  &  \textbf{0.31}  &  98.1  &  23.39  \\
\midrule
\multicolumn{8}{l}{\textbf{Adaptive Methods (w/o Budget Constraint)}} \\
\ablMapadaptGeo~\cite{zheng2024mapadapt}  &  3.52  &  82.33  &  0.61  &  \textbf{0.69}  &  0.39  &  \textbf{26.6}  &  \phantom{1}\textbf{6.34} \\
\ablUncertaintyGeoTexture  &  \textbf{3.35}  &  \textbf{84.36}  &  \textbf{0.62}  &  \textbf{0.69}  &  \textbf{0.35}  &  67.3  &  16.05 \\
\midrule
\multicolumn{8}{l}{\textbf{Adaptive Methods (w/ Budget Constraint)}} \\
\ablUncertaintyGeoTextureRL (75\%)  &  \textbf{3.49}  &  \textbf{82.69}  &  \textbf{0.61}  &  \textbf{0.67}  &  \textbf{0.36}  &  59.5  &  14.20 \\
\ablUncertaintyGeoTextureRL (25\%)  &  3.75  &  78.75  &  0.60  &  \textbf{0.67}  &  0.44  &  \textbf{7.7}  &  \phantom{1}\textbf{1.85} \\
\bottomrule
\end{tabular}
}
\end{table}

\noindent\textbf{Datasets.}
Following MAP-ADAPT~\cite{zheng2024mapadapt}, we evaluate on 81 scenes (all disjoint from the 50 ScanNet scenes used for RL training):
(i) \textbf{\textit{HSSD}}~\cite{hssd2023} provides photorealistic synthetic indoor scenes with noise-free ground-truth geometry, semantics, and textures.
(ii) \textbf{\textit{ScanNet}}~\cite{dai2017scannet} contains real-world RGB-D scans with measurement noise, missing depth, and partial semantic annotations, testing robustness under realistic conditions.

\noindent\textbf{Metrics.}
All metrics are computed on meshes extracted from the adaptive TSDF grids.
(i) \textbf{\textit{Geometry}:} mean completion error and completion ratio within 5\,cm.
(ii) \textbf{\textit{Semantics}:} vertex labels from the fused voxel distribution are rendered into held-out test views via raycasting and compared to ground-truth semantic projections, penalizing both label errors and geometric misalignment.
(iii) \textbf{\textit{Texture}:} MS-SSIM and LPIPS~\cite{zhang2018unreasonable} on raycasted color renderings.
(iv) \textbf{\textit{Efficiency}:} total voxel count, peak memory, and per-frame integration time.
All adaptive methods share the same TSDF integration and meshing pipelines~\cite{zheng2024mapadapt}.

\noindent\textbf{Baselines and Ablations.}
\begin{itemize}
    \item \textbf{Voxblox (8\,cm, 4\,cm, 1\,cm)~\cite{voxblox}:} Fixed-resolution baselines at each of the three levels that the adaptive methods select among during subdivision. The 1\,cm grid provides a quality upper bound at prohibitive memory cost.

    \item \textbf{\ablMapadaptGeo~\cite{zheng2024mapadapt}:} Uses geometric CoC~\cite{pauly2002efficient} and a \emph{manually selected semantic class list}, limiting generalization. We use the official HSSD configuration; for ScanNet, we adopt the same list with unknown classes assigned to coarse resolution.

    \item \textbf{\ablMapadaptSplitAvg~\cite{zheng2024mapadapt}:} MAP-ADAPT averaged over 5 runs with randomly sampled semantic class lists, where each list specifies which class labels trigger fine-resolution subdivision. This evaluates sensitivity to class list selection when no dataset-specific configuration is available.

    \item \textbf{\ablUncertaintyGeoTexture:} Our method using semantic entropy, geometric complexity, and texture complexity as subdivision signals, without user-defined class lists.

    \item \textbf{\ablUncertaintyGeoTextureRL:} Budget-aware RL refinement trained at 25\% and 75\% target voxel-budget regimes, providing budget-aware refinement at inference.
\end{itemize}
We omit Multi-TSDF~\cite{schmid2022panoptic}, which has been shown to underperform MAP-ADAPT.
We further evaluate robustness to the semantic segmentation backbone by replacing RefineNet with SegFormer; \Cref{tab:accuracy_scannet_segformer_full} shows consistent accuracy--memory behavior with the alternative backbone.

\noindent\textbf{Results on ScanNet.}
\Cref{tab:accuracy_scannet_full} shows that our uncertainty-driven refinement outperforms MAP-ADAPT on almost all metrics.
\ablUncertaintyGeoTexture achieves lower completion error and higher \emph{Compl.\,$<$5cm} than MAP-ADAPT, and also improves semantic accuracy and photometric quality.
These improvements primarily arise from \emph{automatically} identifying informative or uncertain regions for refinement, rather than relying on a preselected class list.
Importantly, this enables generalization to scenes containing object categories not explicitly known during system design.
Because our method additionally refines texture-rich regions, which are ignored entirely by MAP-ADAPT, the unconstrained variant naturally allocates more voxels, as expected.
This additional cost directly motivates the budget-aware RL variant, which recovers the efficiency advantage while preserving the broader refinement criteria.

The RL-based variant learns predictable trade-offs between memory and accuracy.
At 75\% of the voxel budget, the reconstruction quality is close to our deterministic model, while memory is reduced by more than half.
These results confirm that the agent learns to allocate resources where they most reduce reconstruction loss.
At extreme budgets (25\%), the policy prioritizes high uncertainty and complexity regions, maintaining key structural and semantic details, and gracefully reducing resolution elsewhere.

\begin{table}[tb]
\caption{\textbf{Ablation of refinement criteria on ScanNet}~\cite{dai2017scannet}.
Adding geometric complexity improves reconstruction completeness and semantic consistency, while incorporating texture complexity further enhances all metrics, demonstrating that all three refinement criteria contribute complementary information.}
\label{tab:ablations_scannet}
\centering
\resizebox{0.56\textwidth}{!}{\begin{tabular}{l|ccc}
\toprule
\textbf{Method} & \textbf{Compl. Err$\downarrow$} & \textbf{Sem. Acc.$\uparrow$} & \textbf{MS-SSIM$\uparrow$} \\
\midrule
\textsc{Sem} & 1.624 & 0.620 & 0.637 \\
\textsc{Sem}+\textsc{Geo} & 1.546 &  0.622 & 0.642 \\
\textsc{Sem}+\textsc{Geo}+\textsc{Tex} &  \textbf{1.483} & \textbf{0.624} & \textbf{0.645} \\
\bottomrule
\end{tabular}}
\end{table}

\noindent\textbf{Results on HSSD.}
\Cref{tab:accuracy_hssd_full} reports results on the synthetic HSSD scenes.
Although the RL agent and all hyperparameters were trained on ScanNet, our uncertainty-driven model again surpasses MAP-ADAPT across geometric and texture metrics.
This demonstrates a key strength of our formulation: since voxel refinement is triggered by modality-specific uncertainty and complexity signals rather than predefined semantic categories, the method naturally generalizes to scenes containing previously unseen objects, materials, and textures.

The RL-based models show the same trend observed on ScanNet.
With a moderate budget (75\%), reconstruction accuracy remains nearly indistinguishable from the unconstrained version, while memory usage is substantially reduced.
Even under tight budgets (25\%), the policy maintains stable reconstructions, consistently preserving fine geometric structures and semantically meaningful regions despite the reduced allocation of high-resolution voxels.

\noindent\textbf{Runtime.}
The average per-frame integration times are shown in \Cref{tab:efficiency_hssd_runtime}.
Voxblox with large voxels (\ablLarge) achieves the highest throughput, but at the cost of substantially lower reconstruction accuracy. The uniform fine-resolution baseline (\ablSmall) remains computationally efficient, but requires considerably higher memory.
Across all adaptive methods, runtime remains within a similar range of 19-26\,FPS.
This demonstrates that both heuristic and learned adaptive subdivision preserve real-time performance.
The RL-based approach achieves a favorable trade-off: it adapts refinement to the available voxel budget, avoiding unnecessary refinement and maintaining consistent frame rates even under tight memory settings.

\noindent\textbf{Memory.}
Uniform fine-grid reconstruction (\ablSmall) requires between 20--25\,GB on larger HSSD scenes, making it impractical for memory-constrained real-time systems.
As shown in \Cref{tab:accuracy_scannet_full,tab:accuracy_hssd_full}, our deterministic uncertainty-driven method (\textsc{Un-Map-Adapt}) prioritizes reconstruction quality by refining regions according to semantic uncertainty, geometric complexity, and texture complexity, but does not explicitly constrain memory usage. The RL-based variants address this limitation by targeting scene-specific voxel budgets and providing explicit control over the memory--accuracy trade-off. Under the 25\% setting, \textsc{UnRL} reduces memory by $9.2\times$ on ScanNet and $12.6\times$ on HSSD relative to the fine-resolution Voxblox baseline, while trading some reconstruction accuracy for substantially lower memory usage.
This budget-aware behavior makes the RL policies particularly well-suited for systems operating under strict memory constraints, such as mobile or embedded platforms.

\begin{figure}[tb]
\centering
\setlength{\tabcolsep}{1pt}
\renewcommand{\arraystretch}{0.75}

\resizebox{\linewidth}{!}{\begin{tabular}{cccccc}
\scriptsize \textsc{Voxblox} (8cm) &
\scriptsize \textsc{UnRL} (25\%) &
\scriptsize \textsc{UnRL} (75\%) &
\scriptsize \ablMapadaptGeo &
\scriptsize \textsc{Un-Map-Adapt} &
\scriptsize \textsc{Voxblox} (1cm) \\

\includegraphics[width=0.155\linewidth]{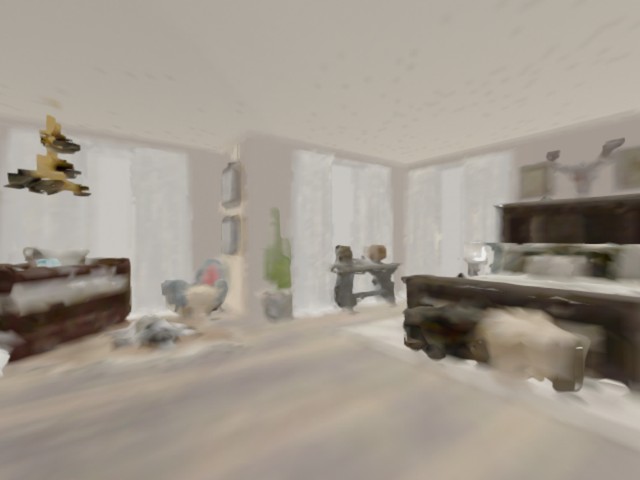} &
\includegraphics[width=0.155\linewidth]{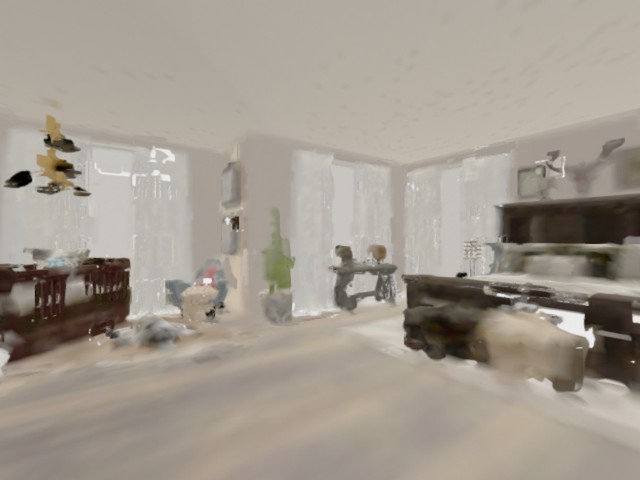} &
\includegraphics[width=0.155\linewidth]{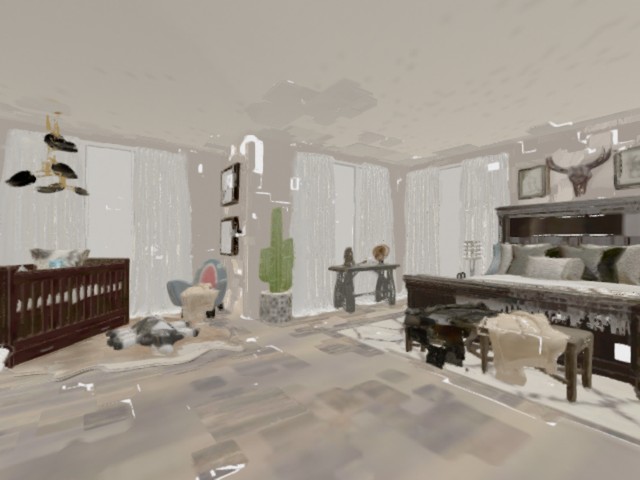} &
\includegraphics[width=0.155\linewidth]{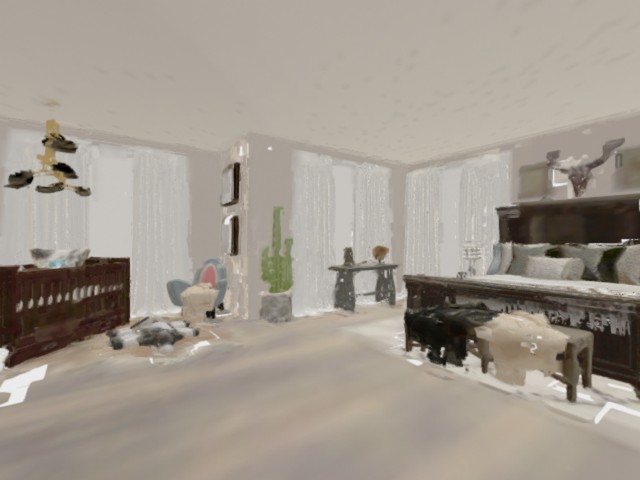} &
\includegraphics[width=0.155\linewidth]{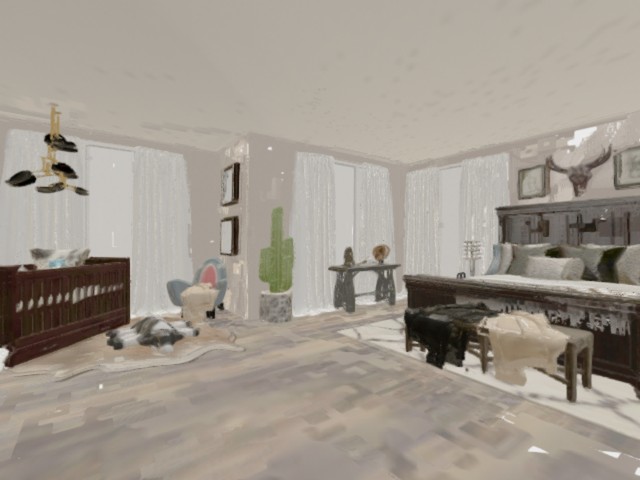} &
\includegraphics[width=0.155\linewidth]{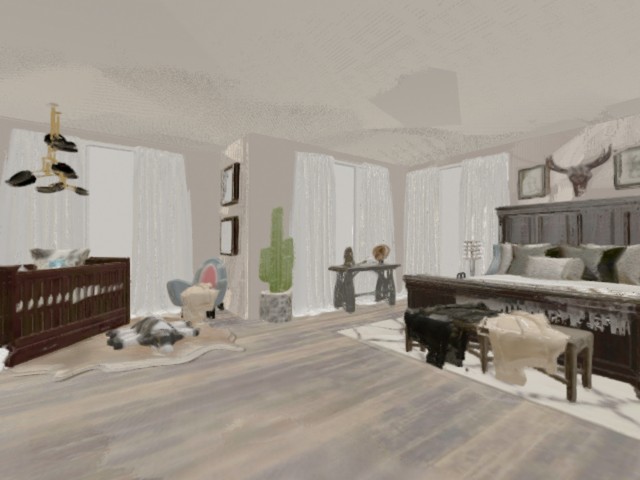} \\

\includegraphics[width=0.155\linewidth]{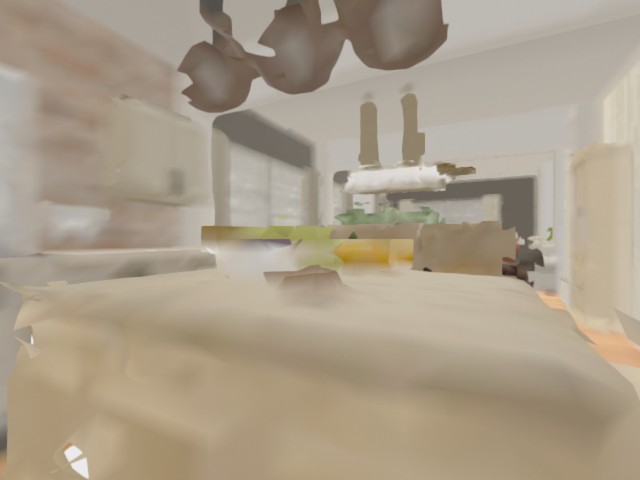} &
\includegraphics[width=0.155\linewidth]{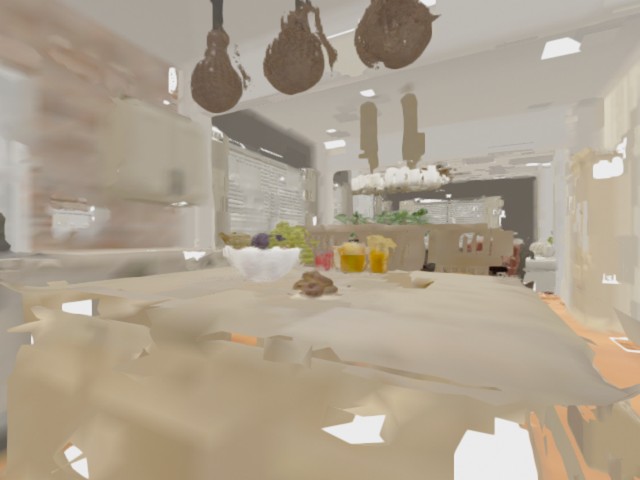} &
\includegraphics[width=0.155\linewidth]{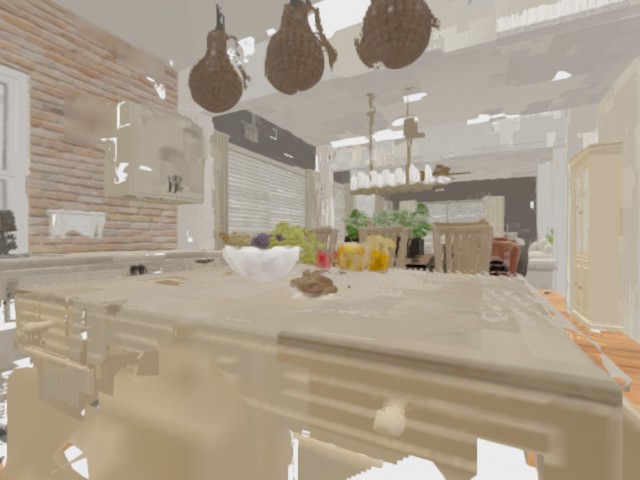} &
\includegraphics[width=0.155\linewidth]{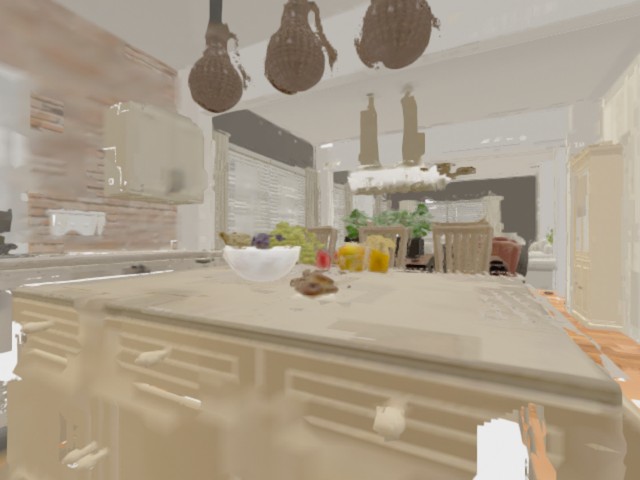} &
\includegraphics[width=0.155\linewidth]{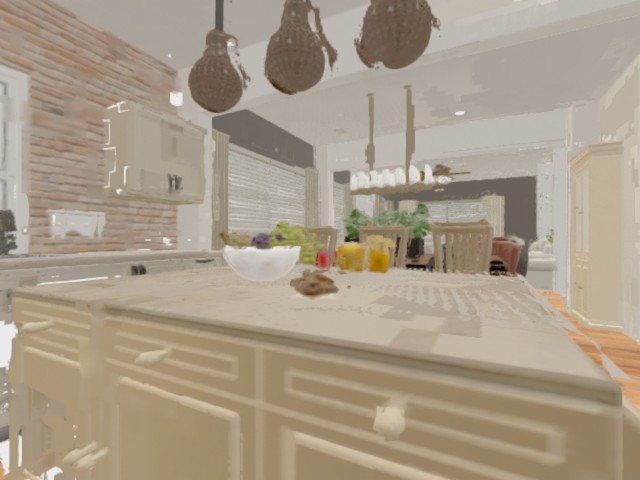} &
\includegraphics[width=0.155\linewidth]{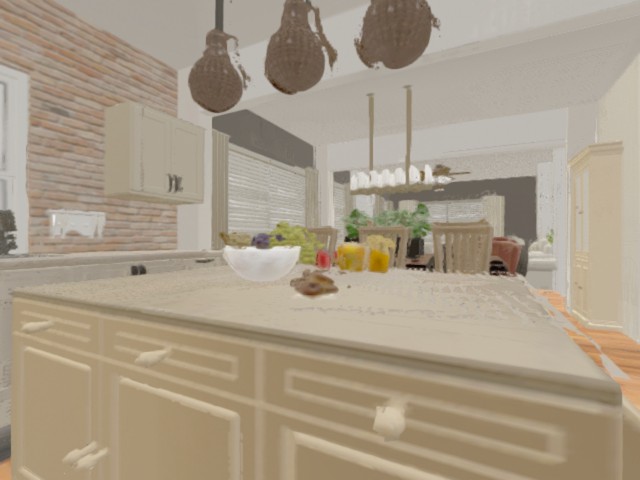} \\

\includegraphics[width=0.155\linewidth]{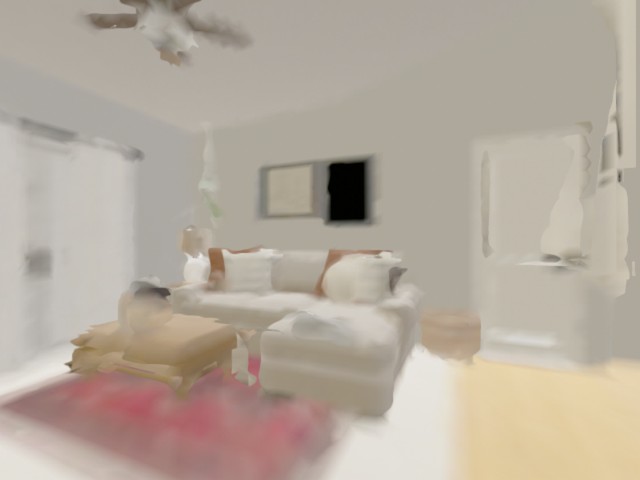} &
\includegraphics[width=0.155\linewidth]{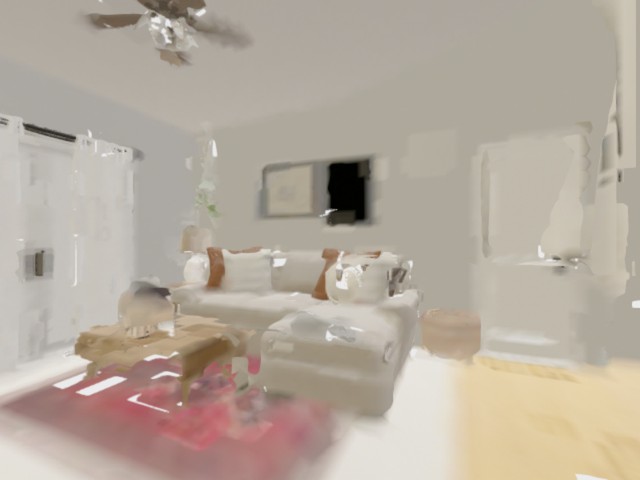} &
\includegraphics[width=0.155\linewidth]{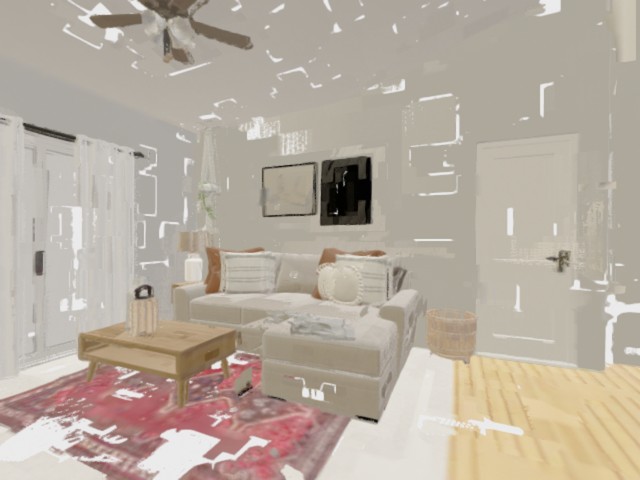} &
\includegraphics[width=0.155\linewidth]{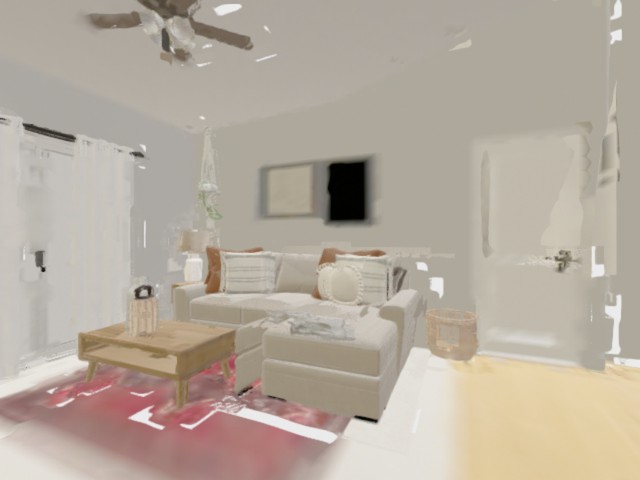} &
\includegraphics[width=0.155\linewidth]{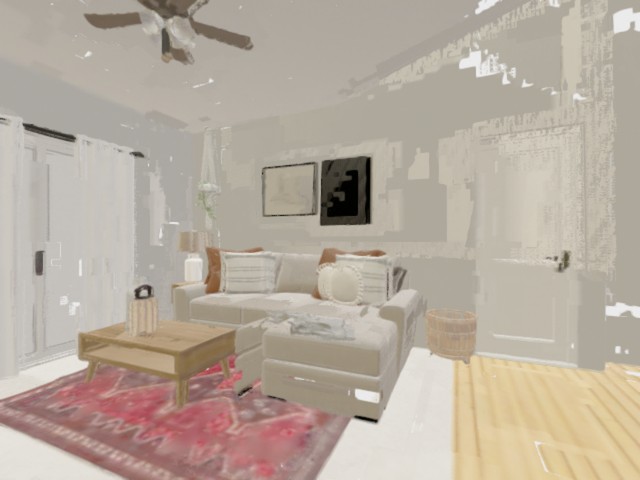} &
\includegraphics[width=0.155\linewidth]{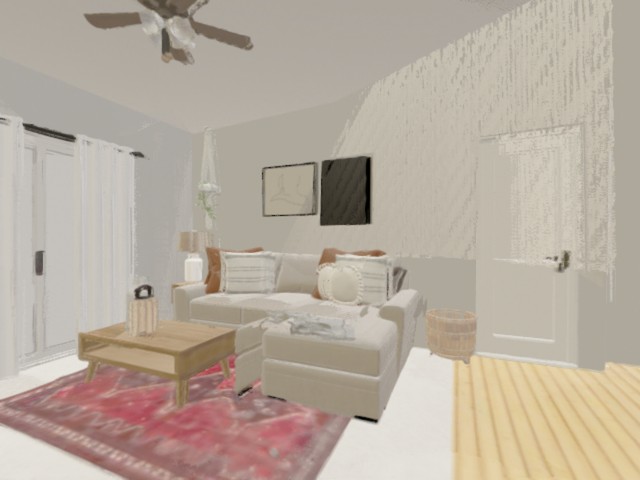} \\

\includegraphics[width=0.155\linewidth]{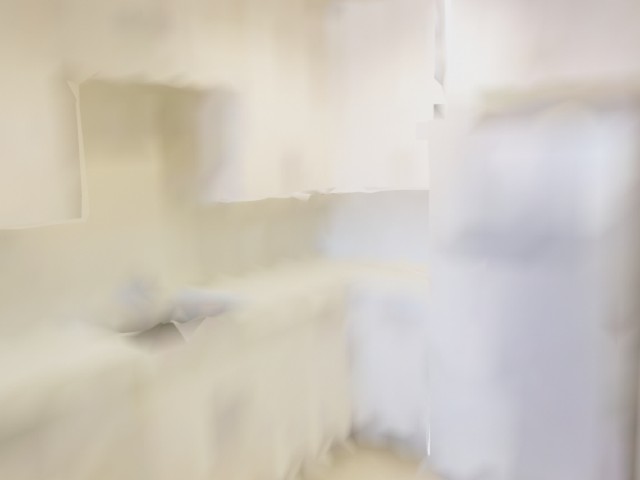} &
\includegraphics[width=0.155\linewidth]{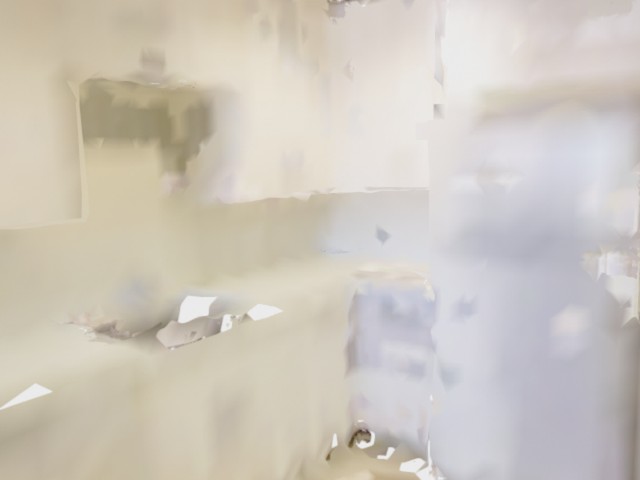} &
\includegraphics[width=0.155\linewidth]{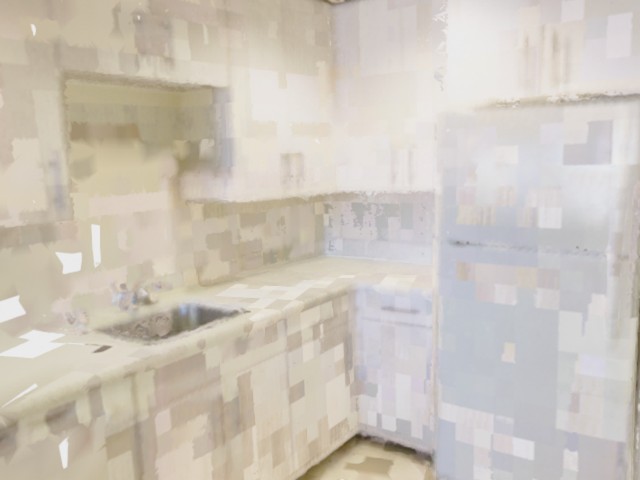} &
\includegraphics[width=0.155\linewidth]{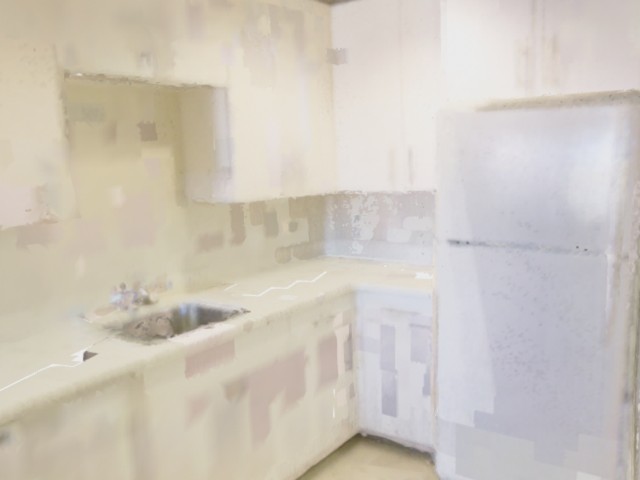} &
\includegraphics[width=0.155\linewidth]{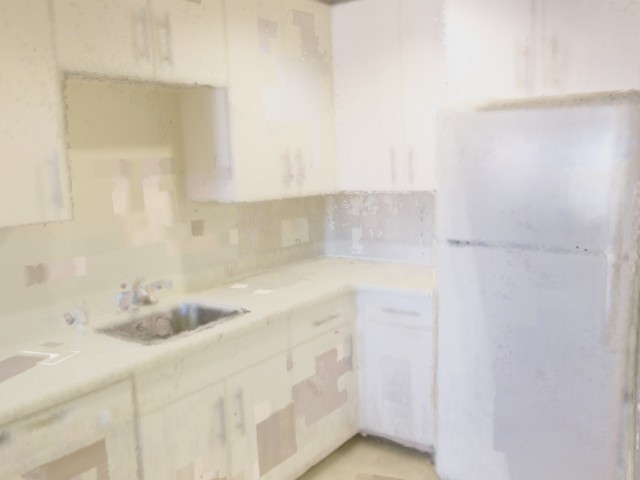} &
\includegraphics[width=0.155\linewidth]{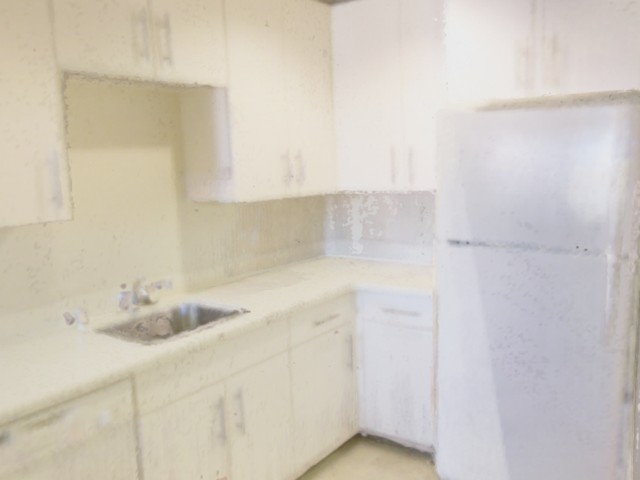} \\

\includegraphics[width=0.155\linewidth]{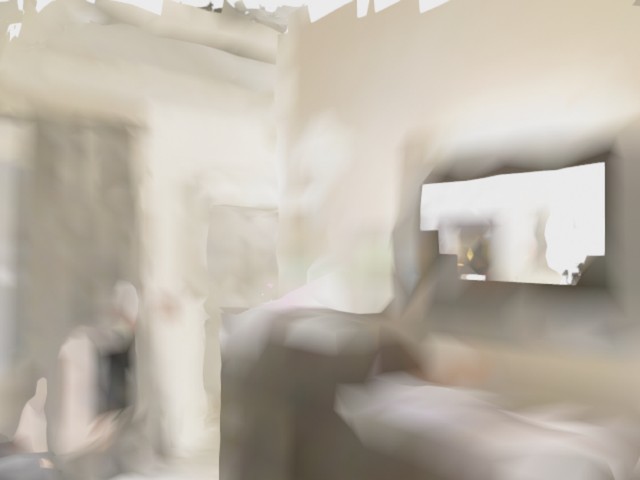} &
\includegraphics[width=0.155\linewidth]{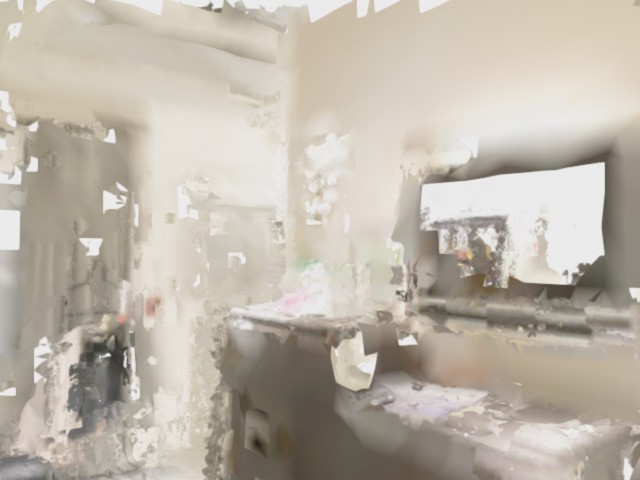} &
\includegraphics[width=0.155\linewidth]{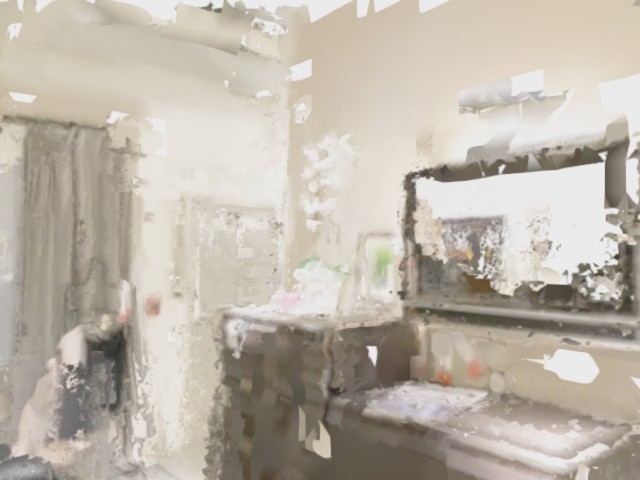} &
\includegraphics[width=0.155\linewidth]{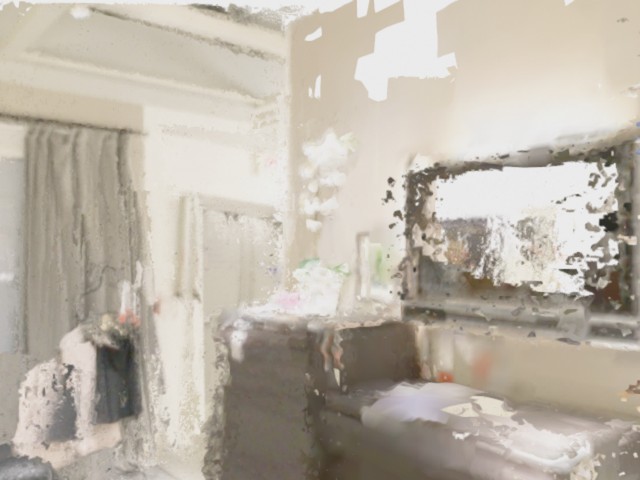} &
\includegraphics[width=0.155\linewidth]{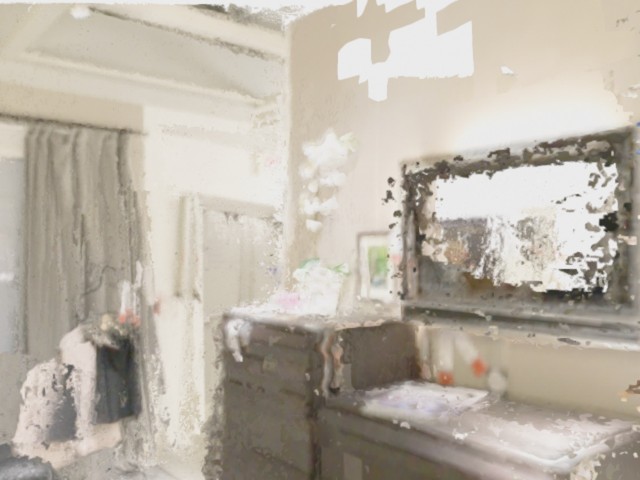} &
\includegraphics[width=0.155\linewidth]{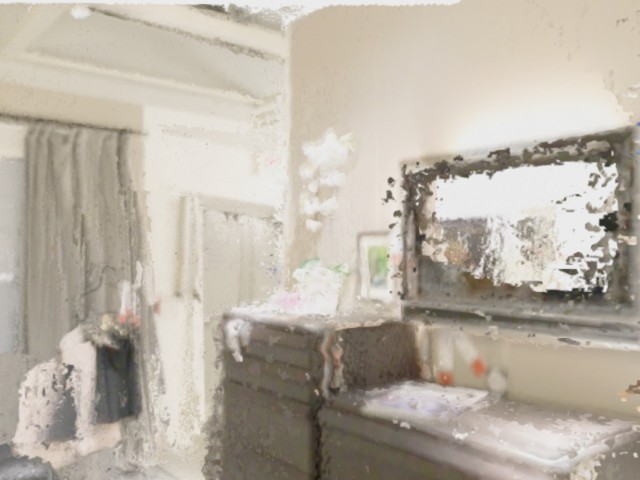} \\

\includegraphics[width=0.155\linewidth]{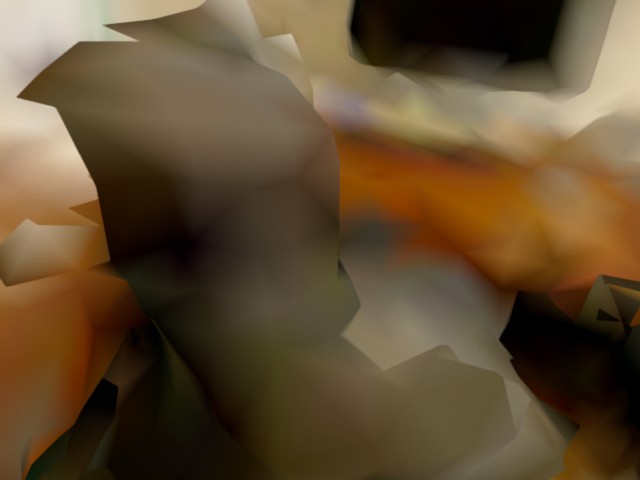} &
\includegraphics[width=0.155\linewidth]{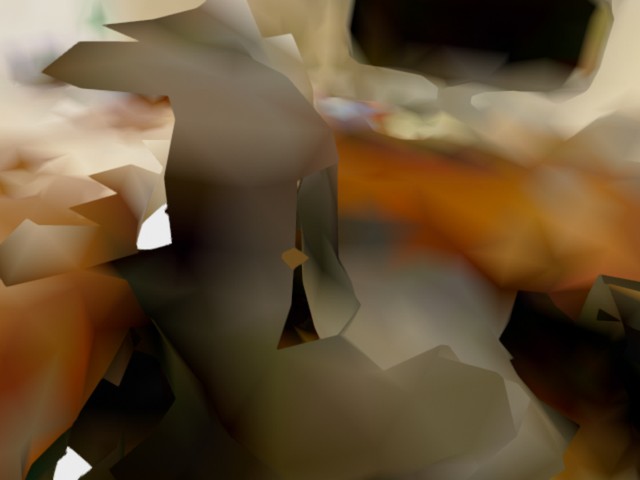} &
\includegraphics[width=0.155\linewidth]{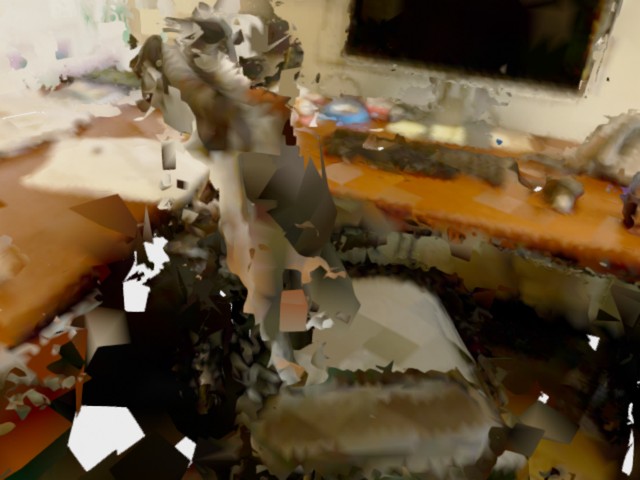} &
\includegraphics[width=0.155\linewidth]{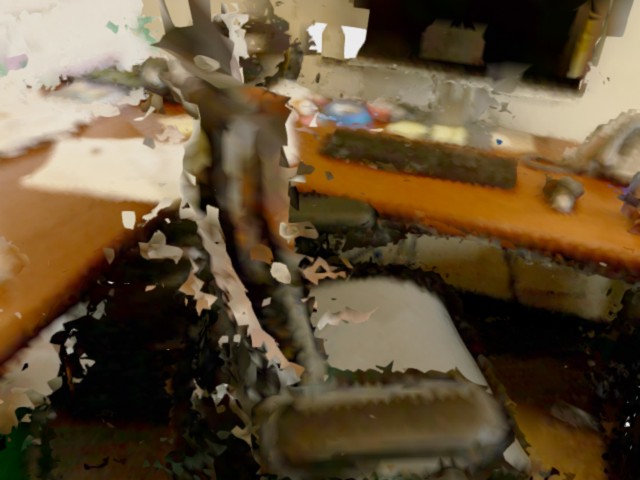} &
\includegraphics[width=0.155\linewidth]{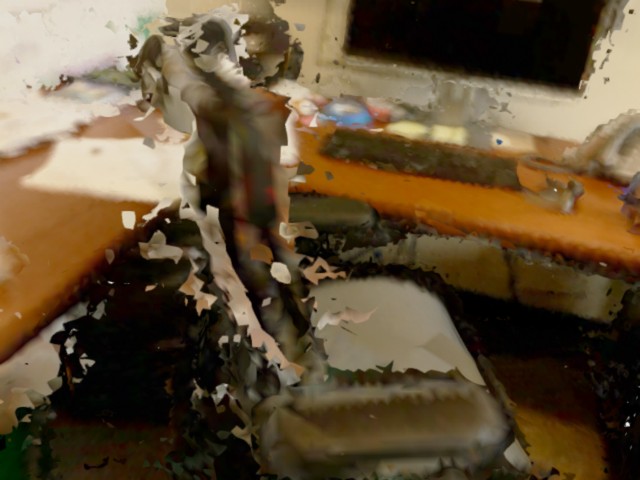} &
\includegraphics[width=0.155\linewidth]{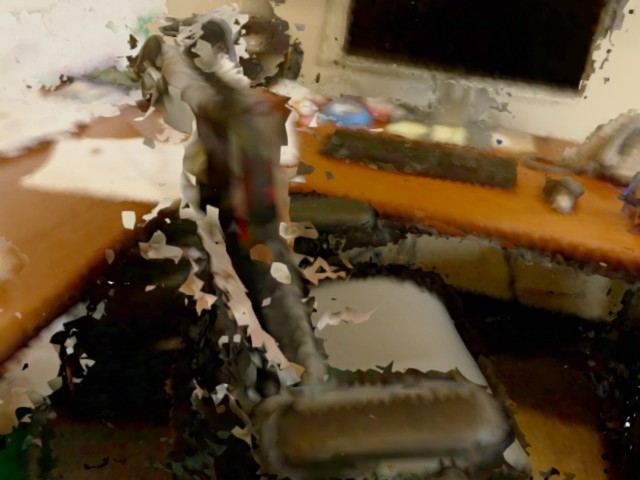} \\

\end{tabular}}
\caption{
\textbf{Qualitative comparison} on HSSD (top three rows) and ScanNet (bottom).
Columns show increasing reconstruction quality and detail from coarse \textsc{Voxblox} (8\,cm) to the fine-grained 1\,cm baseline.
Our deterministic uncertainty-driven method (\textsc{Un-Map-Adapt}) and our budget-aware RL model (\textsc{UnRL}) recover sharper geometry, cleaner object boundaries, and more stable textures than \textsc{Map-Adapt}, without requiring a hand-crafted semantic class list.
The RL variants (25\%, 75\%) adaptively preserve high uncertainty and complexity regions while respecting memory budgets, demonstrating controlled degradation under tight constraints.
}
\label{fig:qualitative_summary}
\end{figure}

\begin{table}[t]
\caption{Per-frame integration speed (FPS) on ScanNet. All methods run in real time.}
\label{tab:efficiency_hssd_runtime}
\centering
\resizebox{\textwidth}{!}{\begin{tabular}{l|cccccc}
\toprule
& \textsc{VoxBlox} (8 cm)
& \textsc{VoxBlox} (1 cm)
& \textsc{MAP-ADAPT}
& \ablUncertaintyGeoTexture
& \ablUncertaintyGeoTextureRL (25\%)
& \ablUncertaintyGeoTextureRL (75\%) \\
\midrule
\textbf{FPS}
& 37.94
& 28.75
& 26.24
& 23.77
& 24.01
& 19.16 \\
\bottomrule
\end{tabular}}
\end{table}

\noindent
\textbf{Ablation Study.}
We further ablate the contribution of the proposed refinement criteria used for adaptive subdivision. Specifically, we evaluate three variants of our approach:
(1) using only \textit{semantic entropy} (Sem),
(2) combining \textit{semantic entropy} and \textit{geometric complexity} (Sem+Geo), and
(3) the full model that additionally incorporates \textit{texture complexity} (Sem+Geo+Tex).

\Cref{tab:ablations_scannet} reports the results on ScanNet, where introducing geometric complexity improves completeness and semantic accuracy by focusing refinement on curved or structurally intricate geometry.
Adding texture complexity further enhances reconstruction fidelity and overall stability, as the model allocates fine voxels to visually rich regions.
These results confirm that integrating complementary refinement criteria leads to more balanced optimization between geometry, appearance, and semantics.

\noindent
\textbf{Qualitative Results.}
\Cref{fig:qualitative_summary} presents qualitative reconstructions on both ScanNet and HSSD, illustrating the effect of adaptive subdivision on geometric detail and texture fidelity.
Coarse Voxblox (8\,cm) produces heavily oversmoothed surfaces, while the fine 1\,cm grid recovers sharp structure but at prohibitive memory cost (as seen in \Cref{tab:accuracy_scannet_full,tab:accuracy_hssd_full}).
Our deterministic uncertainty-driven variant (Un-Map-Adapt) yields visibly sharper object contours, cleaner planar regions, and more stable texture reconstructions, particularly around edges where semantic entropy, geometric complexity, and texture complexity jointly signal the need for fine resolution.
The RL-based policies further demonstrate controlled adaptation: with moderate budgets (\eg, 75\%), reconstructions closely match the unconstrained model, whereas under tight constraints (\eg, 25\%), the system preserves the most informative regions such as object boundaries, high-curvature geometry, and textured surfaces, while simplifying homogeneous areas.
Across all settings, our method achieves a good balance between detail and compactness, and generalizes to HSSD despite the RL agent being trained only on ScanNet.

\section{Conclusion}

This work tackles the absence of explicit, user-controllable accuracy--memory trade-offs in current volumetric mapping pipelines. MAP-ADAPT offers limited control through manual class lists and class-dependent subdivision levels, but this requires expert tuning, fails to generalize, and becomes impractical in scenes with many small or unseen objects. Our method provides a principled and fully automatic alternative: uncertainty- and complexity-driven refinement removes dependence on semantic taxonomies, while a reinforcement-learning policy enables direct specification of a desired memory budget, with the system allocating resolution accordingly. Experiments on real and synthetic datasets show that our approach matches or surpasses MAP-ADAPT in geometric, semantic, and texture accuracy, while delivering predictable and stable performance under varying storage constraints. By making the target memory budget the primary user-facing parameter, our framework provides a flexible and scalable approach to adaptive 3D mapping. Our evaluation is currently limited to the indoor RGB-D setting of ScanNet and HSSD. Nevertheless, the semantic, geometric, and texture cues exploited by our approach are not intrinsically tied to indoor scene categories. Extending and adapting the method to outdoor and larger-scale mapping scenarios is an important direction for future work.

\clearpage
\appendix
\renewcommand{\theHsection}{\Alph{section}}
\section{Experiments with SegFormer}
\label[appendix]{sec:segformer}

This section provides additional experiments and visualizations.
We evaluate our method using SegFormer~\cite{xie2021segformer}, a stronger but heavier segmentation model than RefineNet~\cite{lin2017refinenet} used in the main paper.
We present extended qualitative comparisons, and analyze the behavior and generalization of the reinforcement learning (RL) agents.
Our goal is to demonstrate that (i) the proposed entropy-driven framework is robust to changes in the semantic backbone, (ii) the accuracy-memory trade-off remains consistent under different segmentation signals, and (iii) methods relying on fixed semantic class heuristics, such as MAP-ADAPT, are inherently brittle across datasets and segmentation models.

\subsection{Qualitative Comparisons}

\Cref{fig:qualitative_summary_sem} and \Cref{fig:qualitative_summary_rgb_segformer} present visualizations on HSSD and ScanNet for RefineNet and SegFormer, respectively.
We visualize textured reconstructions and semantic renderings from multiple viewpoints and across all ablations.

The coarse Voxblox baseline loses fine structure and produces over-smoothed surfaces. MAP-ADAPT improves detail, but remains sensitive to its hand-crafted class list, which often fails in the presence of new or ambiguous object categories. Our entropy-based variant yields crisp object boundaries and improved structural detail, while the RL models preserve high-entropy regions even under strong memory constraints. The 25\% and 50\% policies demonstrate controlled simplification, whereas the 75\% policy closely matches the full-resolution entropy-driven method.

Using SegFormer sharpens semantic boundaries and reveals a clearer separation between the methods: improvements in semantic quality are most visible for \ablUncertaintyGeoTexture and its RL variants, confirming the relevance of entropy-driven subdivision.

\subsection{ScanNet Benchmark with SegFormer}

\Cref{tab:accuracy_scannet_segformer_full} reports quantitative results on ScanNet with SegFormer. Reconstruction and texture metrics remain consistent with the main paper, while semantic accuracy increases across all methods due to the improved segmentation model.
We additionally evaluate an RL agent with a 50\% budget. As expected, accuracy increases smoothly from 25\% to 75\%, showing that the learned policy produces a stable and monotonic accuracy -- memory trade-off.

To study MAP-ADAPT's sensitivity to class heuristics, we introduce a random class-to-resolution assignment. Performance degrades sharply across all metrics, underscoring the method's reliance on user-engineered class lists and highlighting its lack of adaptability. As scenes, datasets, or segmentation backbones change, these heuristics may not transfer reliably. In contrast, our uncertainty- and complexity-driven method is class agnostic and relies on observed semantic, geometric, and texture cues rather than predefined semantic class assignments, supporting transfer across the tested segmentation backbones.

\subsection{Generalization of the RL Agents}

A key question is whether the RL policies learned on RefineNet generalize to SegFormer.
The policy does not consume semantic class identities. Instead, its semantic component
is based on normalized prediction entropy, while geometric and texture complexity provide
complementary class-independent cues. Specifically, the per-voxel proxy used to compute
the block-level statistics observed by the policy is
\begin{equation}
    h_i =
    \tilde{w}_{\mathrm{sem}} \tilde{H}_i^{\mathrm{sem}} +
    \tilde{w}_{\mathrm{geo}} \tilde{g}_i +
    \tilde{w}_{\mathrm{tex}} \tilde{t}_i,
    \qquad
    \tilde{H}_i^{\mathrm{sem}}
    =
    -\frac{1}{\log C}
    \sum_{c=1}^{C} \hat{P}_{i}^{\mathrm{sem}}(c) \log \hat{P}_{i}^{\mathrm{sem}}(c).
\end{equation}
Here, $\tilde{H}_i^{\mathrm{sem}}$ depends only on the shape of the predicted
class-probability distribution rather than the identity or ordering of the classes,
while $\tilde{g}_i$ and $\tilde{t}_i$ are independent of semantic class identities.
The reward similarly combines semantic, geometric, and texture reconstruction losses.
Consequently, the policy representation does not rely on a particular semantic class
configuration.

Results in \Cref{tab:accuracy_scannet_segformer_full} confirm this: the RL policies trained entirely on RefineNet preserve their ordering, monotonicity, and accuracy--memory trade-off when deployed with SegFormer, despite the two backbones differing substantially in architecture, capacity, and prediction characteristics.
This demonstrates that the agent learns a structural refinement strategy based on class-agnostic uncertainty and complexity cues, rather than overfitting to the statistics of a particular pretrained network.

\subsection{Qualitative Comparison: Semantic Maps}

\begin{figure}[t]
\centering
\setlength{\tabcolsep}{1pt}
\renewcommand{\arraystretch}{0.75}

\resizebox{\linewidth}{!}{\begin{tabular}{ccccccc}
\scriptsize \textsc{Ground Truth} &
\scriptsize \textsc{Voxblox} (8cm) &
\scriptsize \ablUncertaintyGeoTextureRL (25\%) &
\scriptsize \ablUncertaintyGeoTextureRL (75\%) &
\scriptsize \ablMapadaptGeo &
\scriptsize \ablUncertaintyGeoTexture &
\scriptsize \textsc{Voxblox} (1cm) \\

\includegraphics[width=0.155\linewidth]{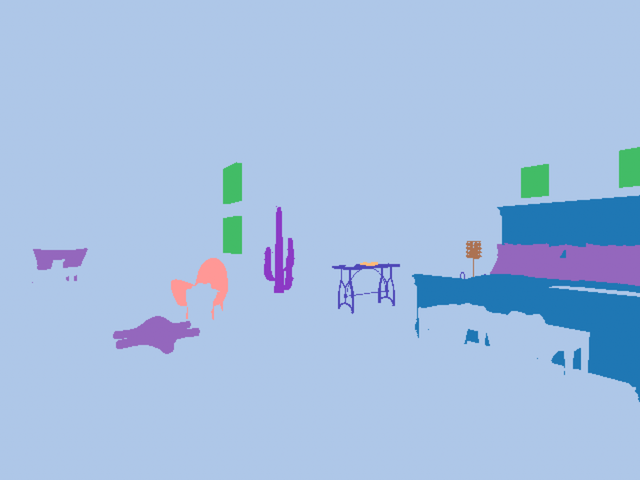} &
\includegraphics[width=0.155\linewidth]{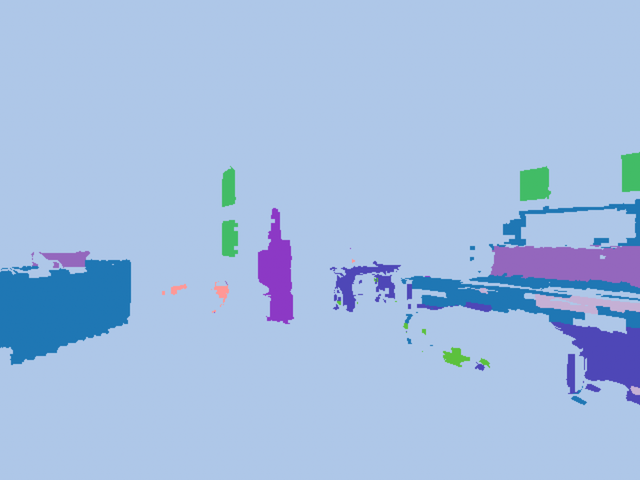} &
\includegraphics[width=0.155\linewidth]{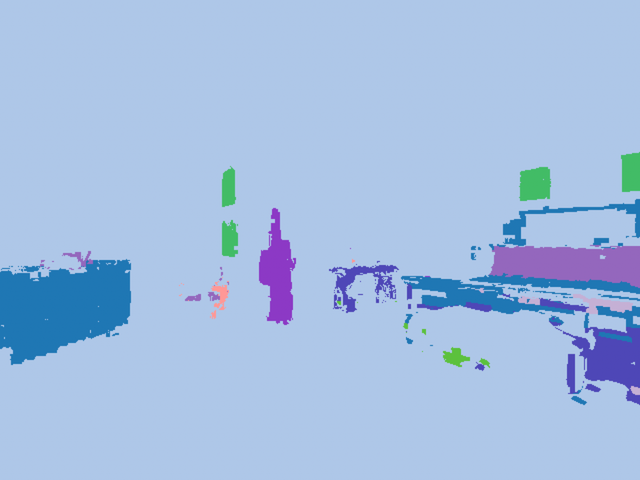} &
\includegraphics[width=0.155\linewidth]{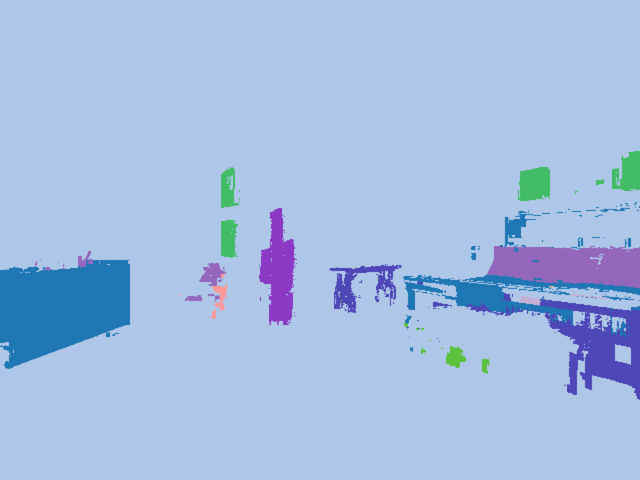} &
\includegraphics[width=0.155\linewidth]{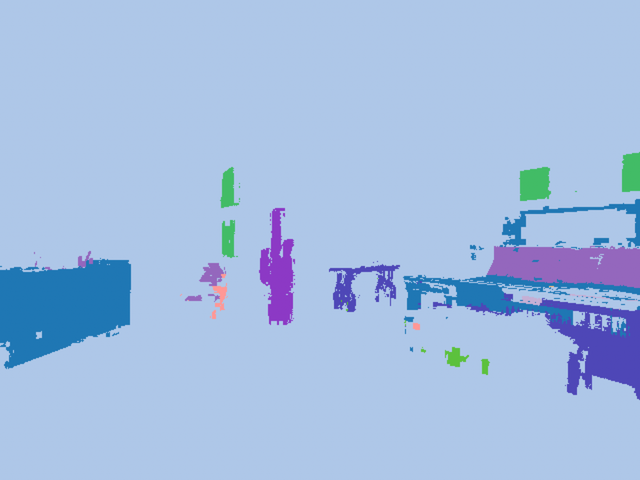} &
\includegraphics[width=0.155\linewidth]{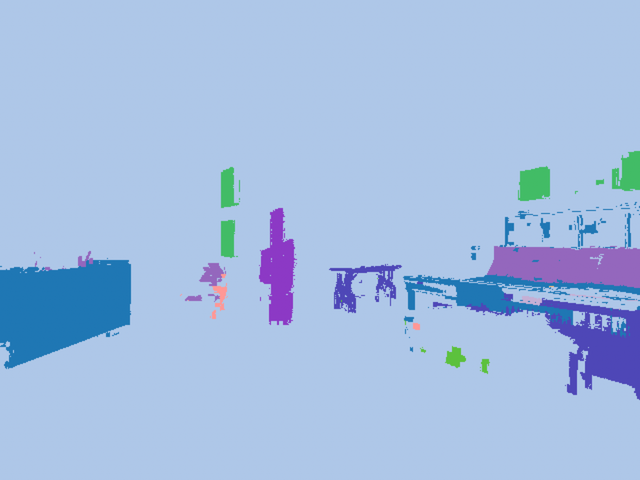} &
\includegraphics[width=0.155\linewidth]{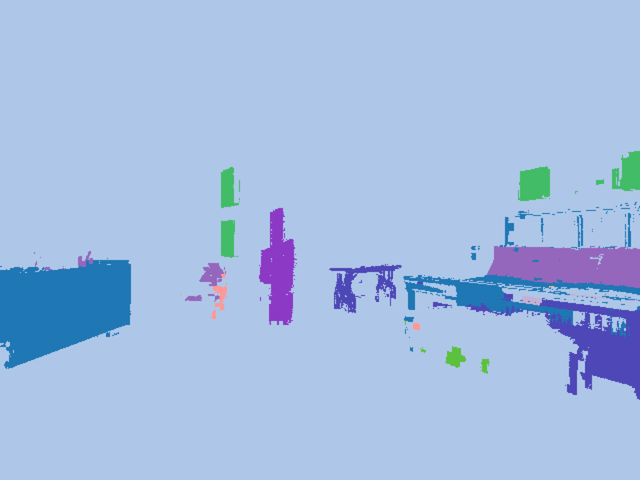} \\

\includegraphics[width=0.155\linewidth]{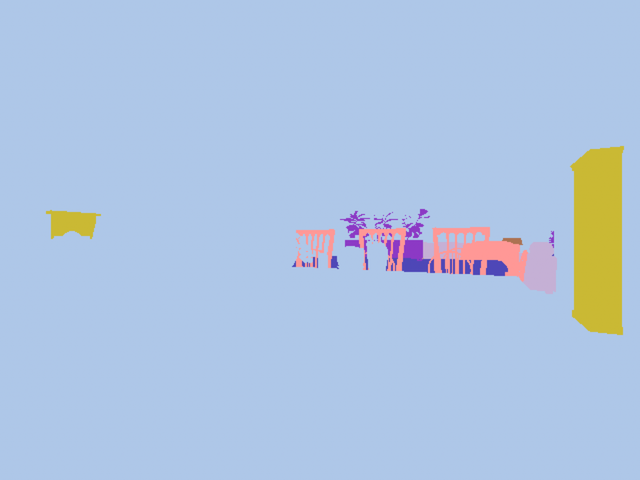} &
\includegraphics[width=0.155\linewidth]{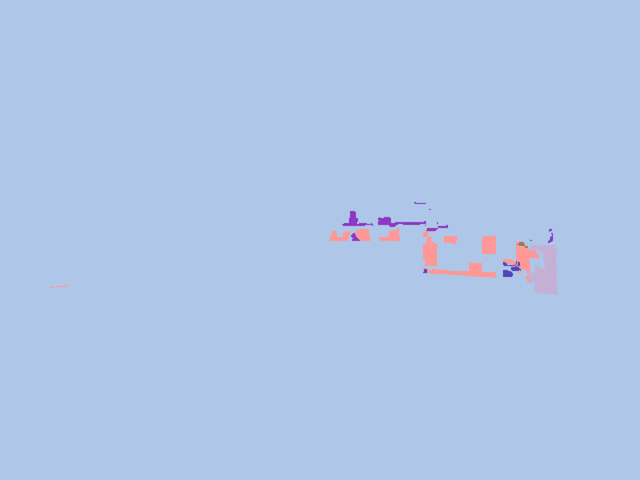} &
\includegraphics[width=0.155\linewidth]{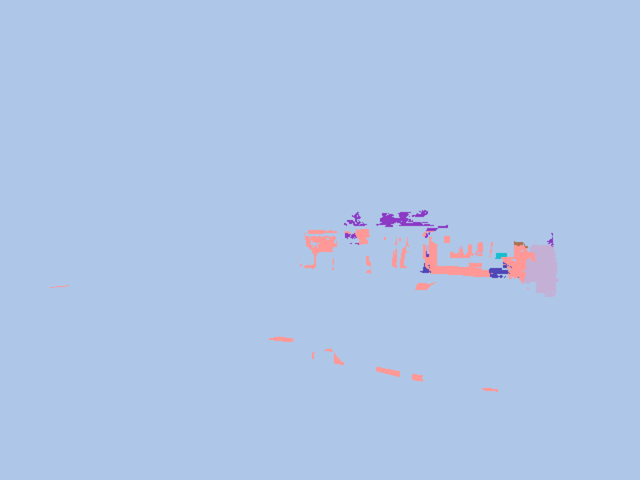} &
\includegraphics[width=0.155\linewidth]{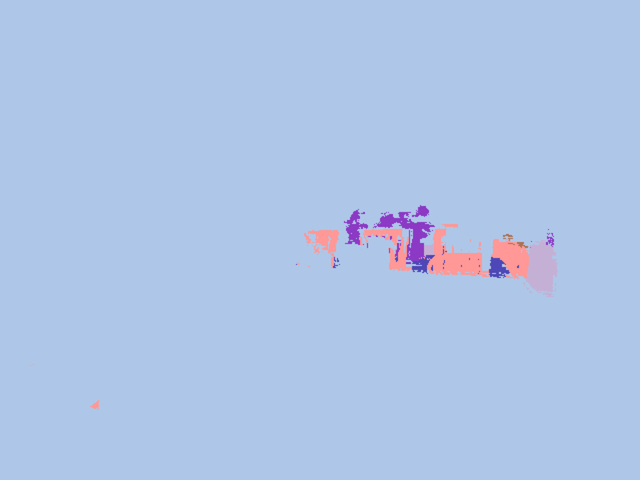} &
\includegraphics[width=0.155\linewidth]{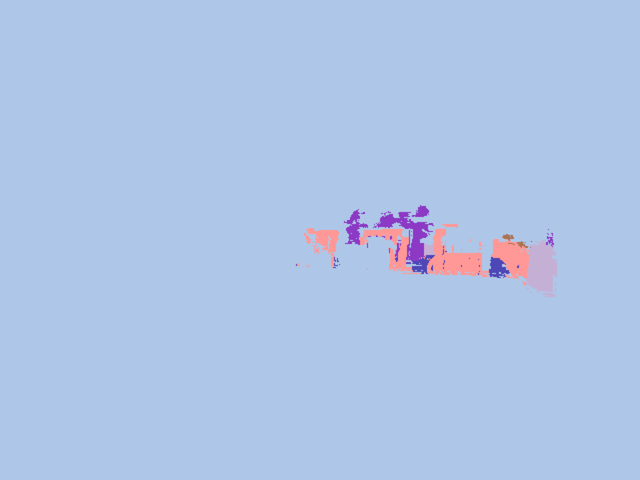} &
\includegraphics[width=0.155\linewidth]{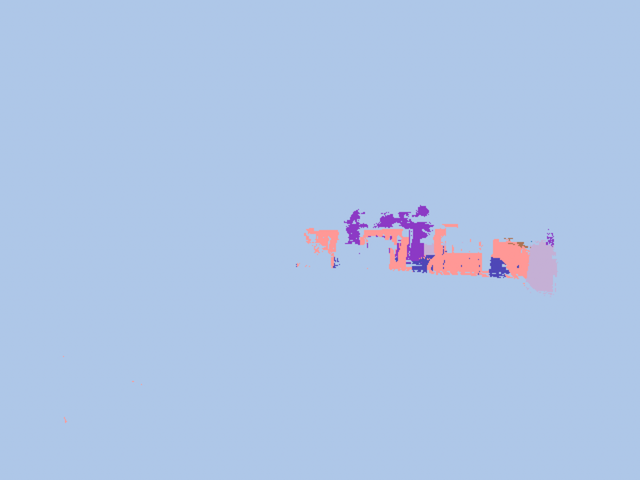} &
\includegraphics[width=0.155\linewidth]{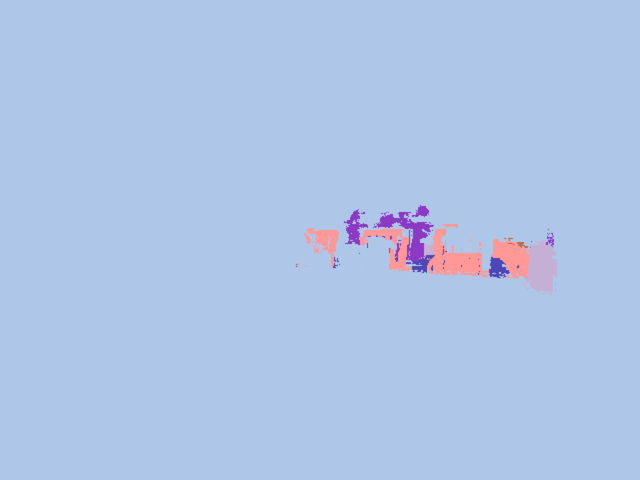} \\

\includegraphics[width=0.155\linewidth]{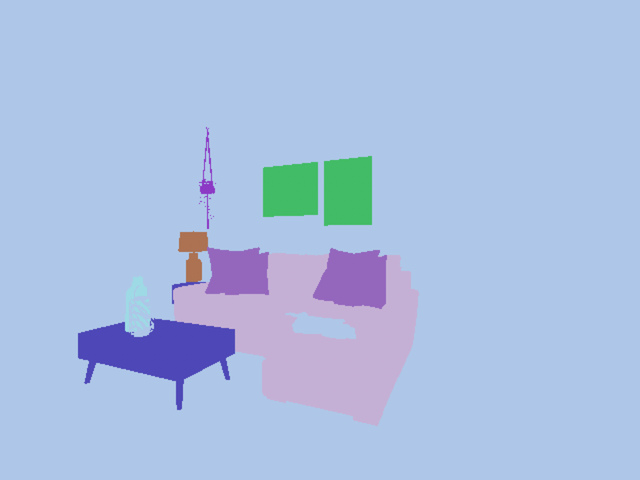} &
\includegraphics[width=0.155\linewidth]{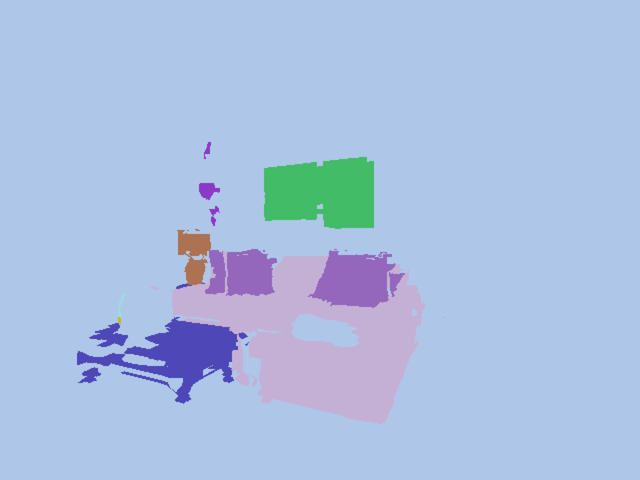} &
\includegraphics[width=0.155\linewidth]{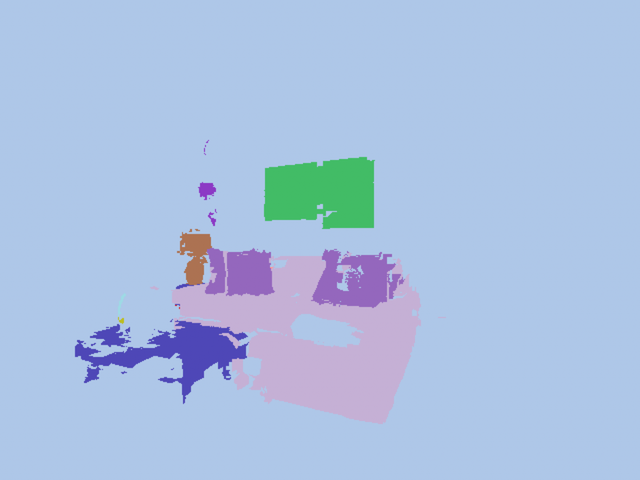} &
\includegraphics[width=0.155\linewidth]{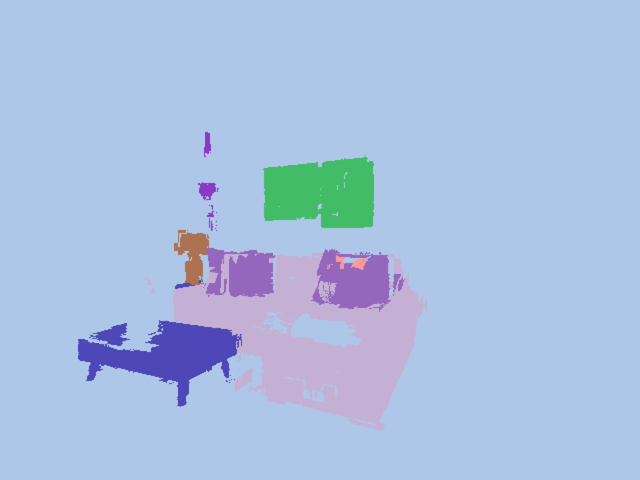} &
\includegraphics[width=0.155\linewidth]{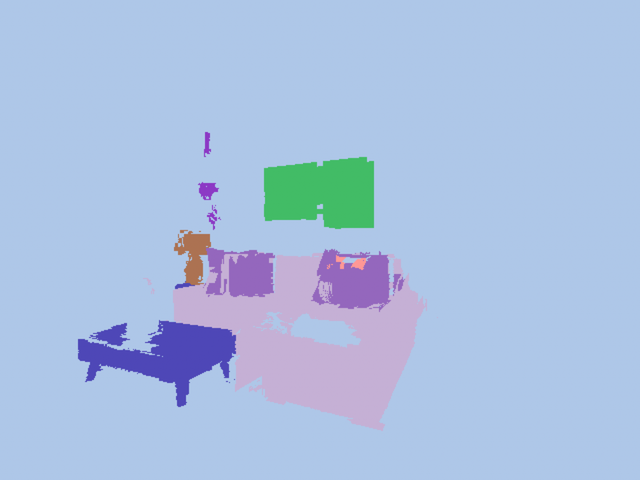} &
\includegraphics[width=0.155\linewidth]{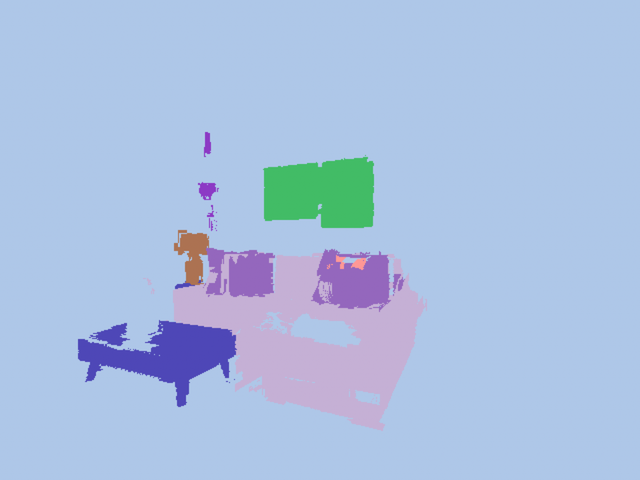} &
\includegraphics[width=0.155\linewidth]{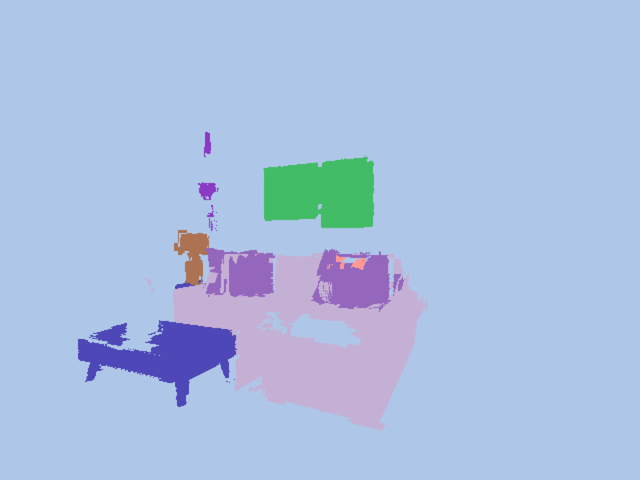} \\

\includegraphics[width=0.155\linewidth]{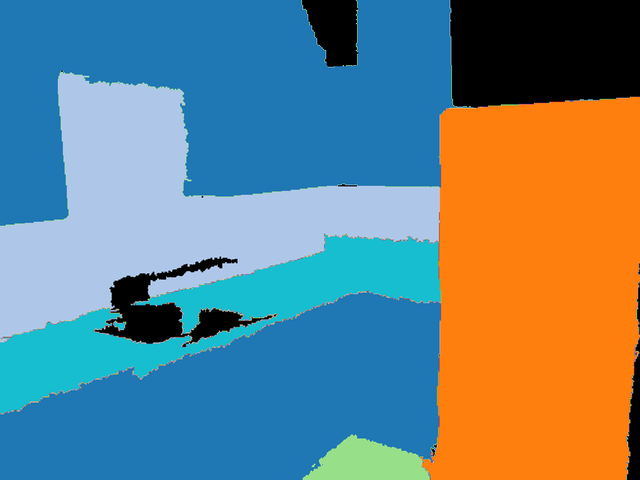} &
\includegraphics[width=0.155\linewidth]{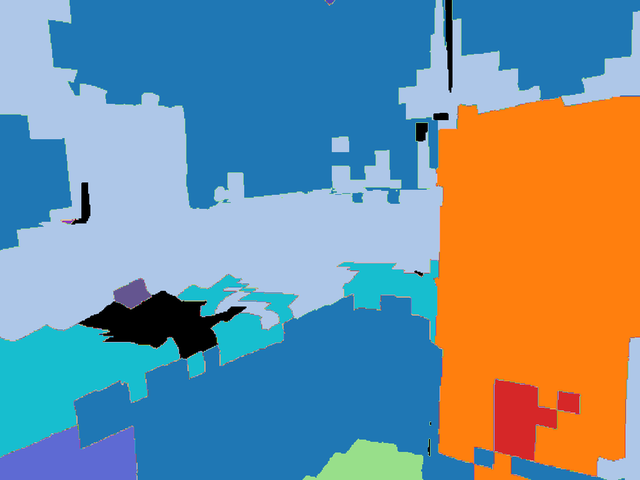} &
\includegraphics[width=0.155\linewidth]{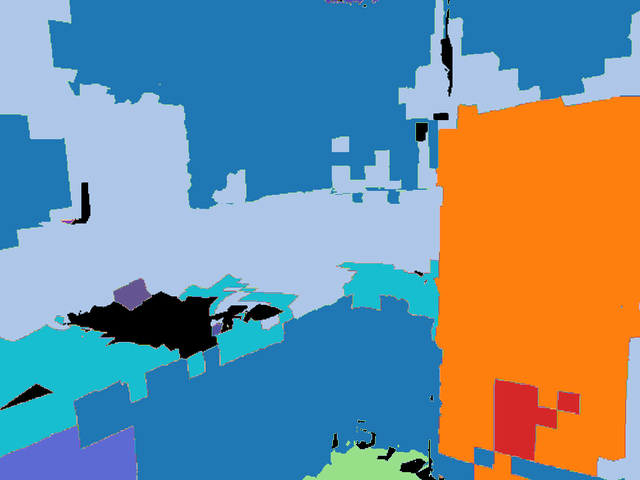} &
\includegraphics[width=0.155\linewidth]{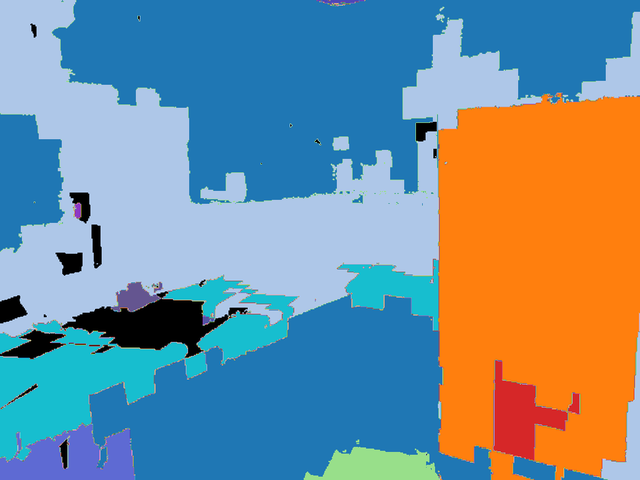} &
\includegraphics[width=0.155\linewidth]{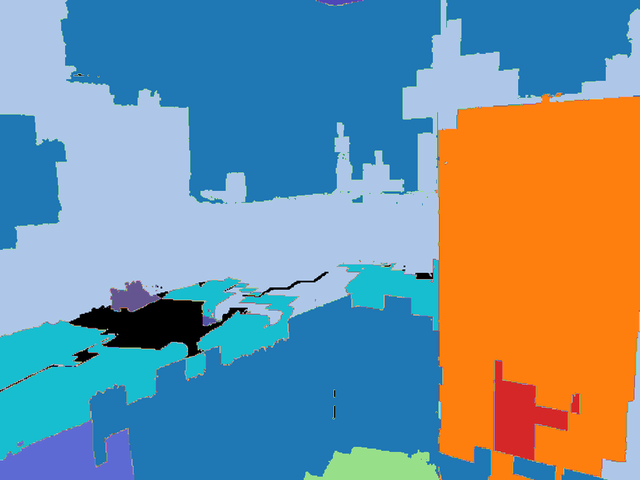} &
\includegraphics[width=0.155\linewidth]{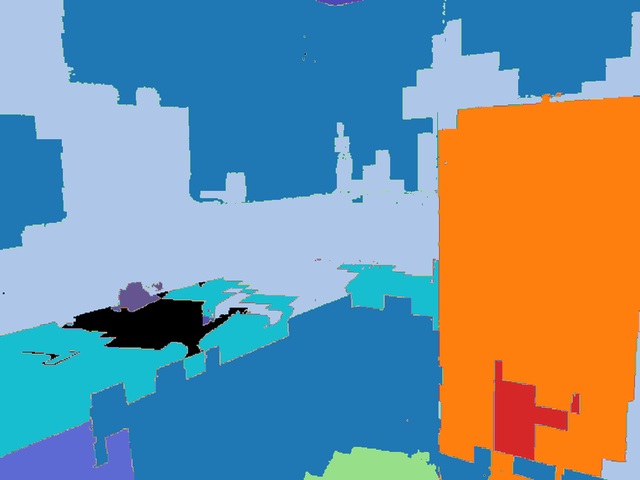} &
\includegraphics[width=0.155\linewidth]{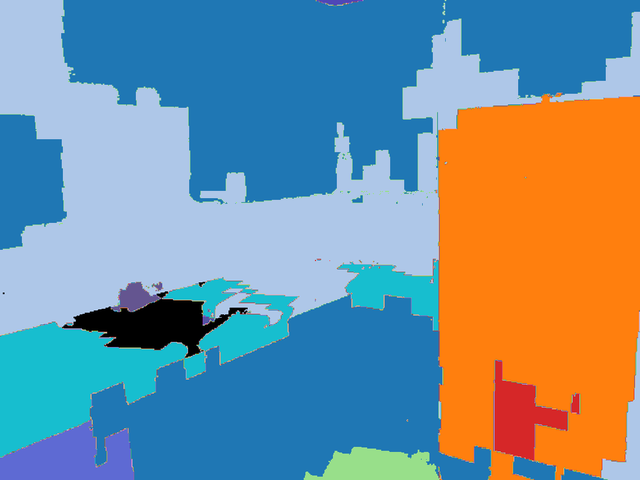} \\

\includegraphics[width=0.155\linewidth]{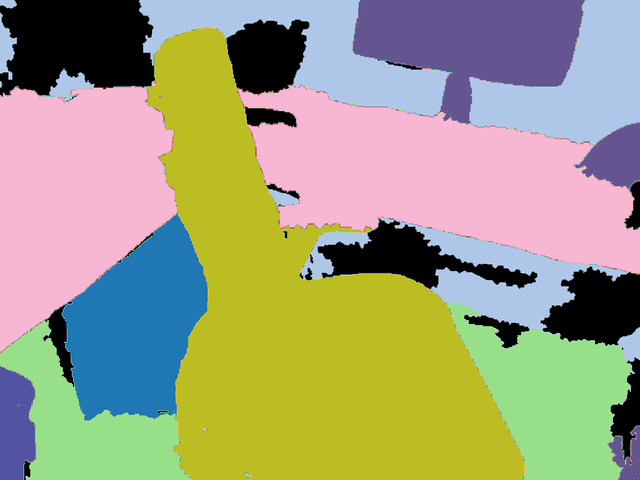} &
\includegraphics[width=0.155\linewidth]{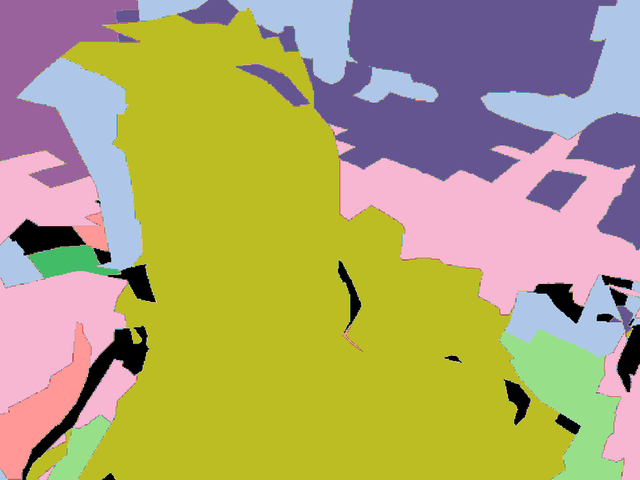} &
\includegraphics[width=0.155\linewidth]{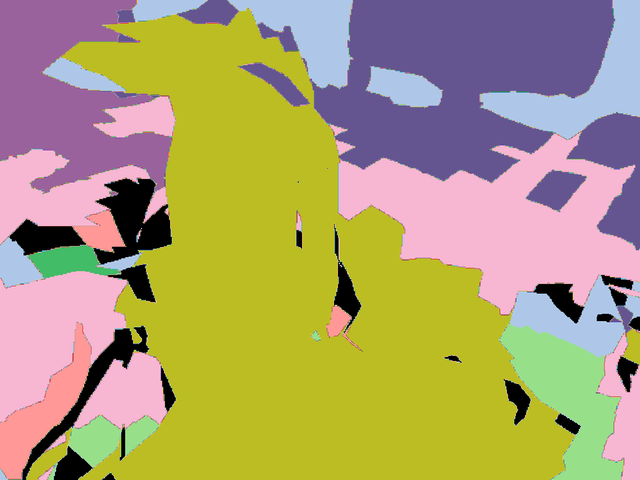} &
\includegraphics[width=0.155\linewidth]{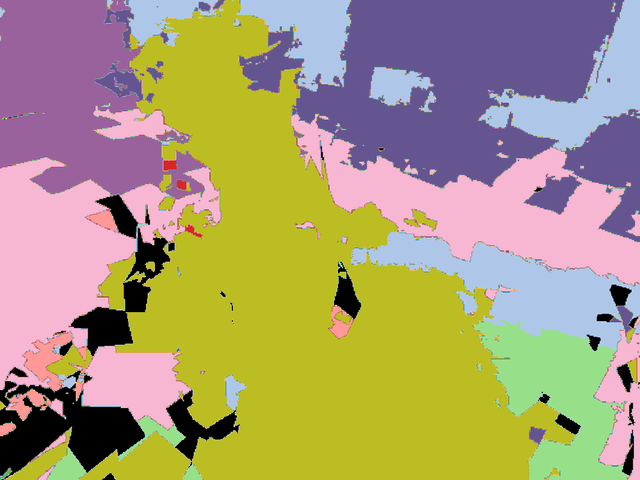} &
\includegraphics[width=0.155\linewidth]{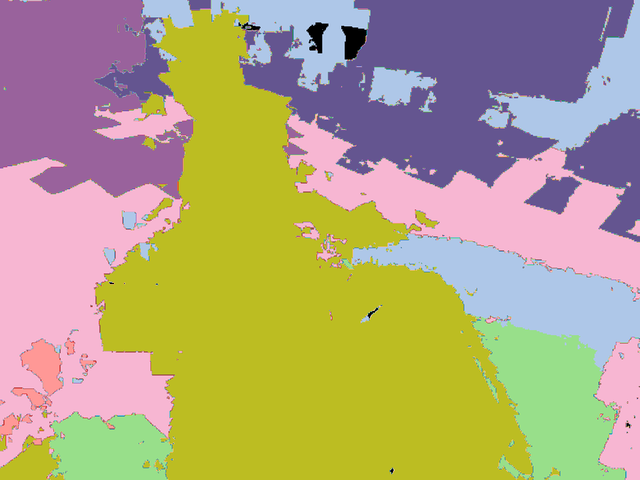} &
\includegraphics[width=0.155\linewidth]{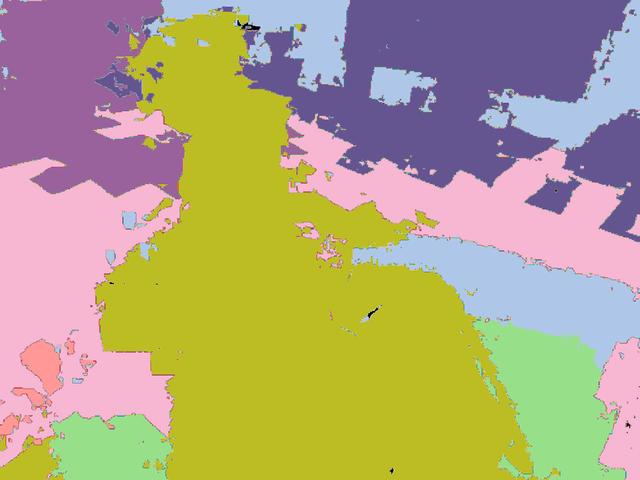} &
\includegraphics[width=0.155\linewidth]{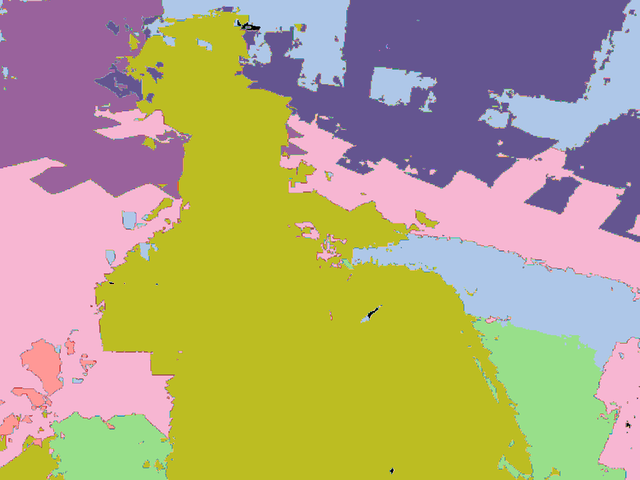} \\

\includegraphics[width=0.155\linewidth]{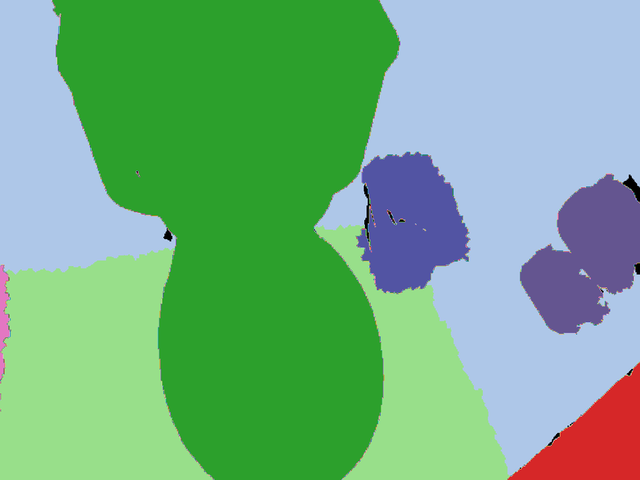} &
\includegraphics[width=0.155\linewidth]{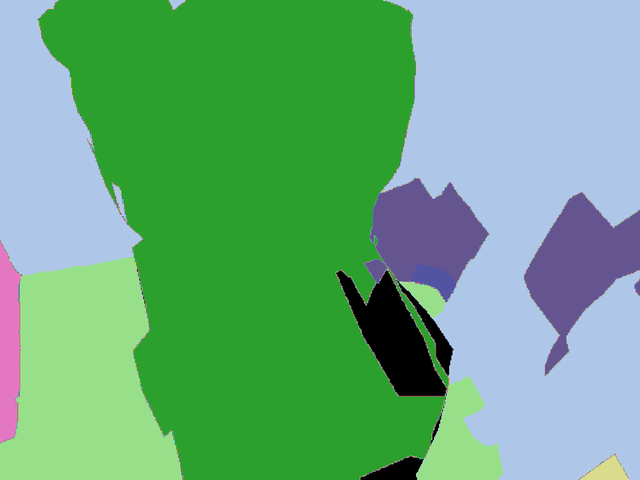} &
\includegraphics[width=0.155\linewidth]{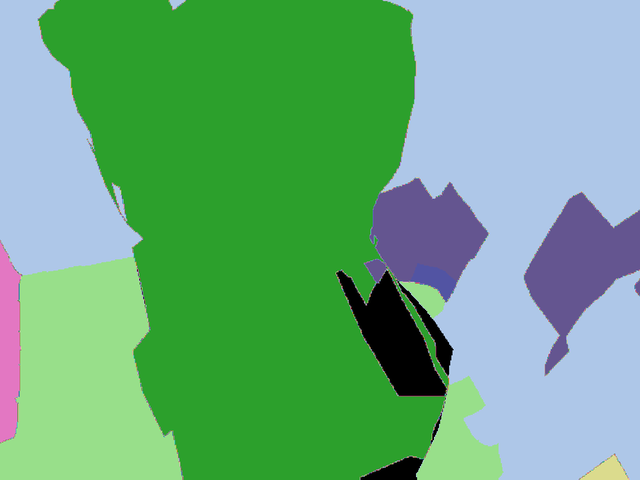} &
\includegraphics[width=0.155\linewidth]{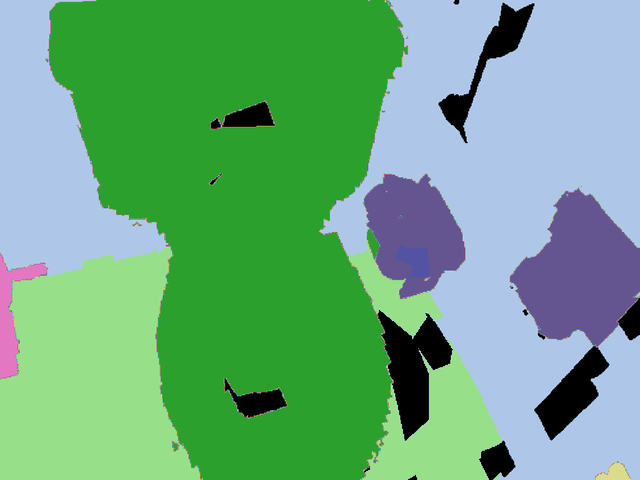} &
\includegraphics[width=0.155\linewidth]{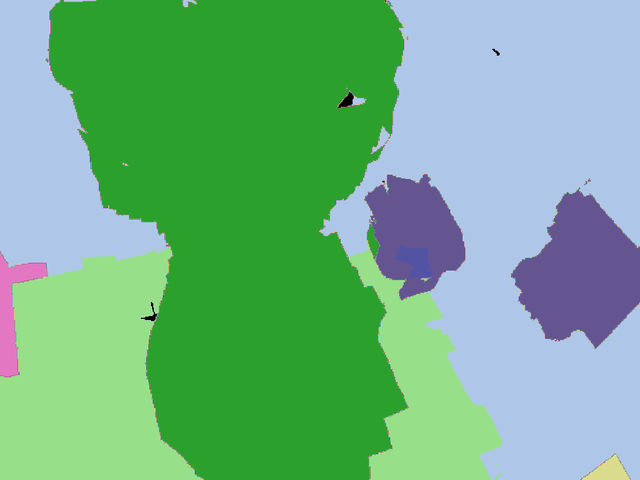} &
\includegraphics[width=0.155\linewidth]{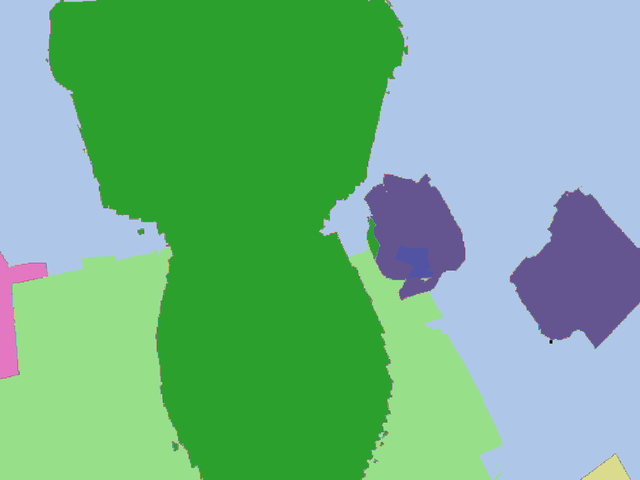} &
\includegraphics[width=0.155\linewidth]{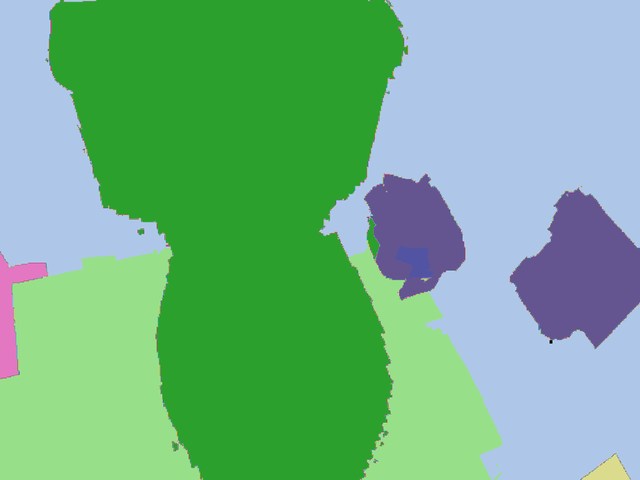} \\

\end{tabular}}
\caption{
\textbf{Semantic segmentations} rendered from the reconstructed scene meshes on HSSD (top three rows) and ScanNet (bottom) using RefineNet~\cite{lin2017refinenet} to provide per-frame semantic segmentations.
Columns show increasing reconstruction quality and detail from coarse \textsc{Voxblox} (8\,cm) to the fine-grained 1\,cm baseline.
Our deterministic entropy-based method (\textsc{Un-Map-Adapt}) and our budget-aware RL model (\textsc{UnRL}) recover sharper geometry, cleaner object boundaries, and more stable textures than \textsc{Map-Adapt}, without requiring a hand-crafted semantic class list.
The RL variants (25\%, 75\%) adaptively preserve high-uncertainty regions while respecting memory budgets, demonstrating controlled degradation under tight constraints.
}
\label{fig:qualitative_summary_sem}
\end{figure}

\begin{figure}[t]
\centering
\setlength{\tabcolsep}{1pt}
\renewcommand{\arraystretch}{0.75}

\resizebox{\linewidth}{!}{%
\begin{tabular}{ccccccccc}
\scriptsize \textsc{Ground Truth} &
\scriptsize \textsc{Voxblox} (8cm) &
\scriptsize \ablUncertaintyGeoTextureRL (25\%) &
\scriptsize \ablUncertaintyGeoTextureRL (50\%) &
\scriptsize \ablUncertaintyGeoTextureRL (75\%) &
\scriptsize \ablMapadaptGeo &
\scriptsize \ablUncertaintyGeoTexture &
\scriptsize \textsc{Voxblox} (4cm) &
\scriptsize \textsc{Voxblox} (1cm) \\

\includegraphics[width=0.155\linewidth]{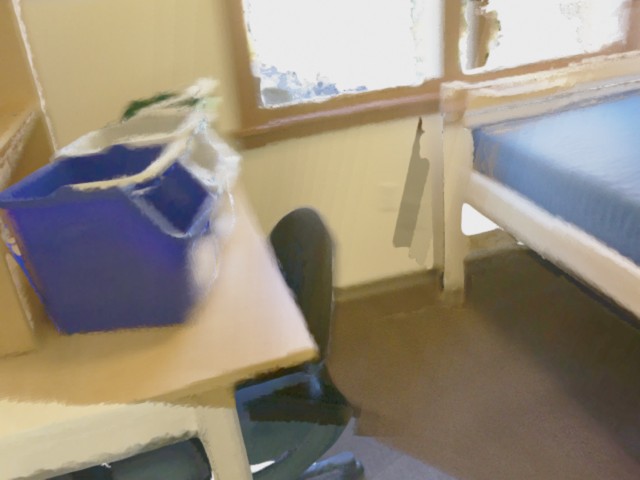} &
\includegraphics[width=0.155\linewidth]{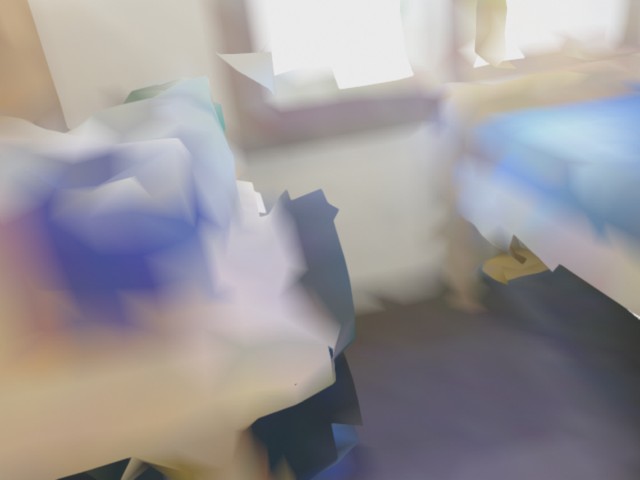} &
\includegraphics[width=0.155\linewidth]{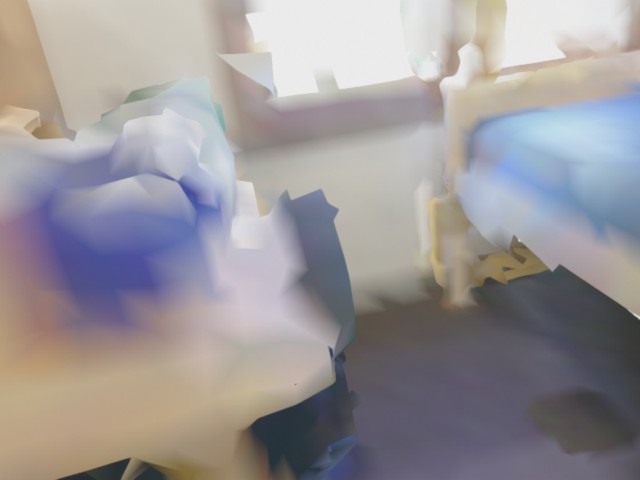} &
\includegraphics[width=0.155\linewidth]{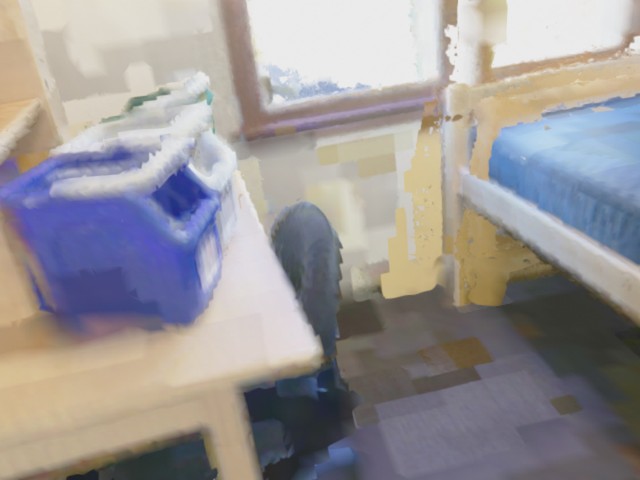} &
\includegraphics[width=0.155\linewidth]{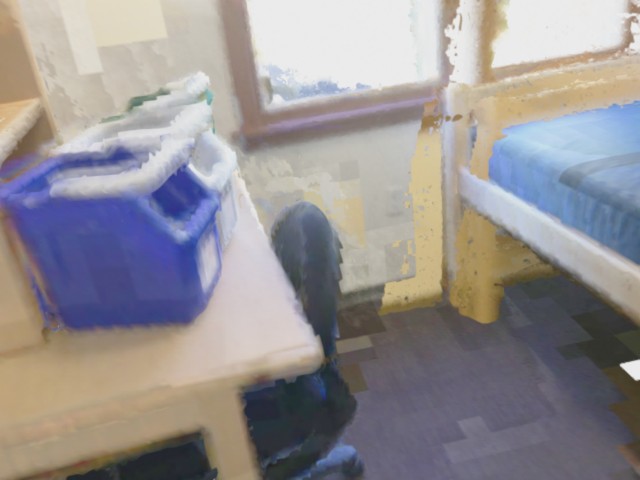} &
\includegraphics[width=0.155\linewidth]{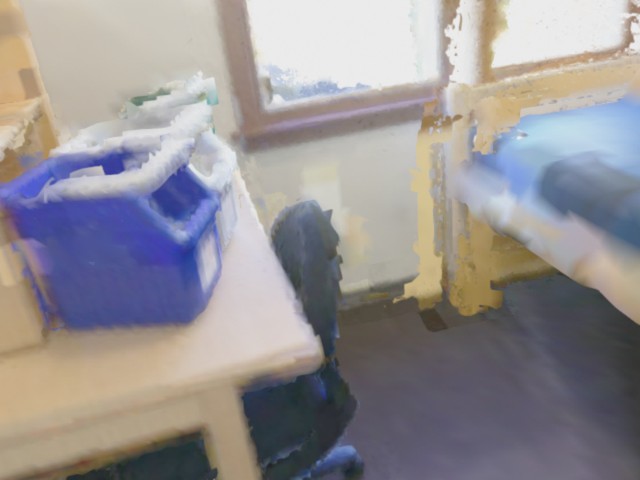} &
\includegraphics[width=0.155\linewidth]{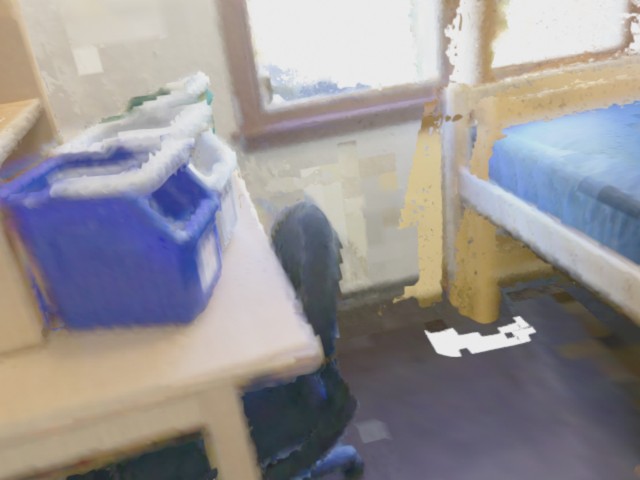} &
\includegraphics[width=0.155\linewidth]{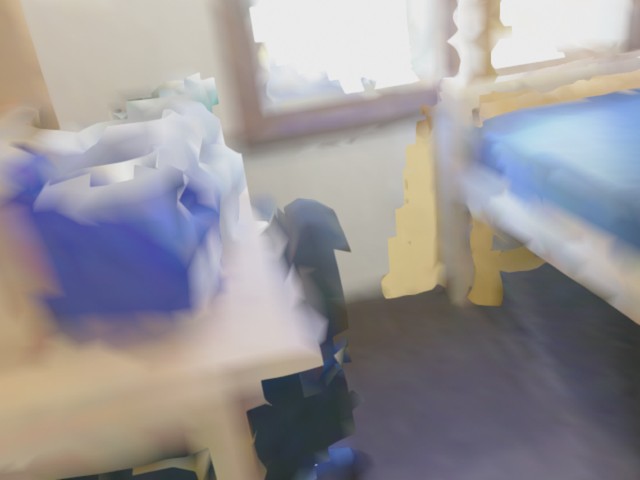} &
\includegraphics[width=0.155\linewidth]{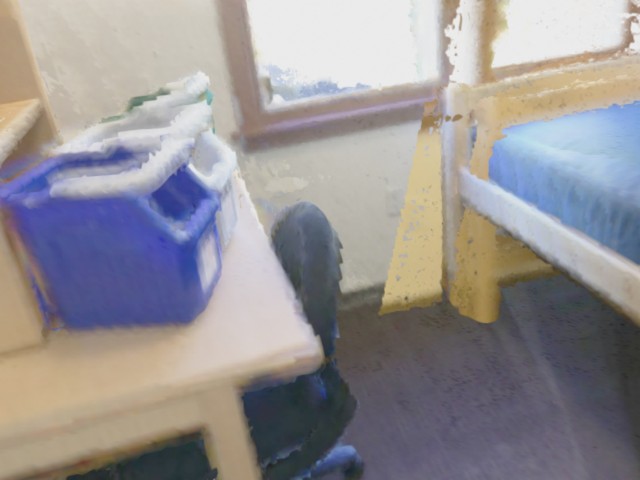} \\

\includegraphics[width=0.155\linewidth]{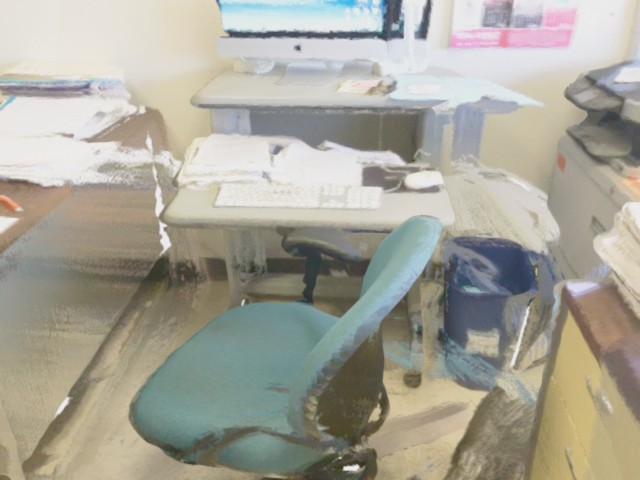} &
\includegraphics[width=0.155\linewidth]{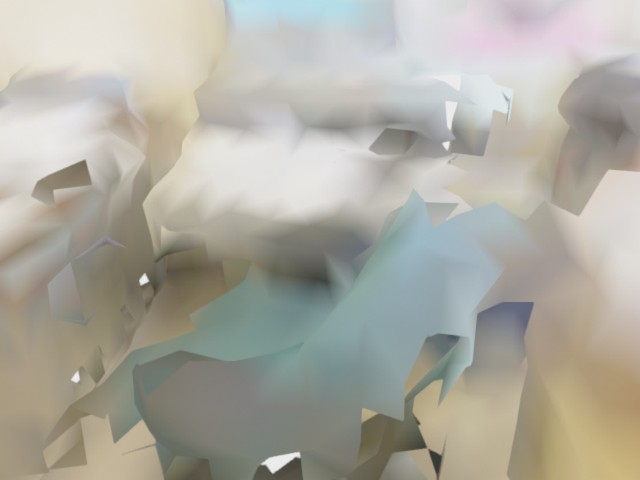} &
\includegraphics[width=0.155\linewidth]{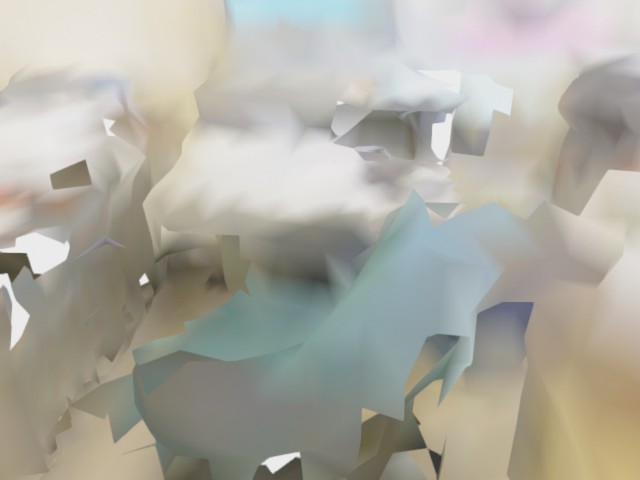} &
\includegraphics[width=0.155\linewidth]{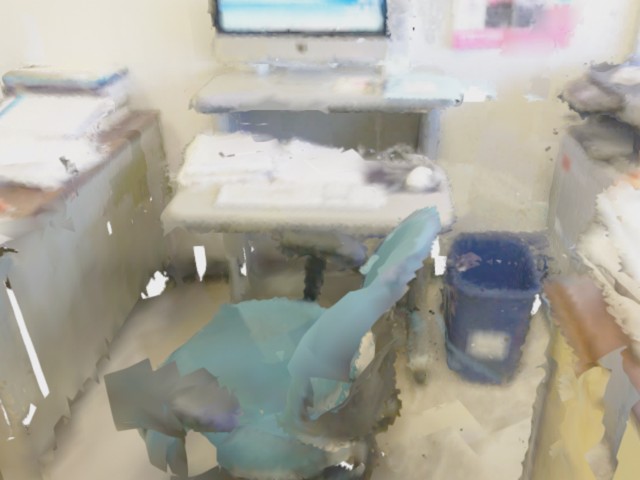} &
\includegraphics[width=0.155\linewidth]{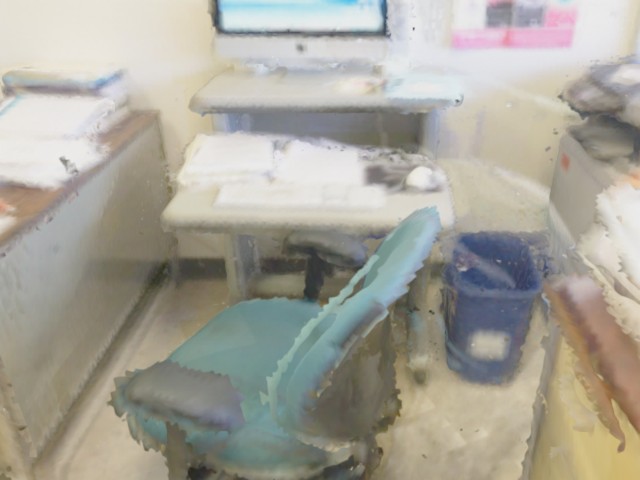} &
\includegraphics[width=0.155\linewidth]{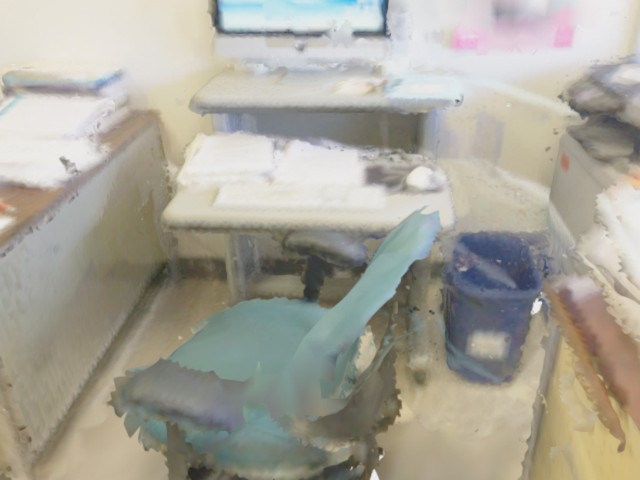} &
\includegraphics[width=0.155\linewidth]{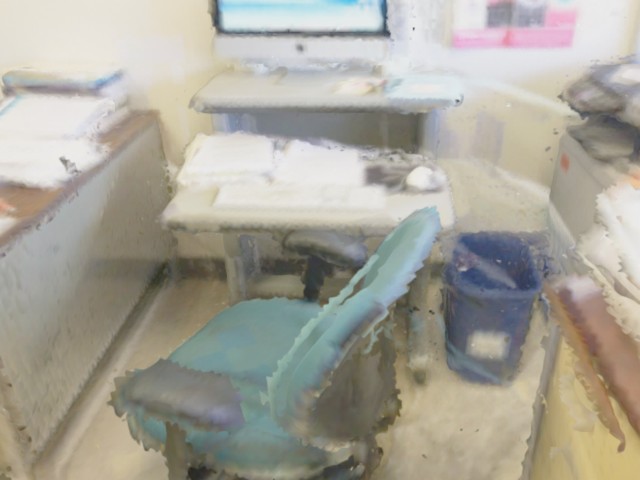} &
\includegraphics[width=0.155\linewidth]{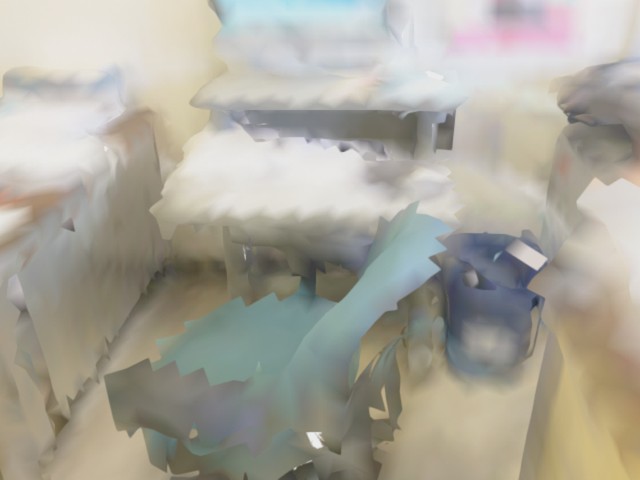} &
\includegraphics[width=0.155\linewidth]{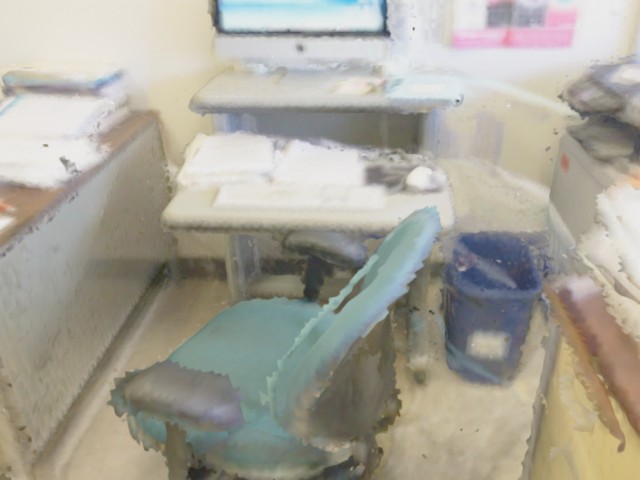} \\

\includegraphics[width=0.155\linewidth]{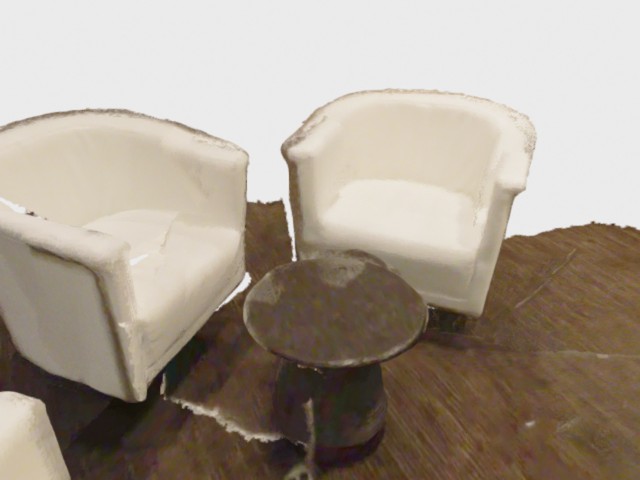} &
\includegraphics[width=0.155\linewidth]{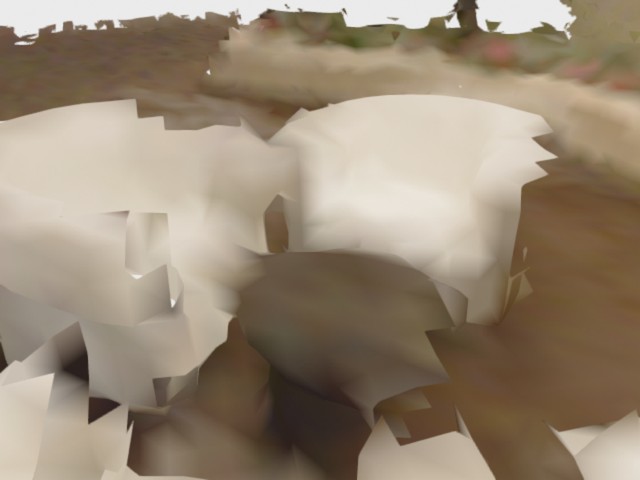} &
\includegraphics[width=0.155\linewidth]{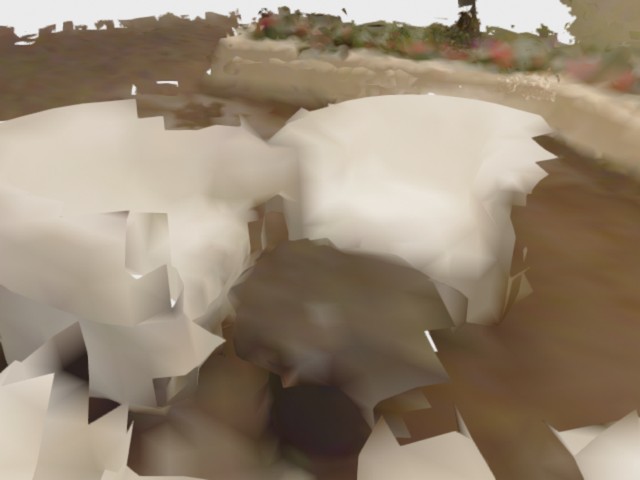} &
\includegraphics[width=0.155\linewidth]{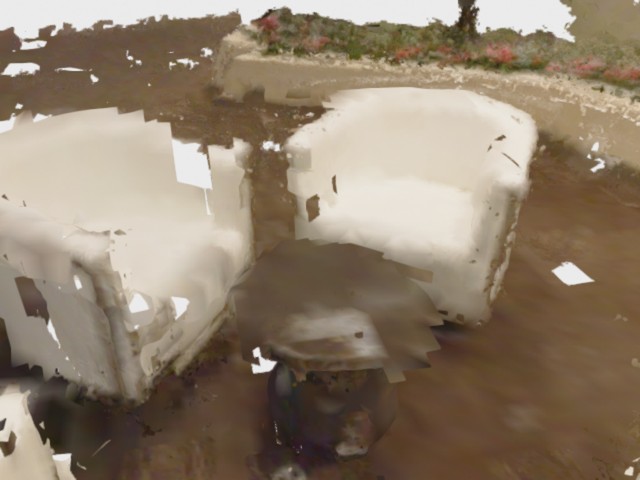} &
\includegraphics[width=0.155\linewidth]{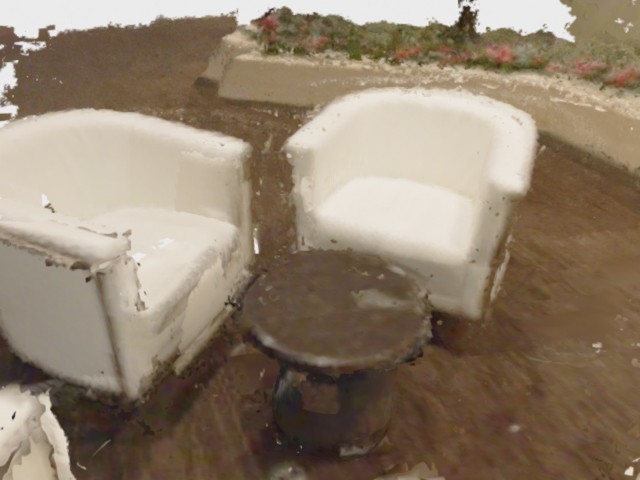} &
\includegraphics[width=0.155\linewidth]{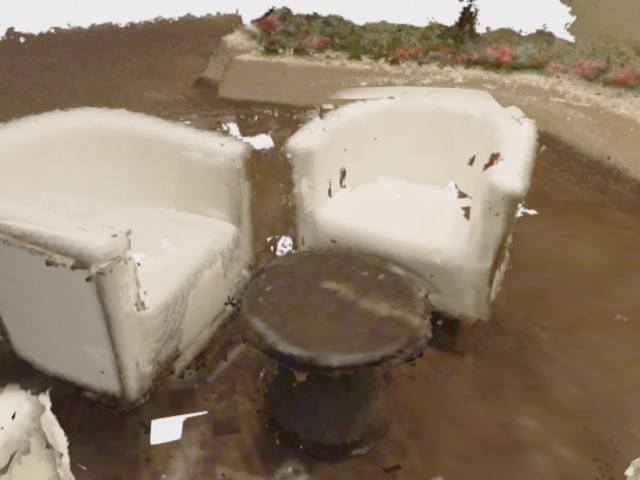} &
\includegraphics[width=0.155\linewidth]{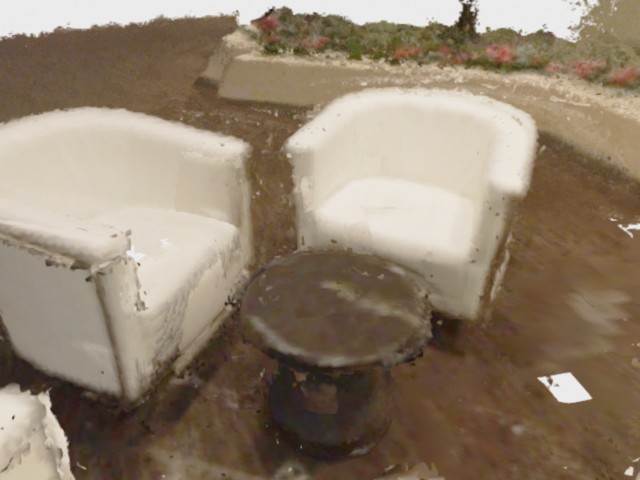} &
\includegraphics[width=0.155\linewidth]{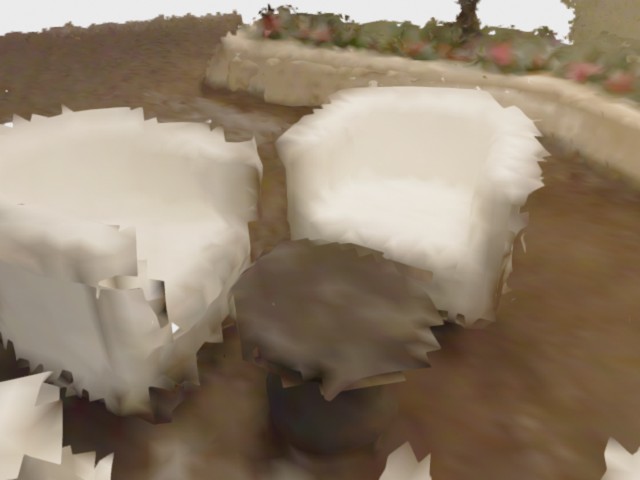} &
\includegraphics[width=0.155\linewidth]{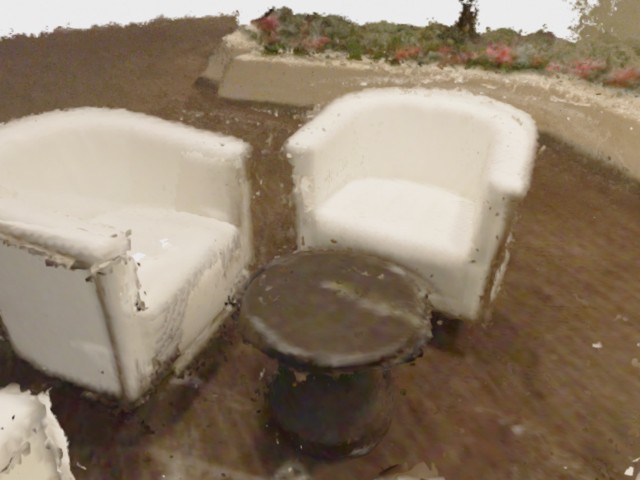} \\

\includegraphics[width=0.155\linewidth]{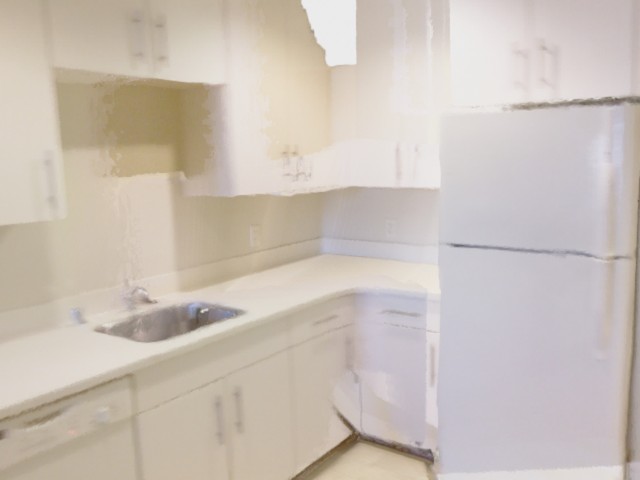} &
\includegraphics[width=0.155\linewidth]{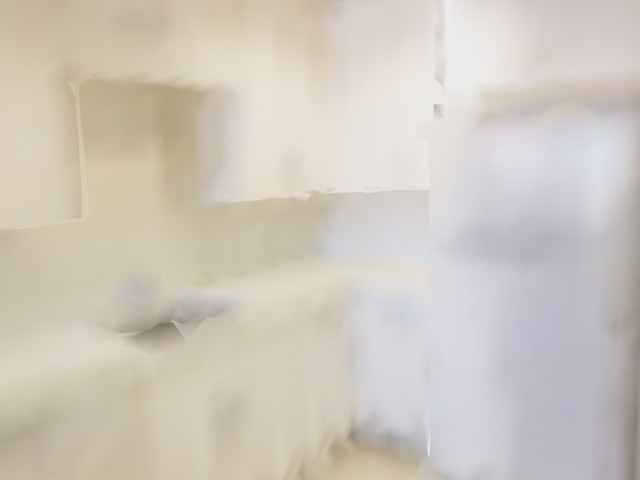} &
\includegraphics[width=0.155\linewidth]{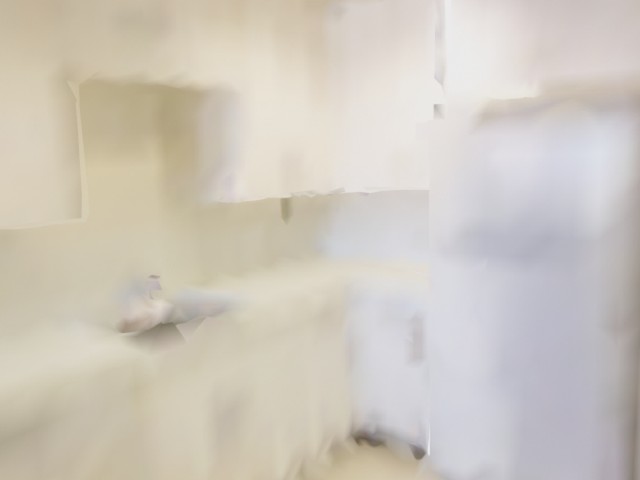} &
\includegraphics[width=0.155\linewidth]{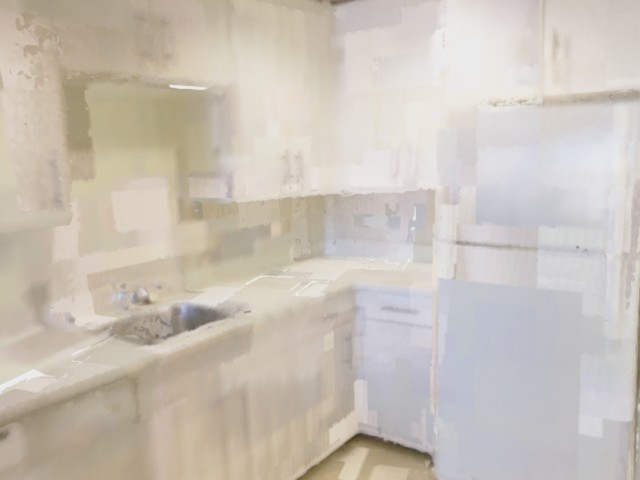} &
\includegraphics[width=0.155\linewidth]{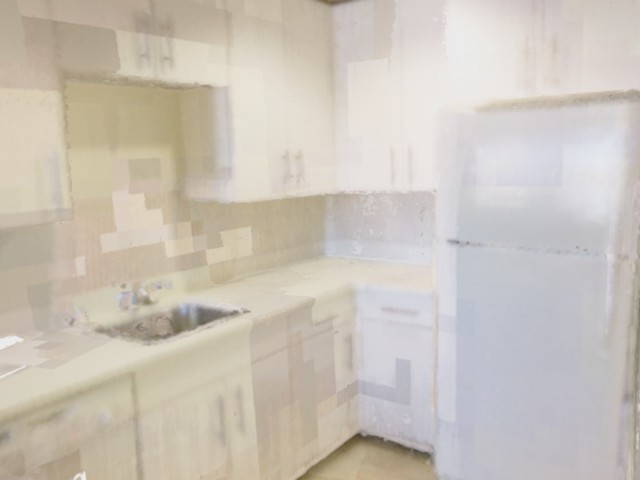} &
\includegraphics[width=0.155\linewidth]{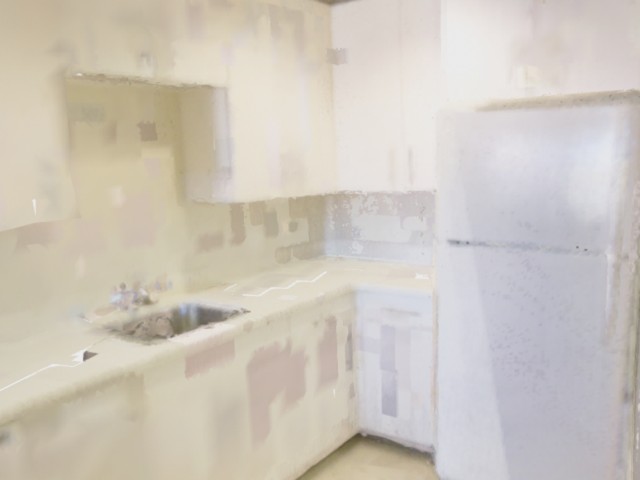} &
\includegraphics[width=0.155\linewidth]{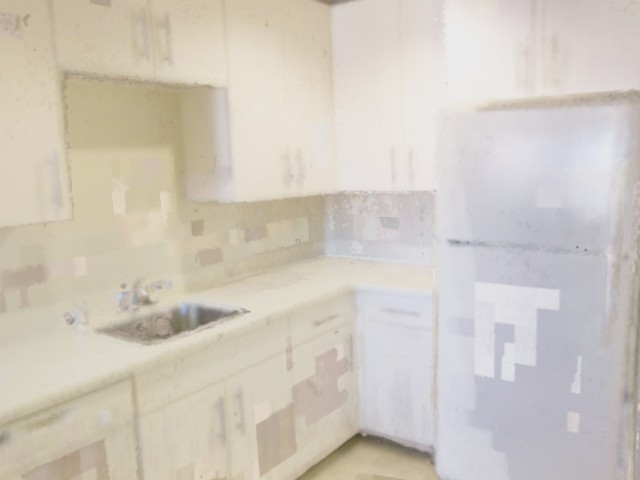} &
\includegraphics[width=0.155\linewidth]{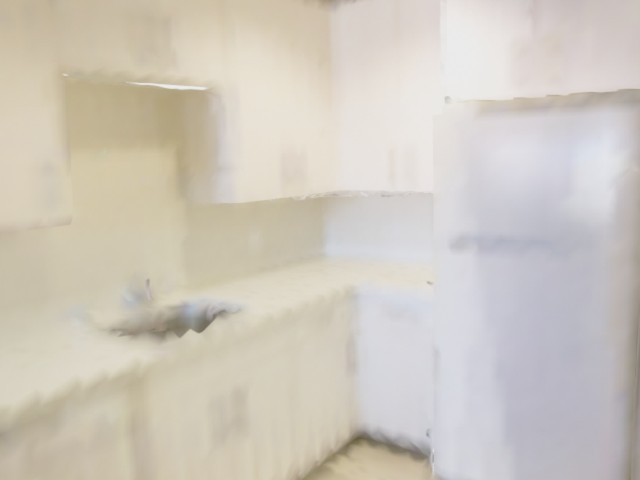} &
\includegraphics[width=0.155\linewidth]{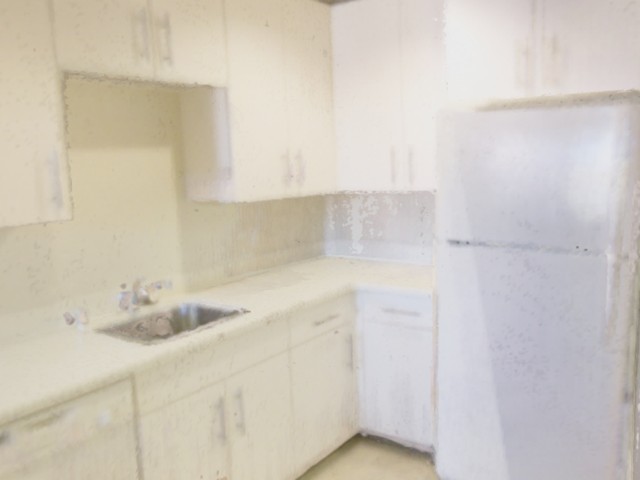} \\

\includegraphics[width=0.155\linewidth]{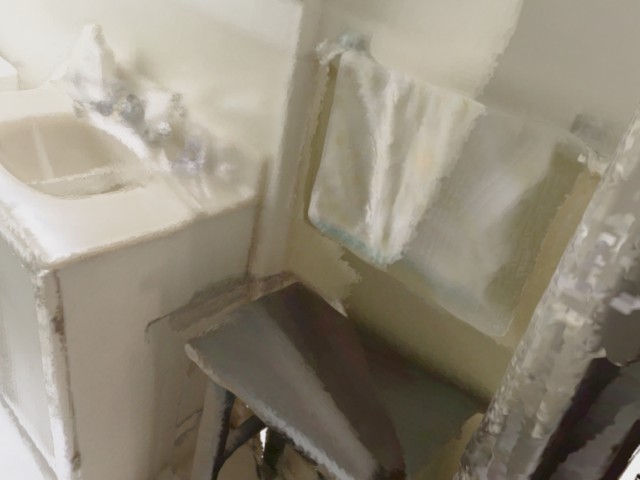} &
\includegraphics[width=0.155\linewidth]{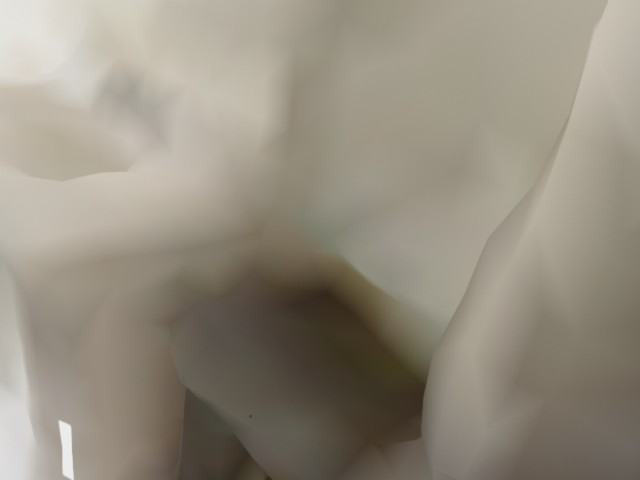} &
\includegraphics[width=0.155\linewidth]{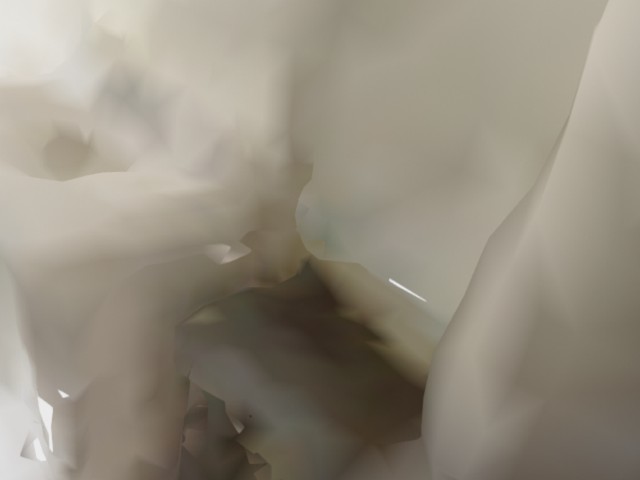} &
\includegraphics[width=0.155\linewidth]{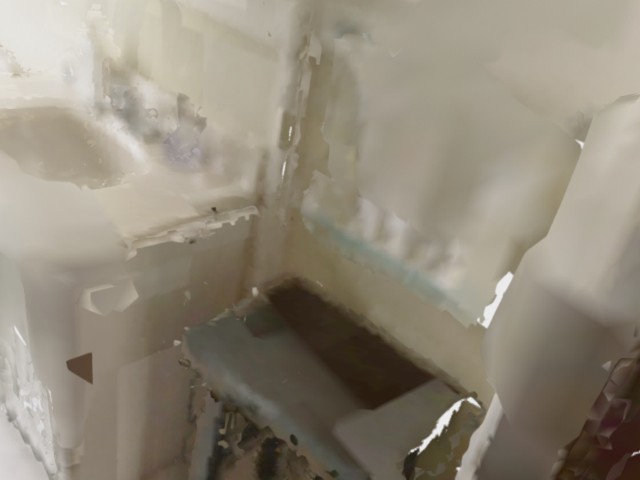} &
\includegraphics[width=0.155\linewidth]{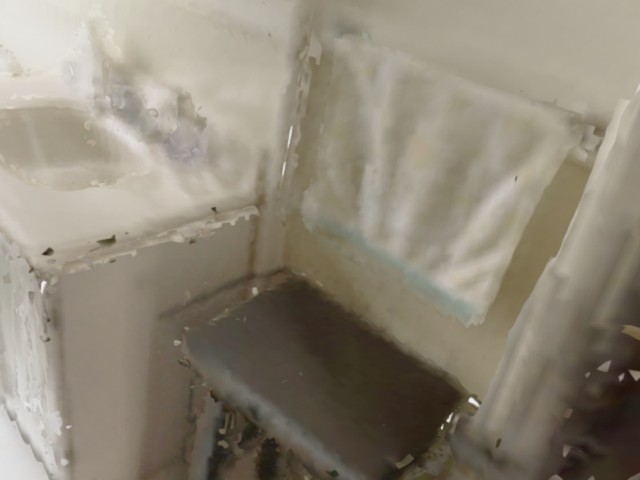} &
\includegraphics[width=0.155\linewidth]{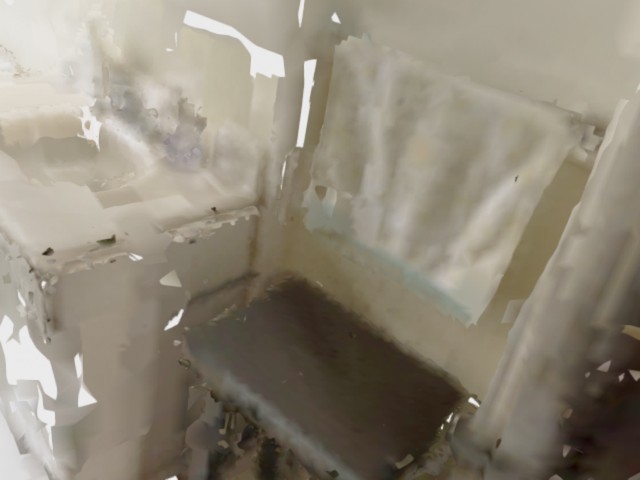} &
\includegraphics[width=0.155\linewidth]{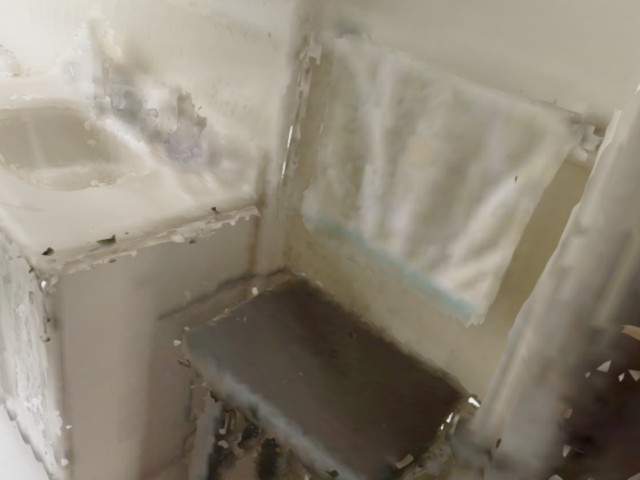} &
\includegraphics[width=0.155\linewidth]{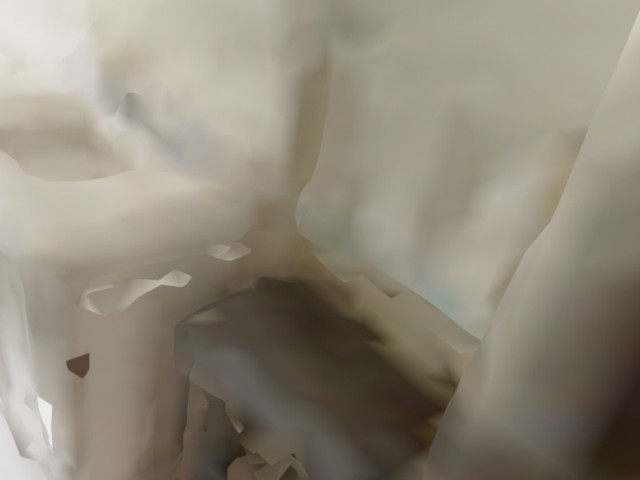} &
\includegraphics[width=0.155\linewidth]{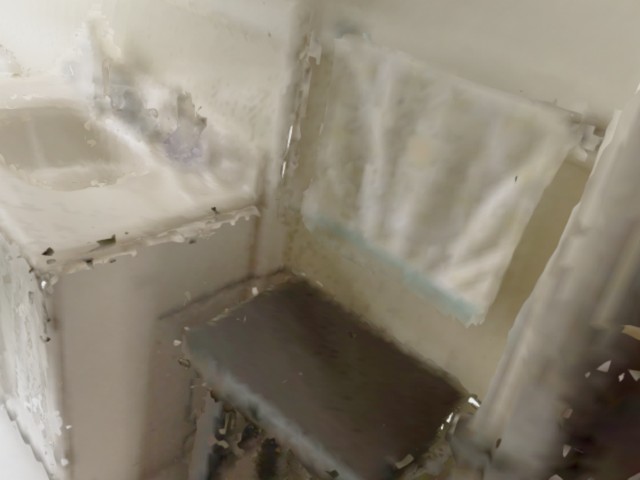} \\

\includegraphics[width=0.155\linewidth]{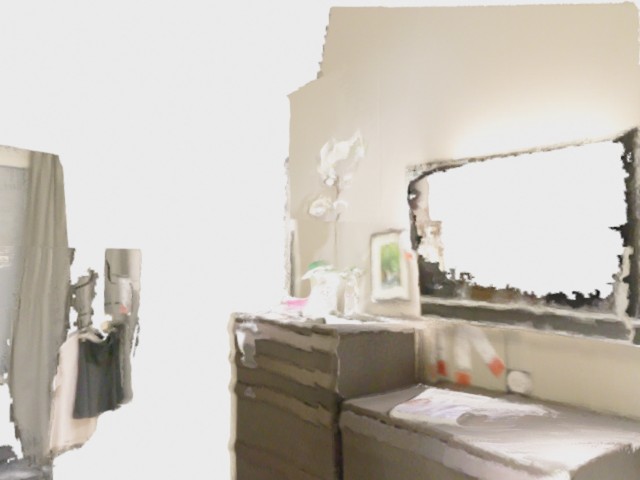} &
\includegraphics[width=0.155\linewidth]{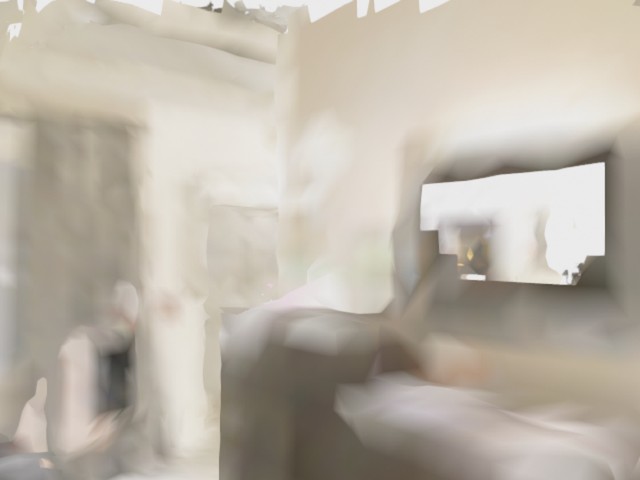} &
\includegraphics[width=0.155\linewidth]{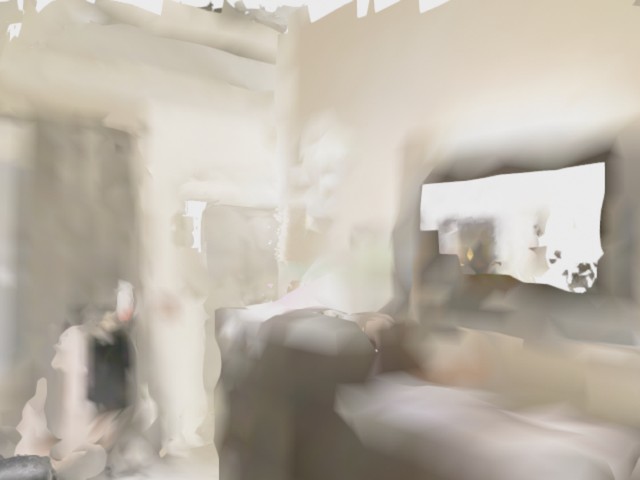} &
\includegraphics[width=0.155\linewidth]{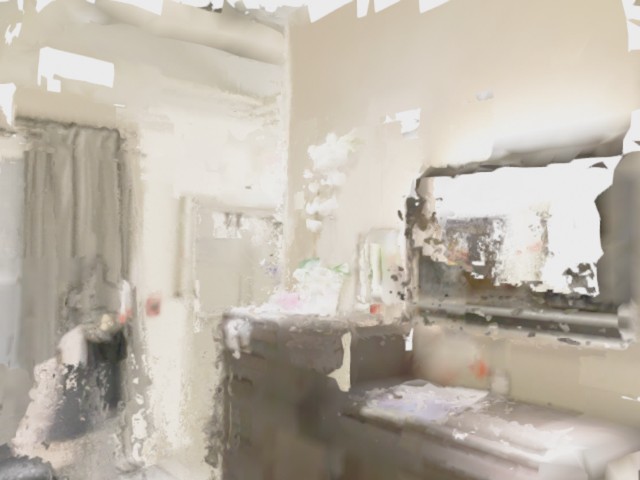} &
\includegraphics[width=0.155\linewidth]{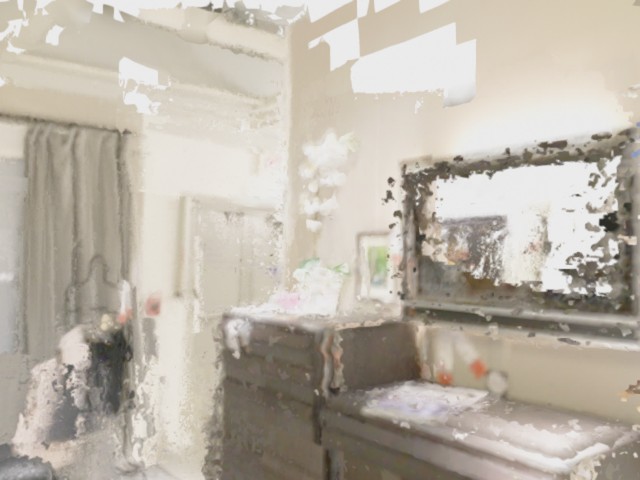} &
\includegraphics[width=0.155\linewidth]{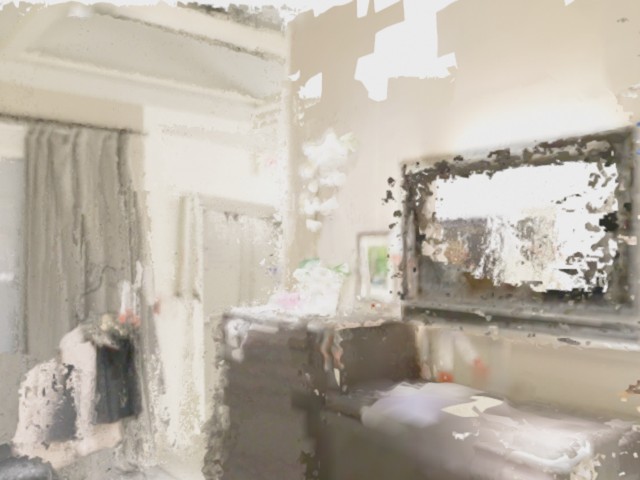} &
\includegraphics[width=0.155\linewidth]{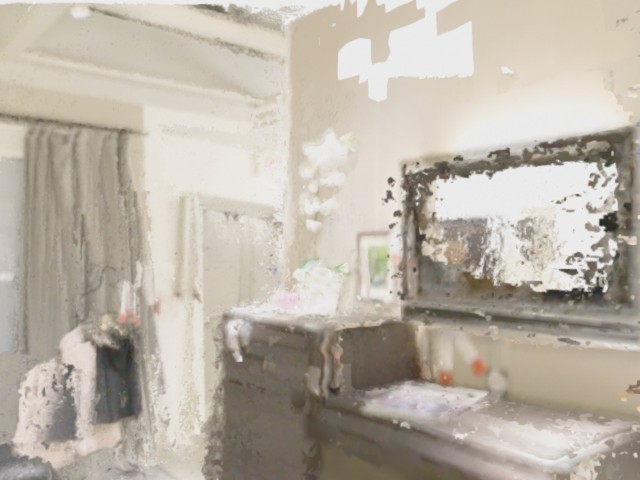} &
\includegraphics[width=0.155\linewidth]{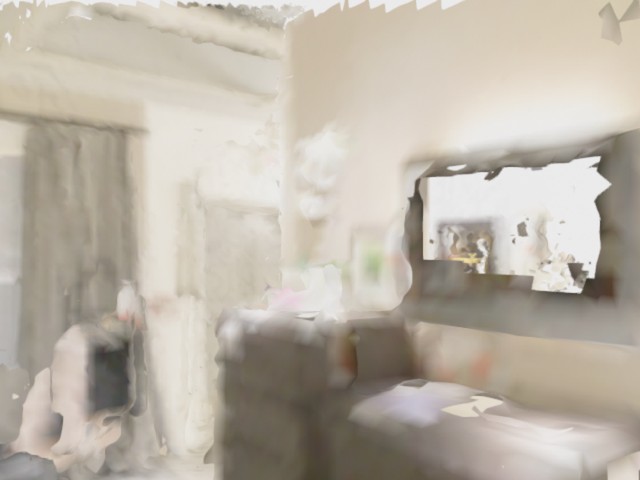} &
\includegraphics[width=0.155\linewidth]{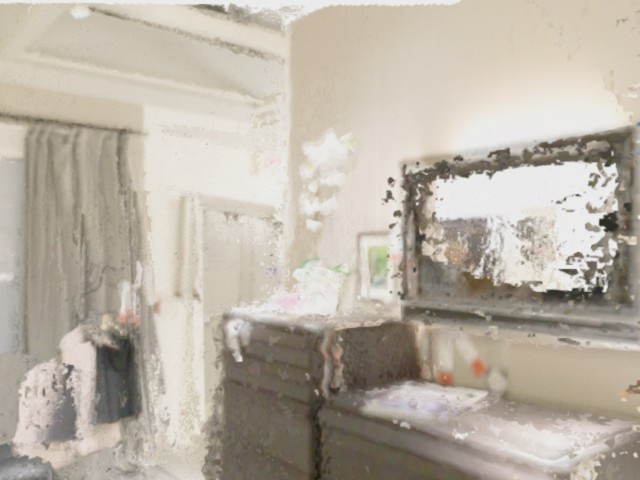} \\

\includegraphics[width=0.155\linewidth]{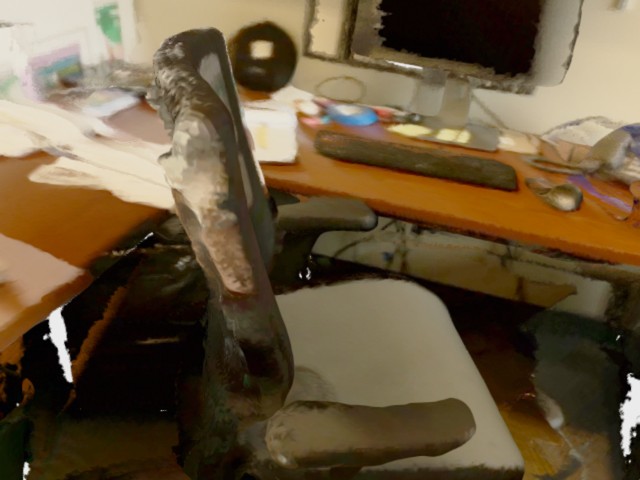} &
\includegraphics[width=0.155\linewidth]{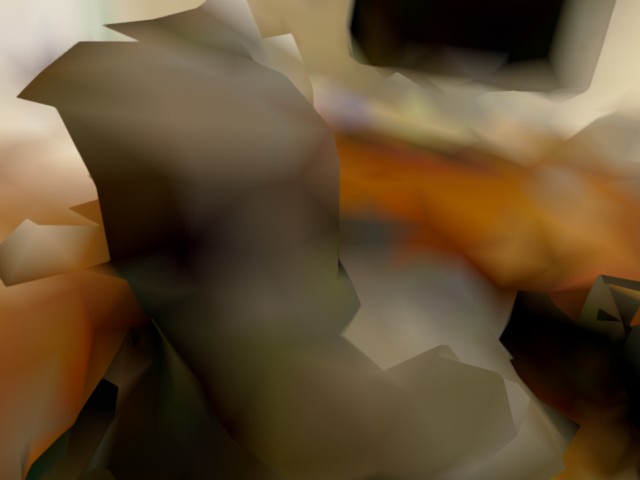} &
\includegraphics[width=0.155\linewidth]{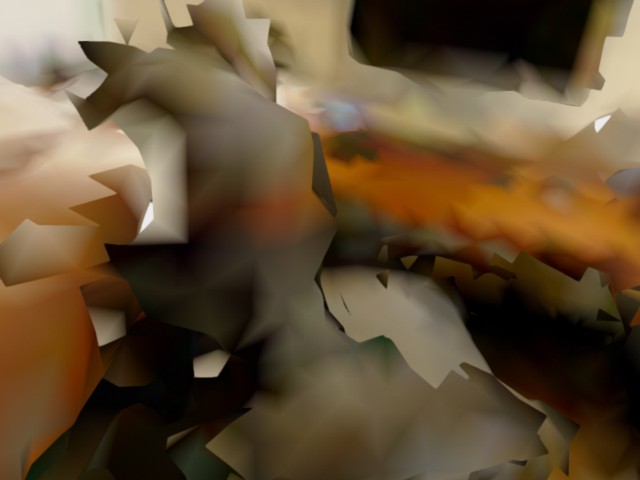} &
\includegraphics[width=0.155\linewidth]{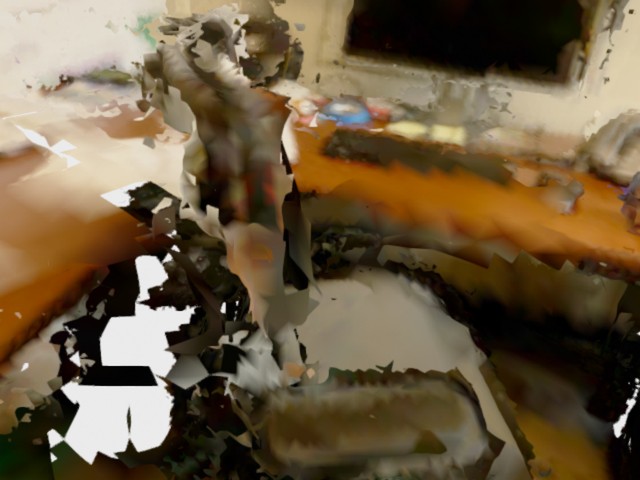} &
\includegraphics[width=0.155\linewidth]{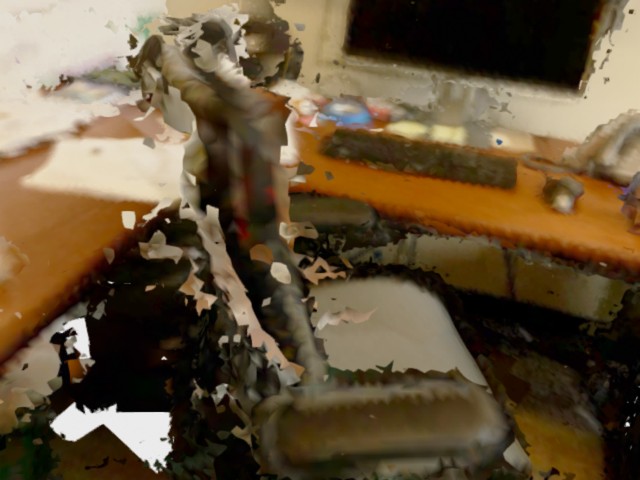} &
\includegraphics[width=0.155\linewidth]{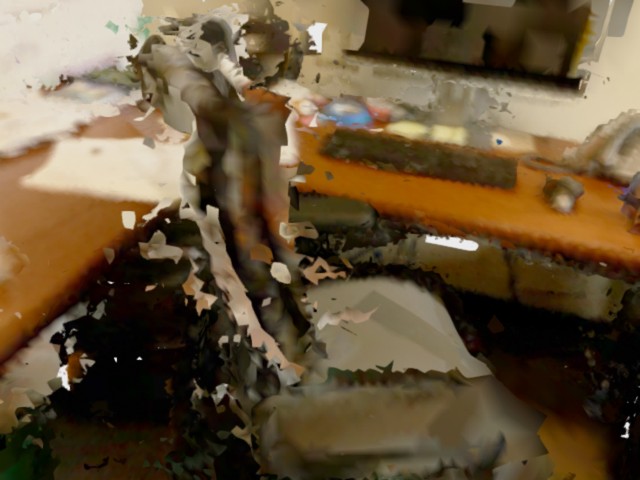} &
\includegraphics[width=0.155\linewidth]{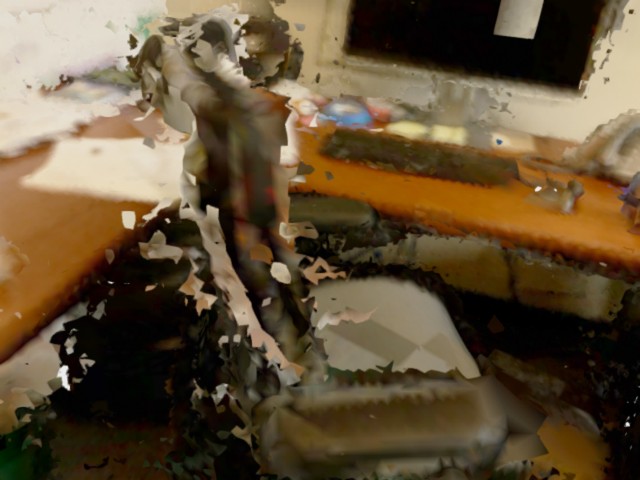} &
\includegraphics[width=0.155\linewidth]{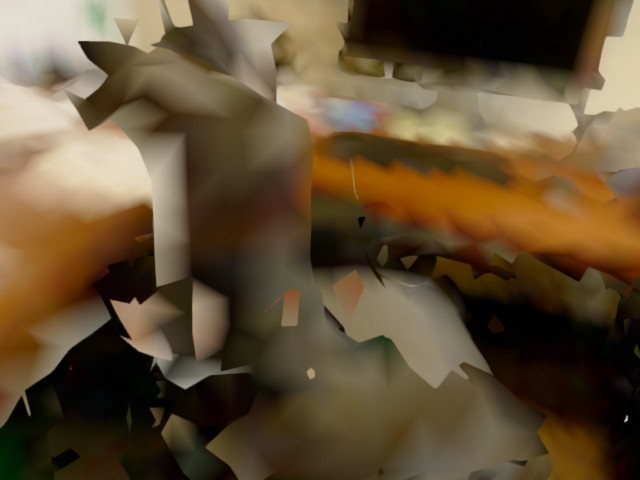} &
\includegraphics[width=0.155\linewidth]{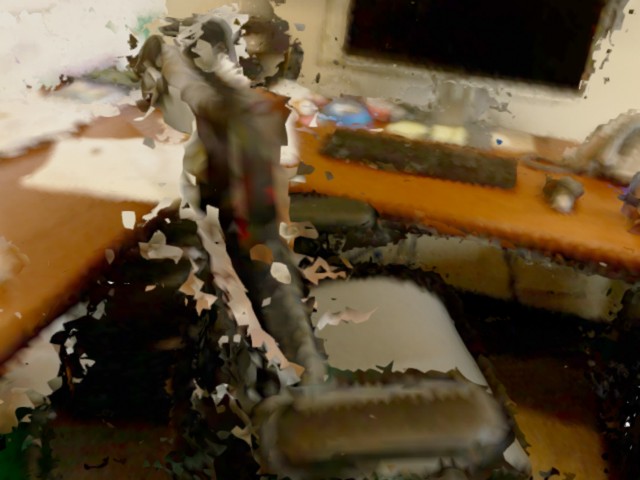} \\

\includegraphics[width=0.155\linewidth]{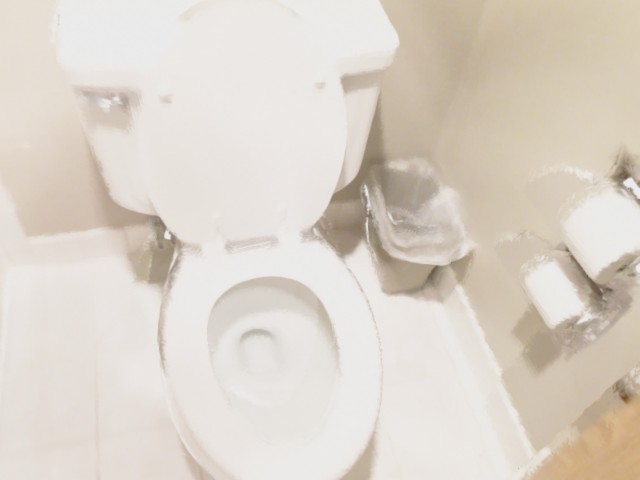} &
\includegraphics[width=0.155\linewidth]{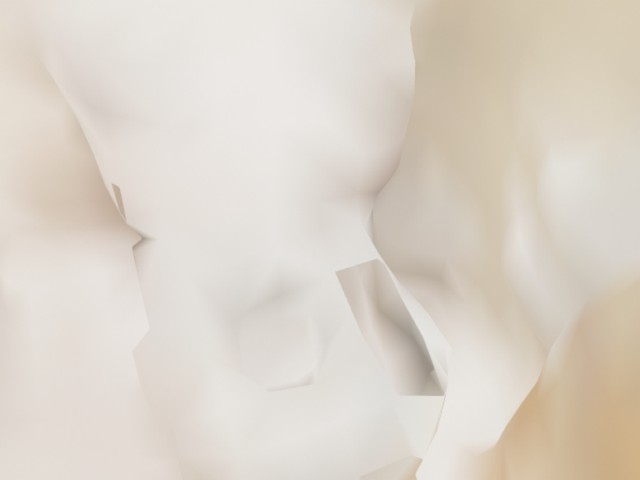} &
\includegraphics[width=0.155\linewidth]{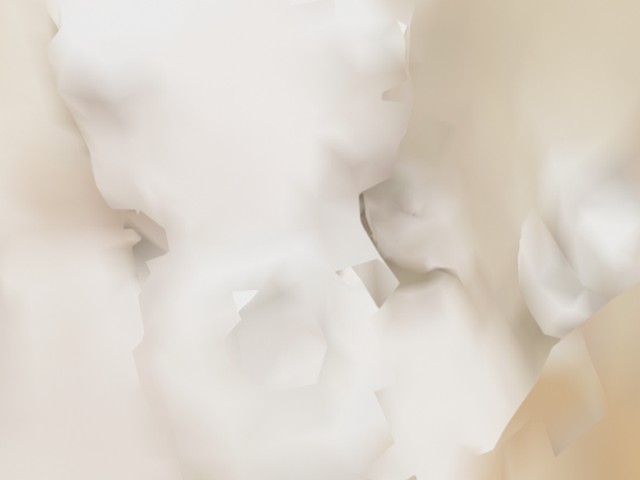} &
\includegraphics[width=0.155\linewidth]{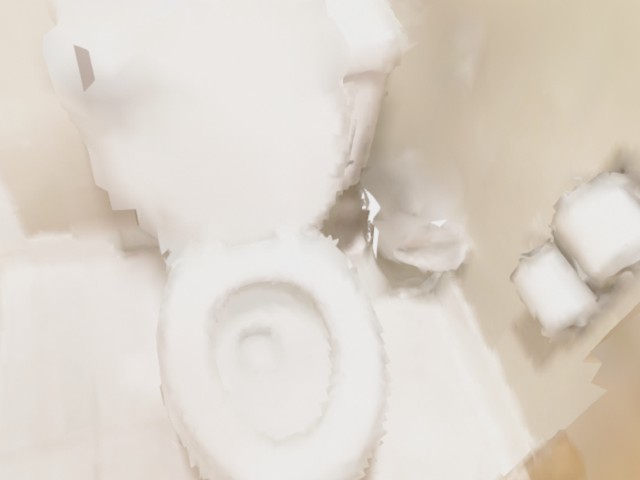} &
\includegraphics[width=0.155\linewidth]{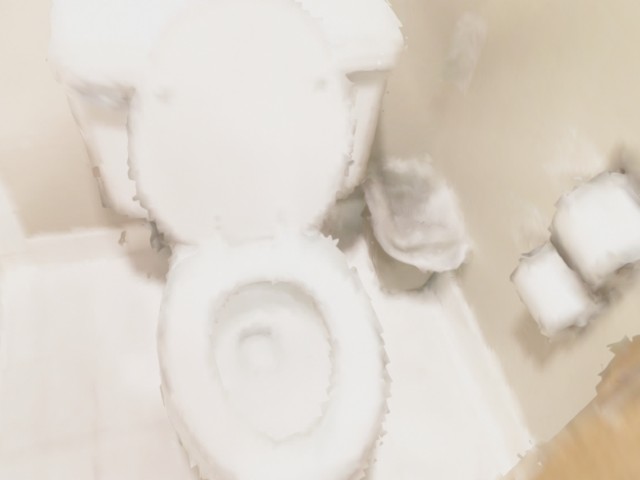} &
\includegraphics[width=0.155\linewidth]{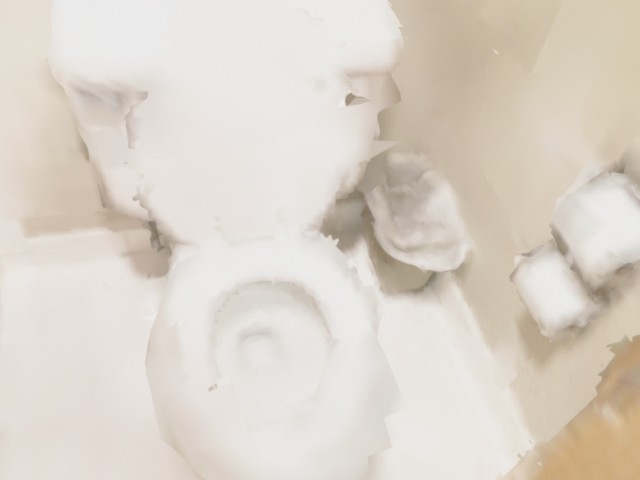} &
\includegraphics[width=0.155\linewidth]{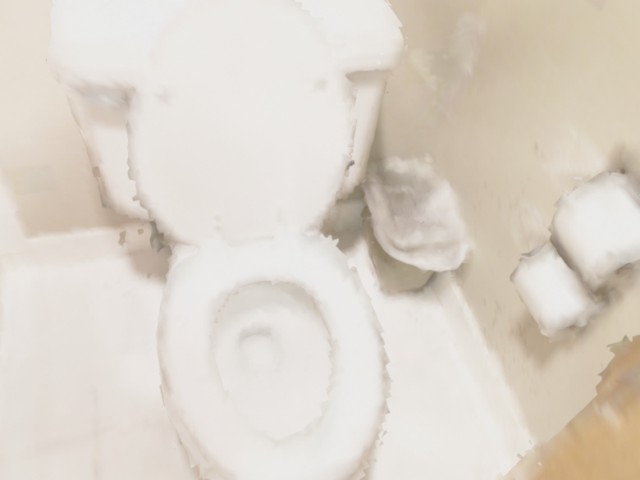} &
\includegraphics[width=0.155\linewidth]{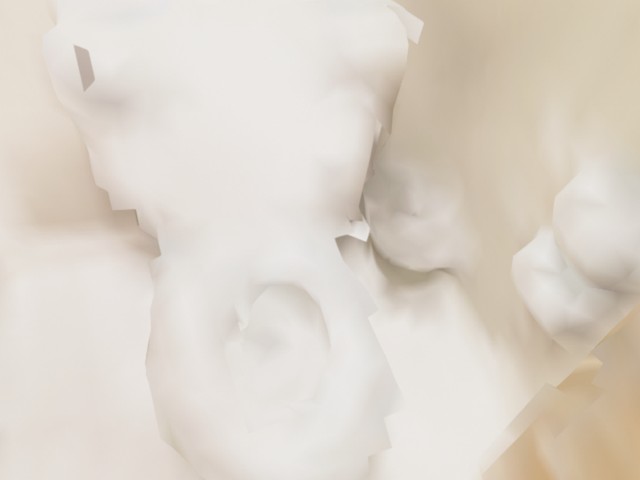} &
\includegraphics[width=0.155\linewidth]{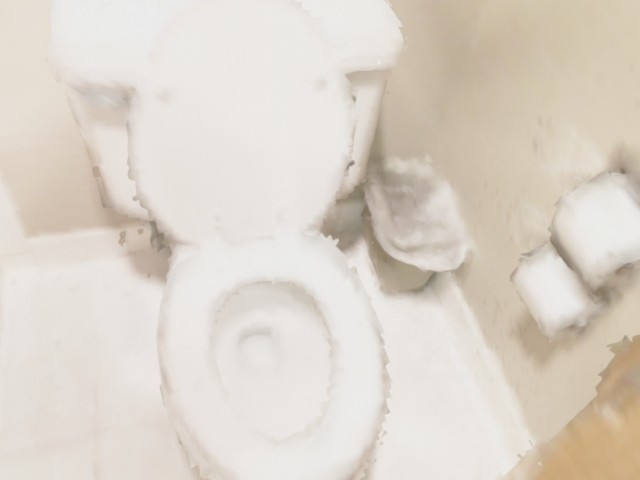} \\

\end{tabular}}
\caption{
\textbf{Rendered images} on the ScanNet dataset from reconstructed meshes using SegFormer~\cite{xie2021segformer} for semantic labeling.
Columns show increasing reconstruction quality from coarse \textsc{Voxblox} (8\,cm), through adaptive methods and a mid-resolution \textsc{Voxblox} (4\,cm), to the fine-grained 1\,cm baseline.
Our deterministic entropy-based method (\textsc{Un-Map-Adapt}) and the budget-aware RL model (\textsc{UnRL}) recover sharper geometry, cleaner object boundaries, and more stable textures than \textsc{Map-Adapt}, without relying on hand-crafted semantic class lists.
The RL variants (25\%, 50\%, 75\%) demonstrate controlled degradation by prioritizing high-uncertainty regions under increasing memory constraints.
}
\label{fig:qualitative_summary_rgb_segformer}
\end{figure}

\begin{figure}[t]
\centering
\setlength{\tabcolsep}{1pt}
\renewcommand{\arraystretch}{0.75}

\resizebox{\linewidth}{!}{%
\begin{tabular}{ccccccccc}
\scriptsize \textsc{Ground Truth} &
\scriptsize \textsc{Voxblox} (8cm) &
\scriptsize \ablUncertaintyGeoTextureRL (25\%) &
\scriptsize \ablUncertaintyGeoTextureRL (50\%) &
\scriptsize \ablUncertaintyGeoTextureRL (75\%) &
\scriptsize \ablMapadaptGeo &
\scriptsize \ablUncertaintyGeoTexture &
\scriptsize \textsc{Voxblox} (4cm) &
\scriptsize \textsc{Voxblox} (1cm) \\

\includegraphics[width=0.155\linewidth]{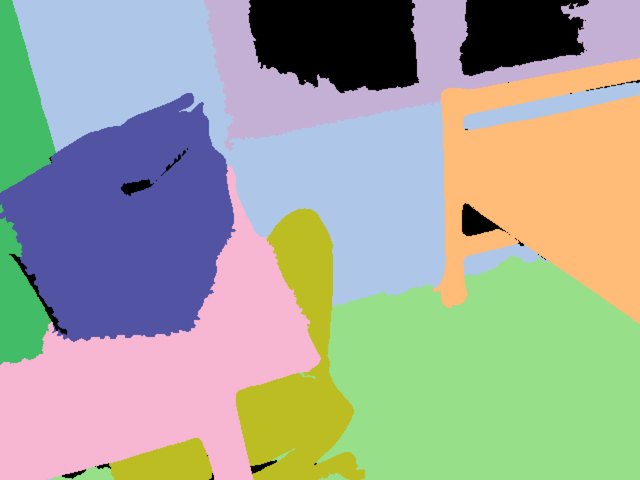} &
\includegraphics[width=0.155\linewidth]{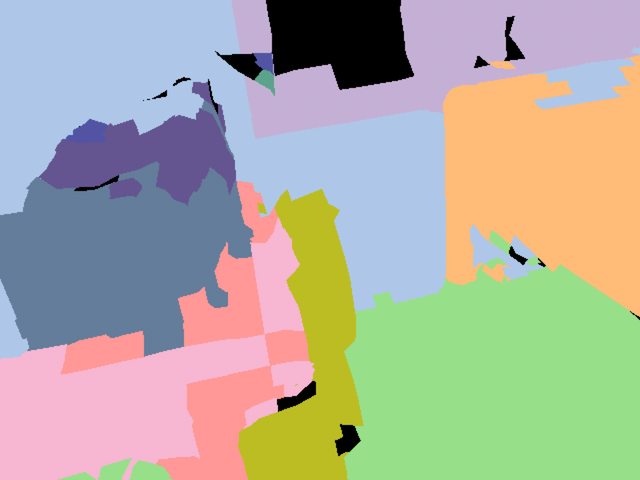} &
\includegraphics[width=0.155\linewidth]{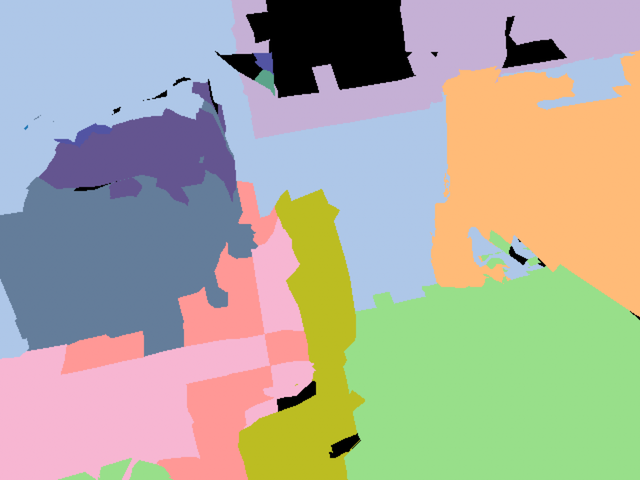} &
\includegraphics[width=0.155\linewidth]{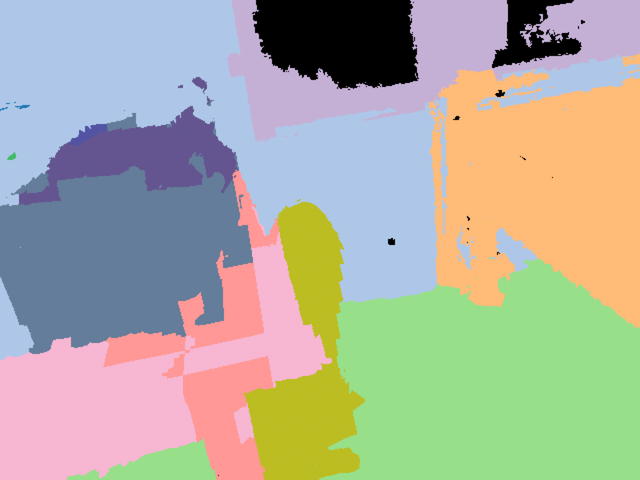} &
\includegraphics[width=0.155\linewidth]{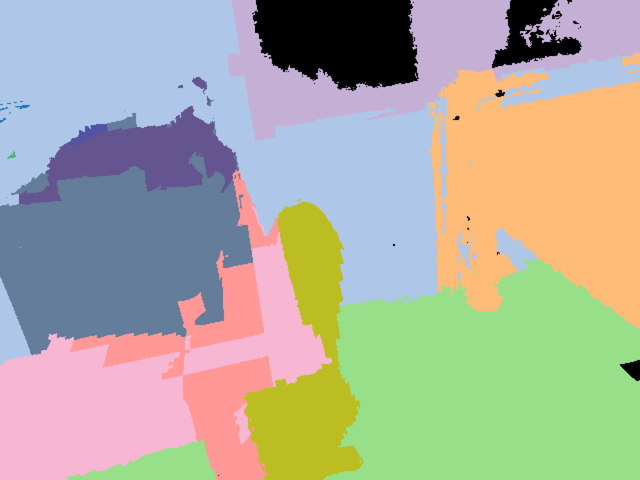} &
\includegraphics[width=0.155\linewidth]{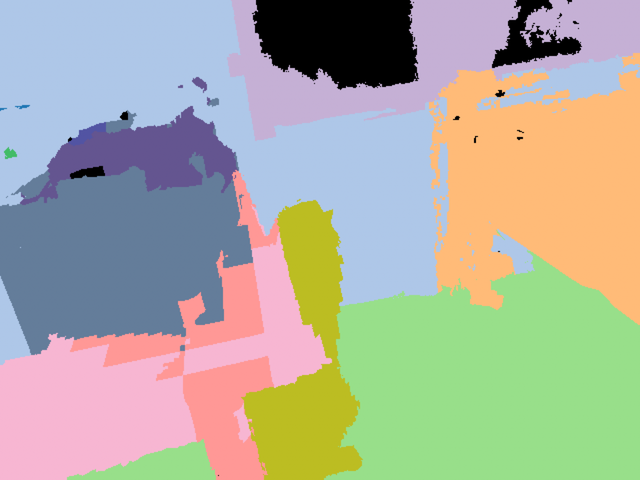} &
\includegraphics[width=0.155\linewidth]{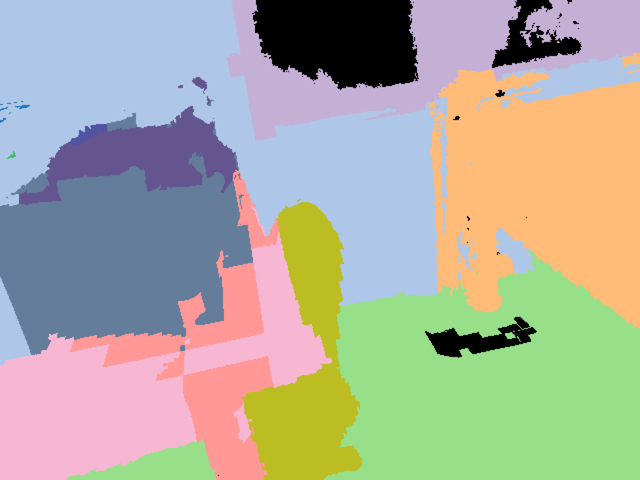} &
\includegraphics[width=0.155\linewidth]{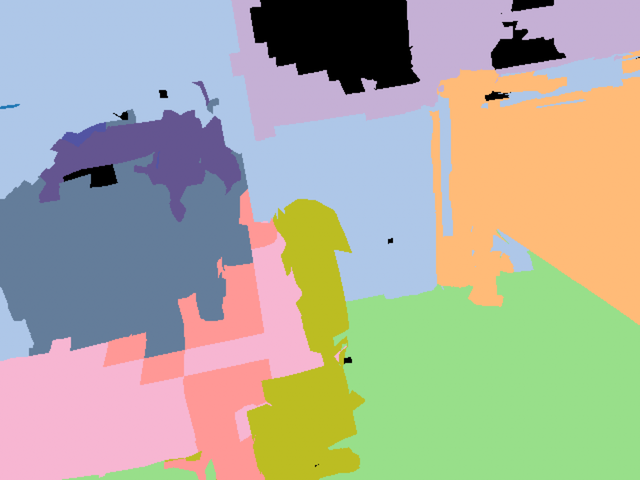} &
\includegraphics[width=0.155\linewidth]{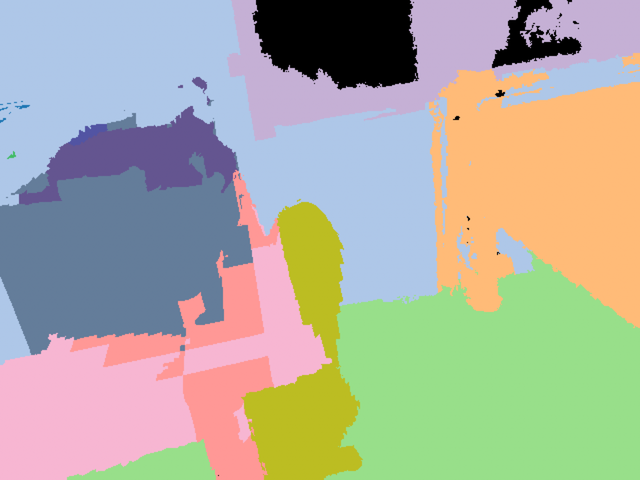} \\

\includegraphics[width=0.155\linewidth]{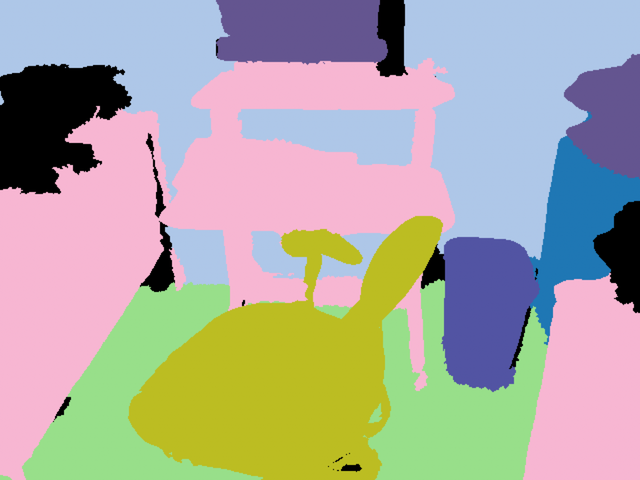} &
\includegraphics[width=0.155\linewidth]{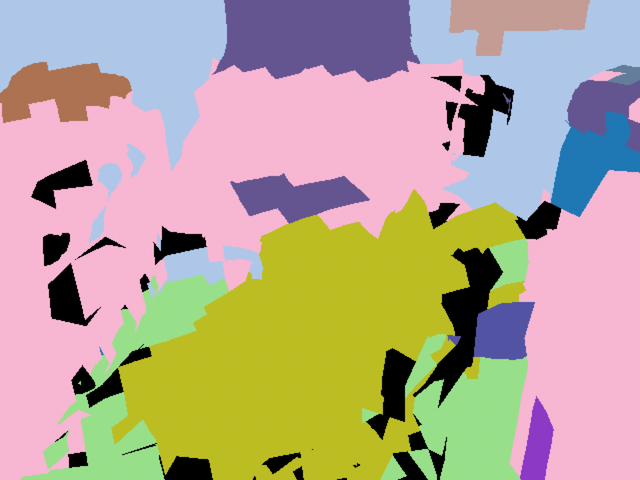} &
\includegraphics[width=0.155\linewidth]{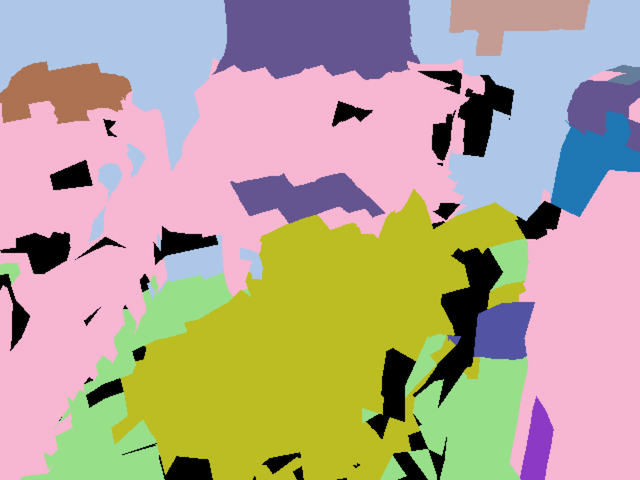} &
\includegraphics[width=0.155\linewidth]{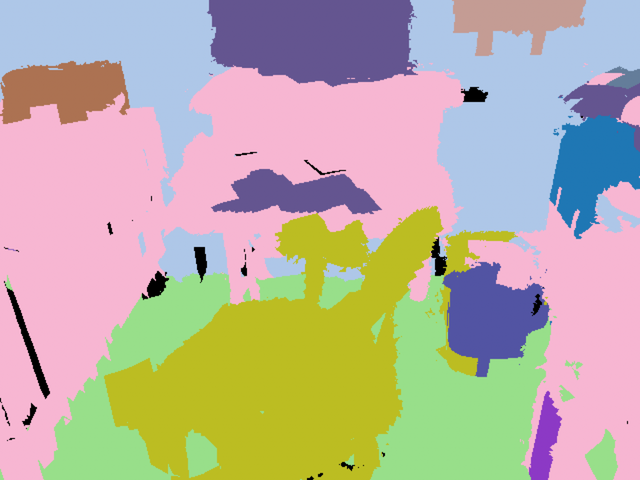} &
\includegraphics[width=0.155\linewidth]{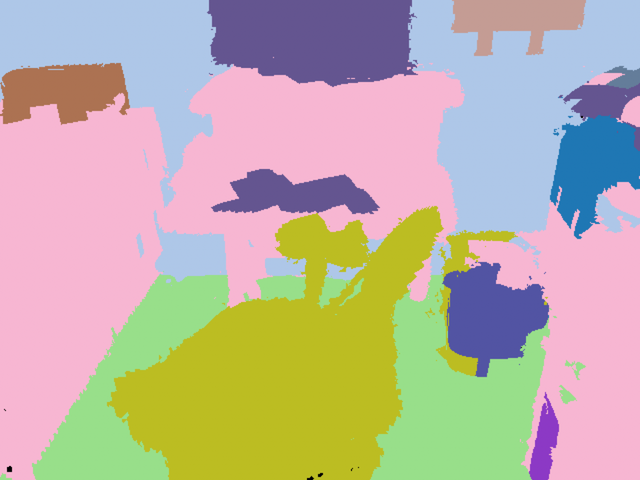} &
\includegraphics[width=0.155\linewidth]{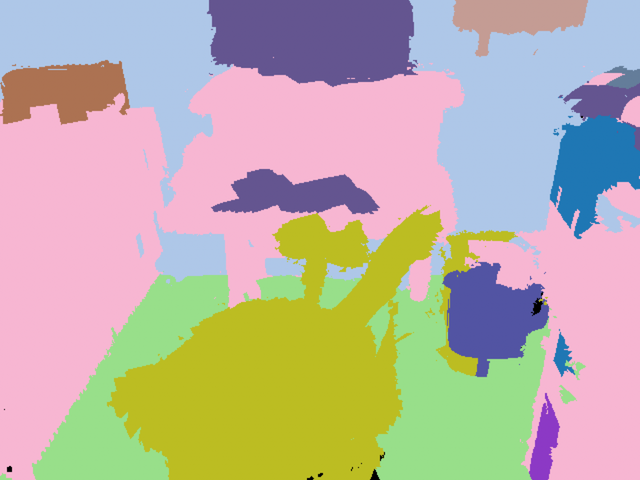} &
\includegraphics[width=0.155\linewidth]{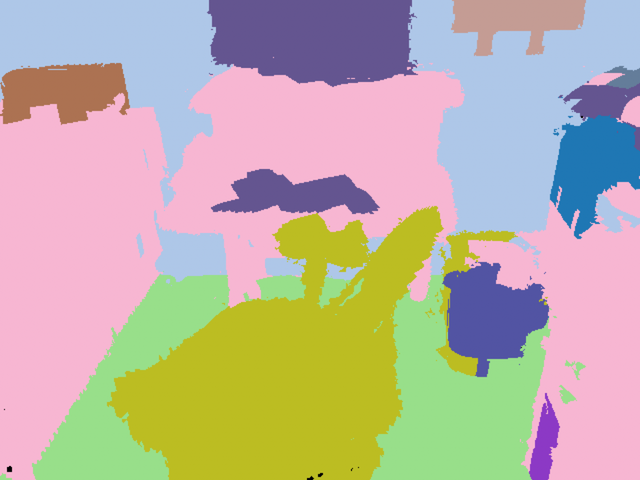} &
\includegraphics[width=0.155\linewidth]{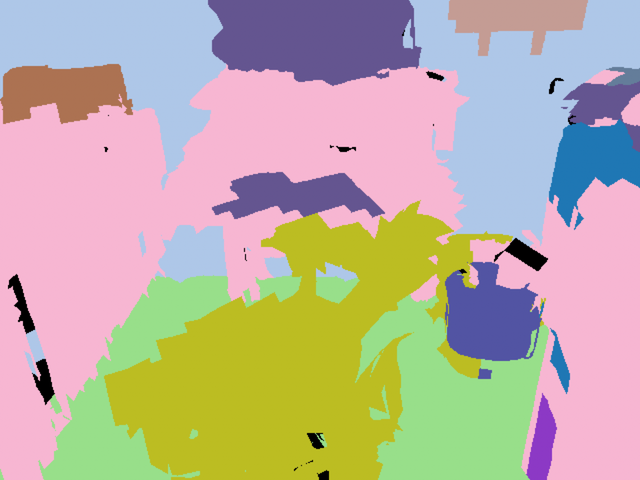} &
\includegraphics[width=0.155\linewidth]{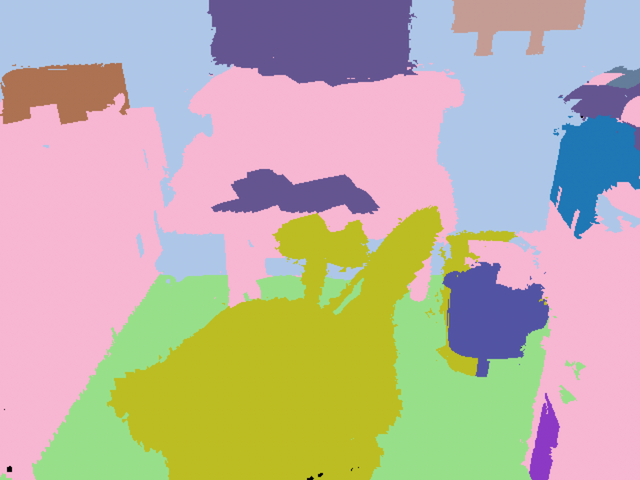} \\

\includegraphics[width=0.155\linewidth]{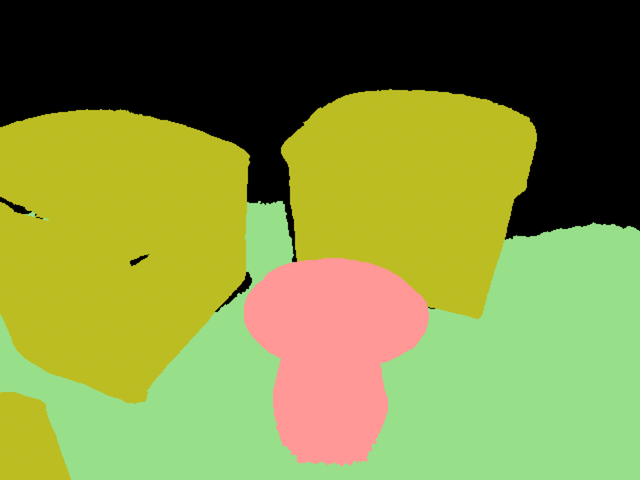} &
\includegraphics[width=0.155\linewidth]{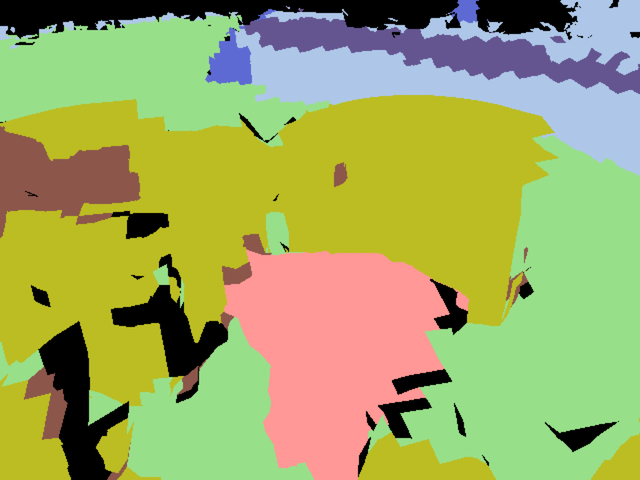} &
\includegraphics[width=0.155\linewidth]{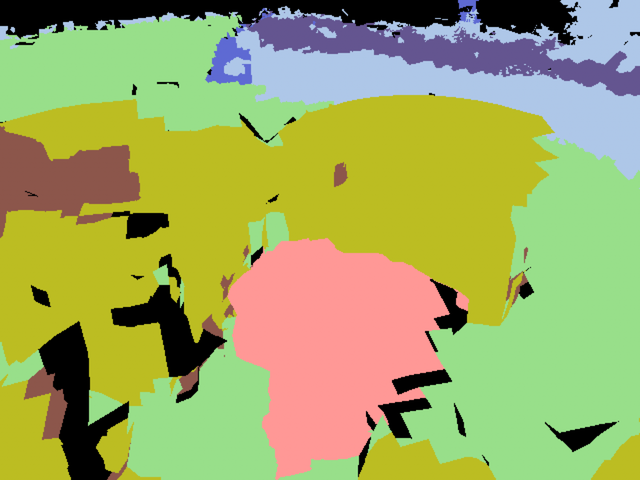} &
\includegraphics[width=0.155\linewidth]{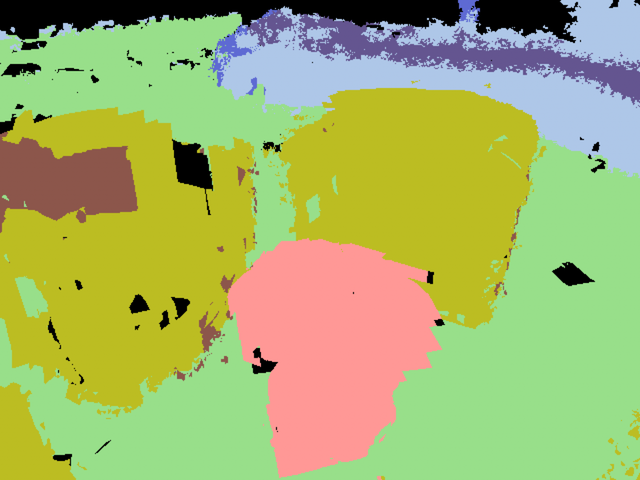} &
\includegraphics[width=0.155\linewidth]{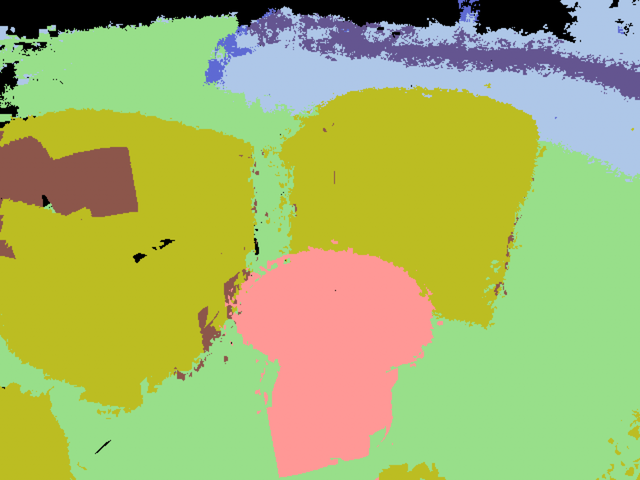} &
\includegraphics[width=0.155\linewidth]{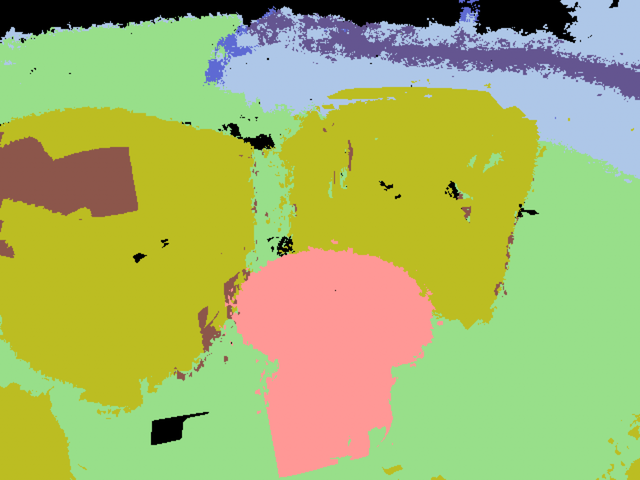} &
\includegraphics[width=0.155\linewidth]{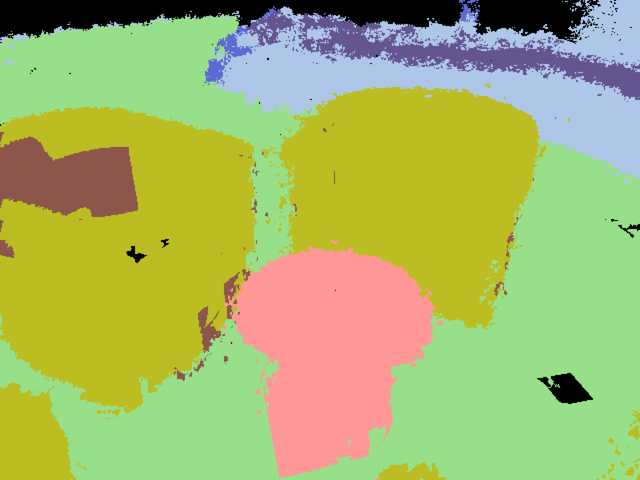} &
\includegraphics[width=0.155\linewidth]{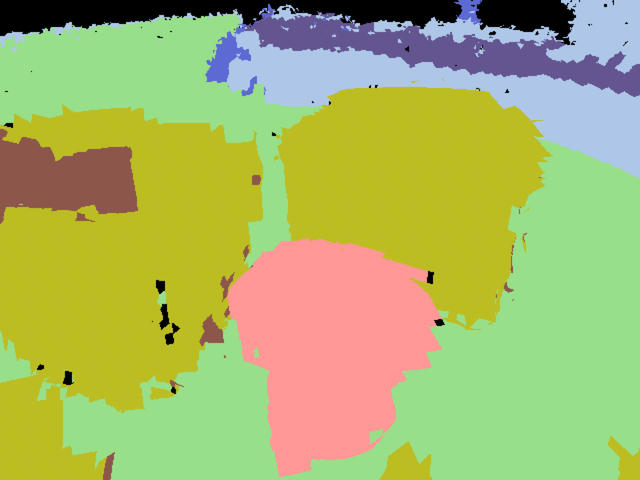} &
\includegraphics[width=0.155\linewidth]{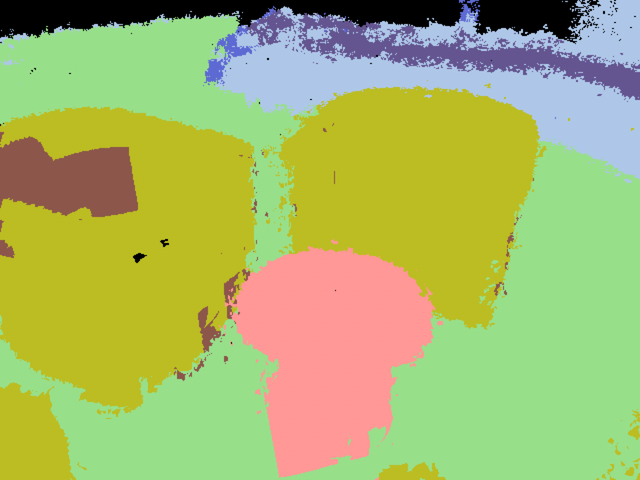} \\

\includegraphics[width=0.155\linewidth]{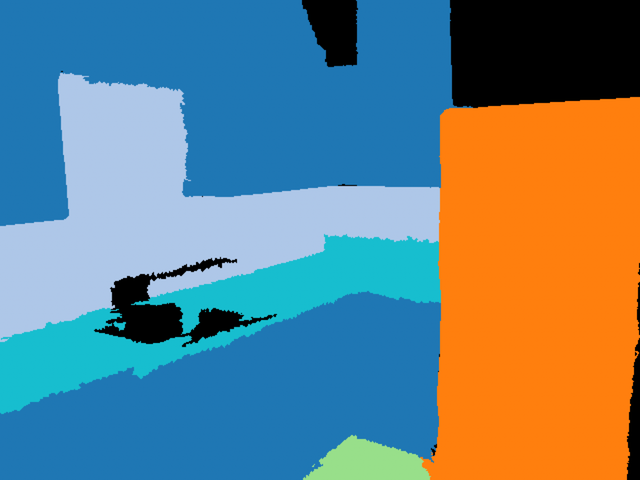} &
\includegraphics[width=0.155\linewidth]{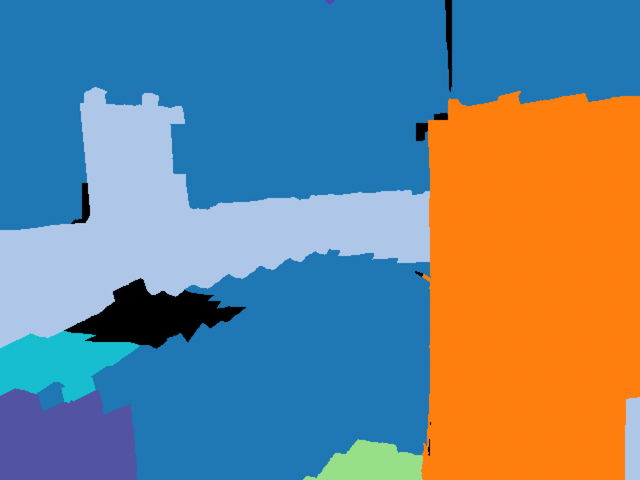} &
\includegraphics[width=0.155\linewidth]{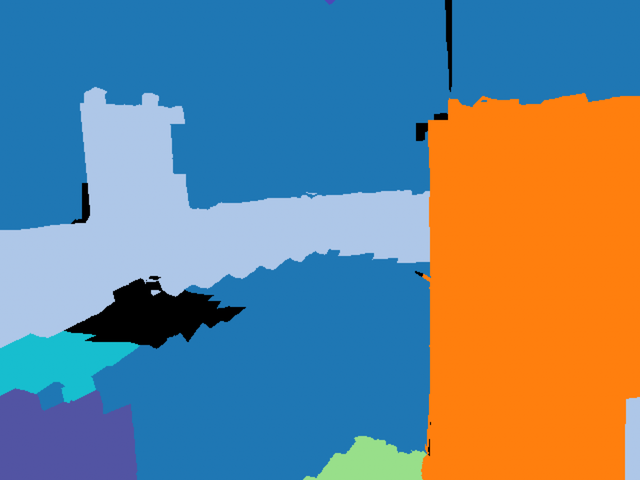} &
\includegraphics[width=0.155\linewidth]{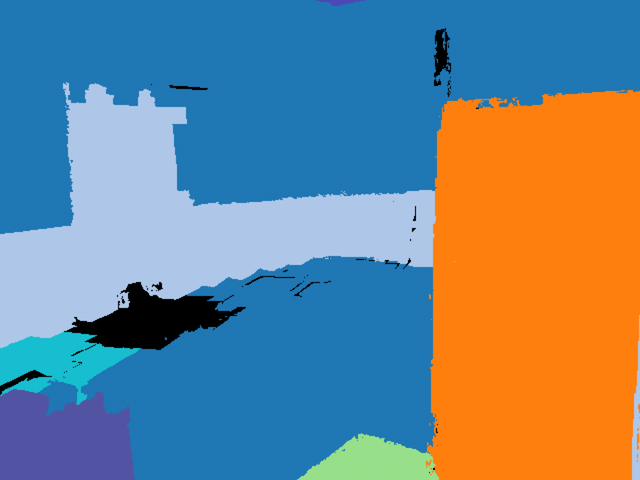} &
\includegraphics[width=0.155\linewidth]{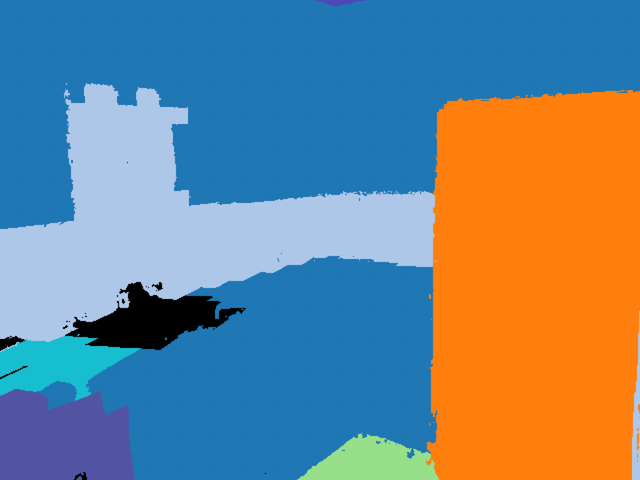} &
\includegraphics[width=0.155\linewidth]{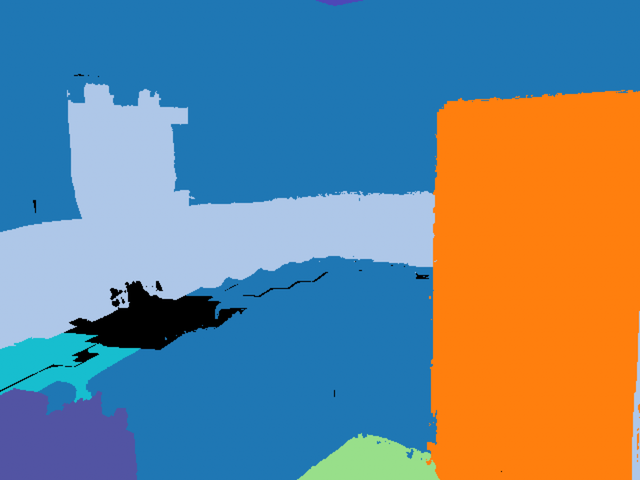} &
\includegraphics[width=0.155\linewidth]{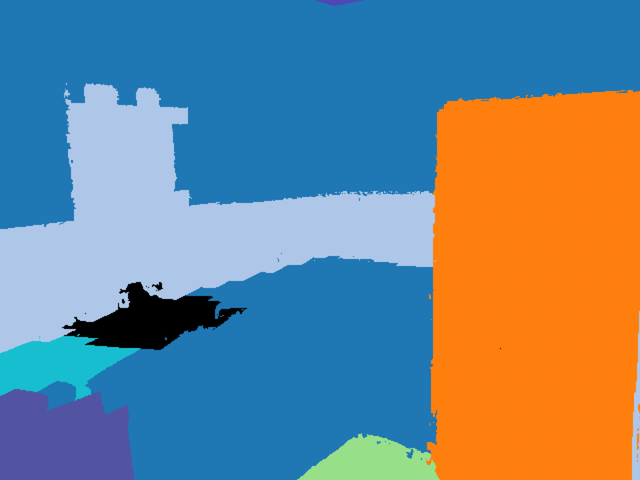} &
\includegraphics[width=0.155\linewidth]{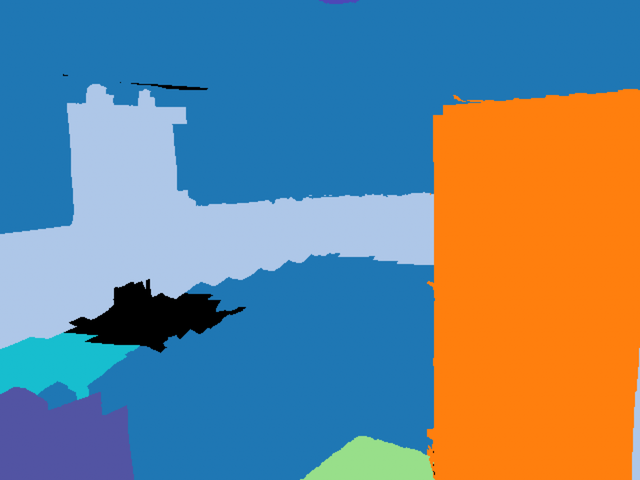} &
\includegraphics[width=0.155\linewidth]{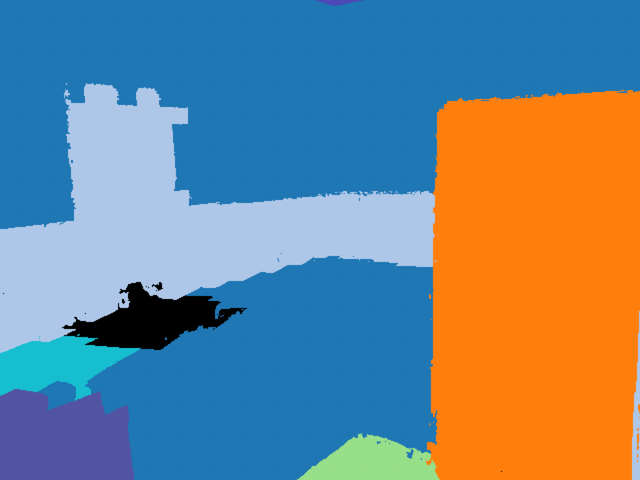} \\

\includegraphics[width=0.155\linewidth]{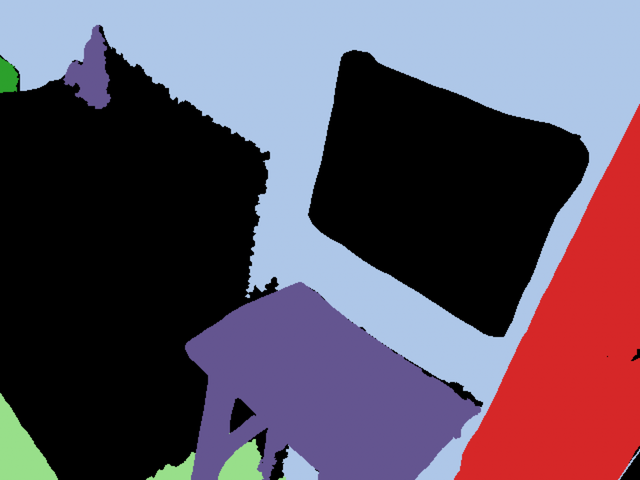} &
\includegraphics[width=0.155\linewidth]{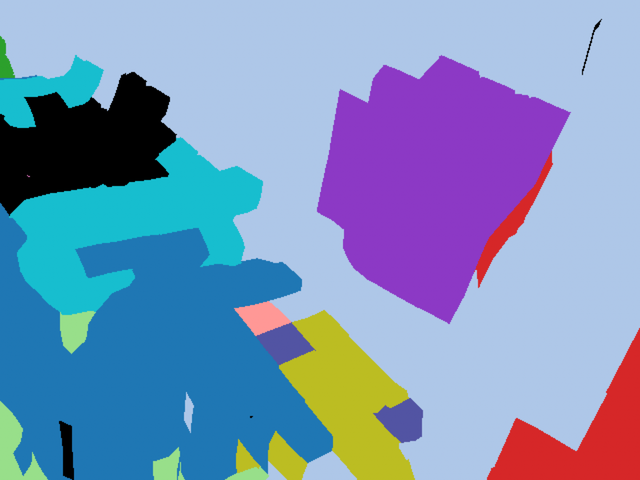} &
\includegraphics[width=0.155\linewidth]{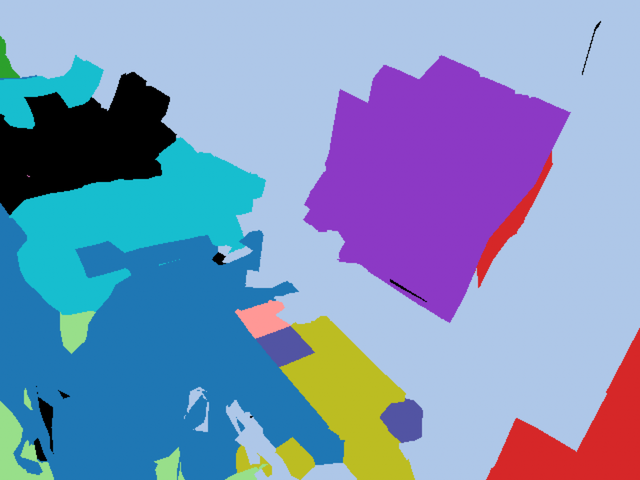} &
\includegraphics[width=0.155\linewidth]{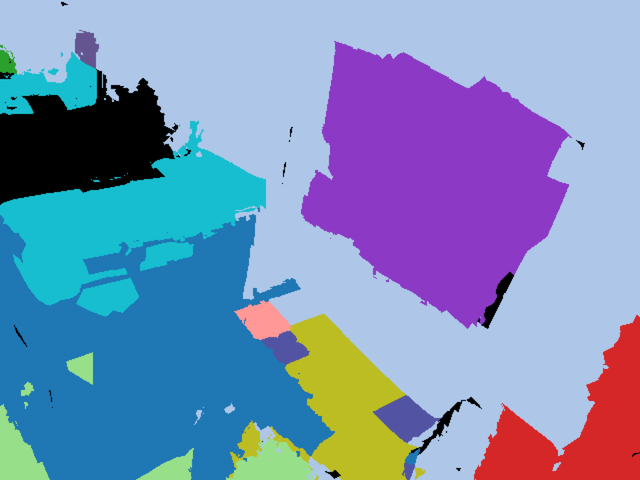} &
\includegraphics[width=0.155\linewidth]{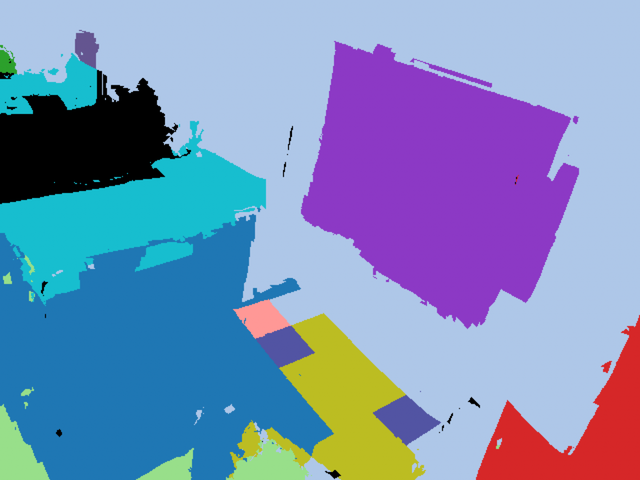} &
\includegraphics[width=0.155\linewidth]{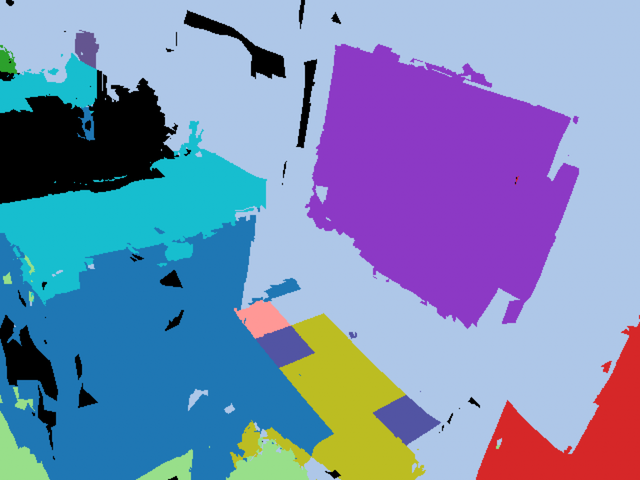} &
\includegraphics[width=0.155\linewidth]{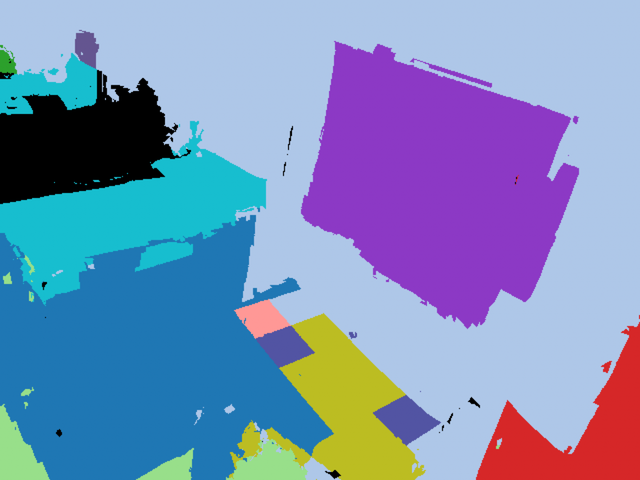} &
\includegraphics[width=0.155\linewidth]{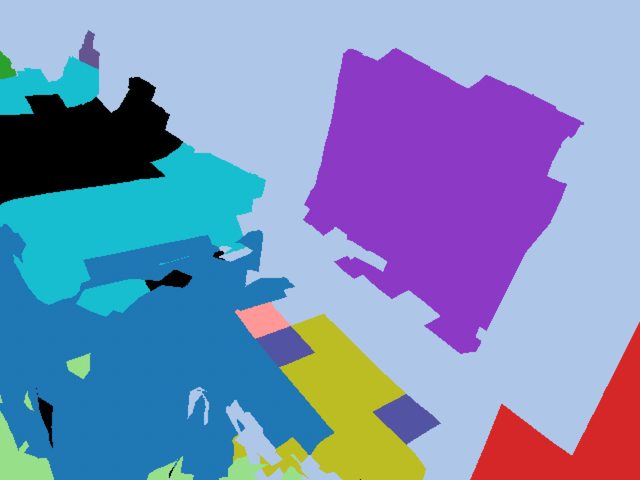} &
\includegraphics[width=0.155\linewidth]{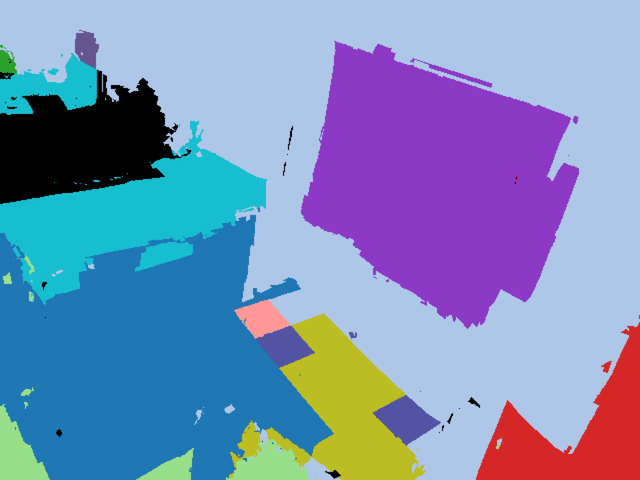} \\

\includegraphics[width=0.155\linewidth]{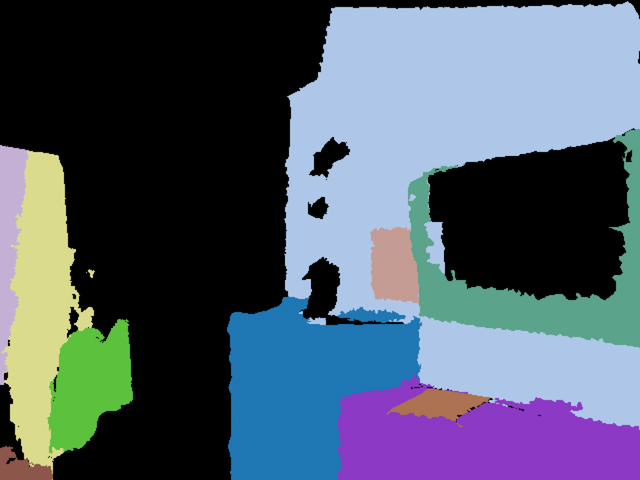} &
\includegraphics[width=0.155\linewidth]{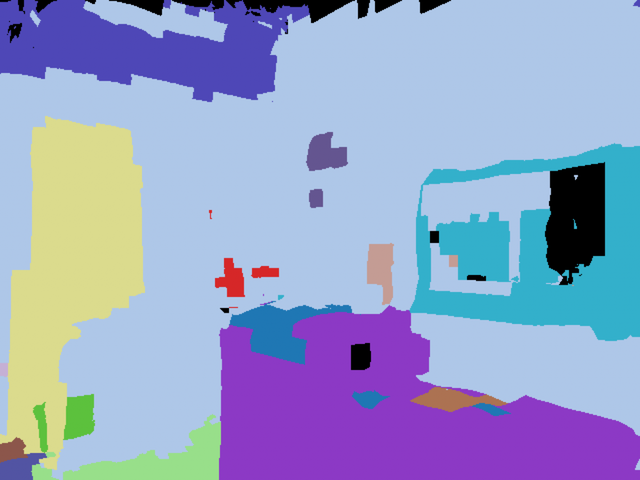} &
\includegraphics[width=0.155\linewidth]{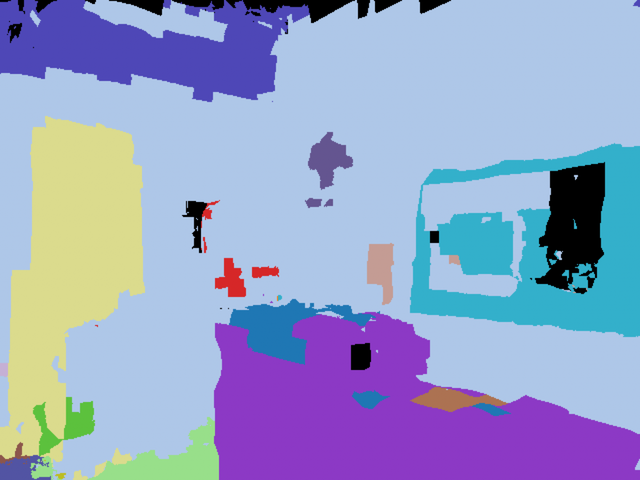} &
\includegraphics[width=0.155\linewidth]{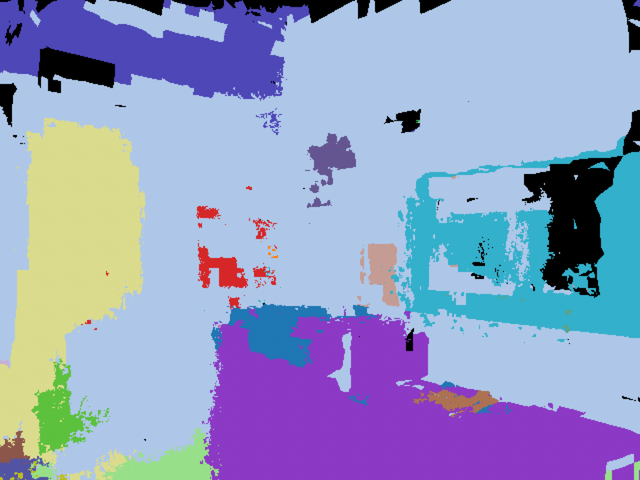} &
\includegraphics[width=0.155\linewidth]{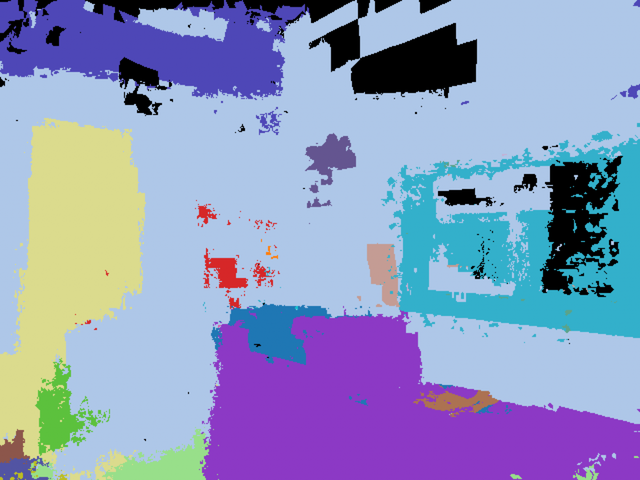} &
\includegraphics[width=0.155\linewidth]{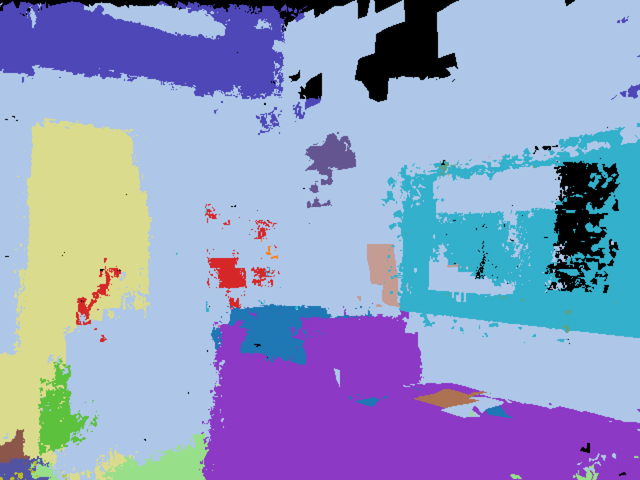} &
\includegraphics[width=0.155\linewidth]{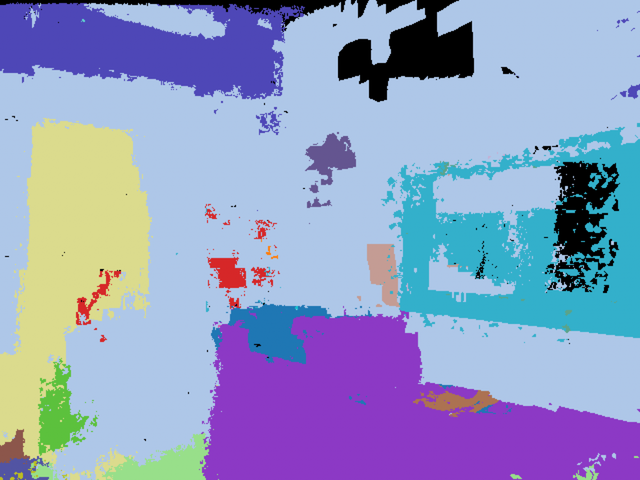} &
\includegraphics[width=0.155\linewidth]{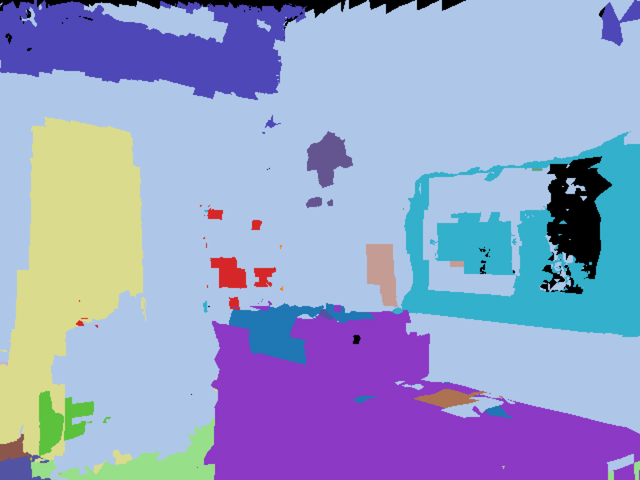} &
\includegraphics[width=0.155\linewidth]{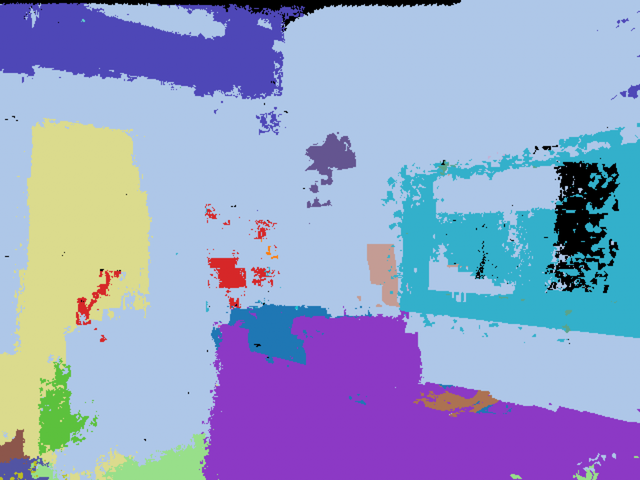} \\

\includegraphics[width=0.155\linewidth]{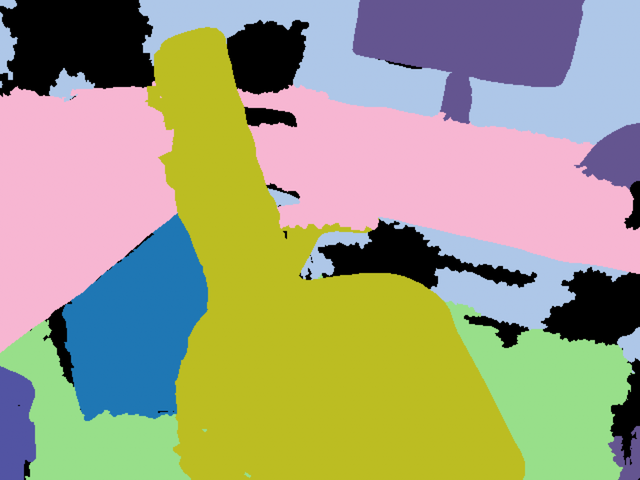} &
\includegraphics[width=0.155\linewidth]{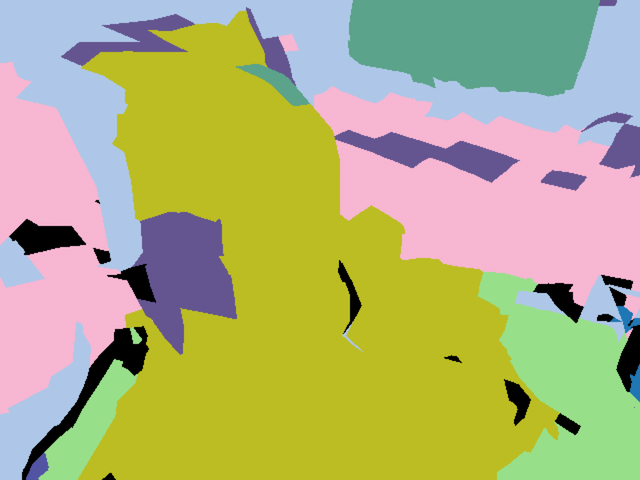} &
\includegraphics[width=0.155\linewidth]{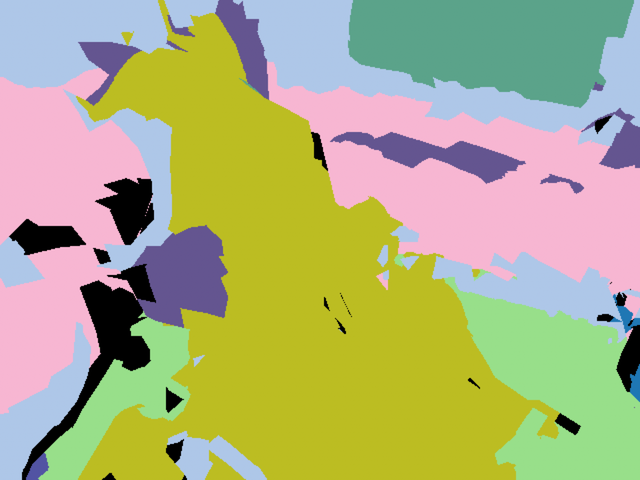} &
\includegraphics[width=0.155\linewidth]{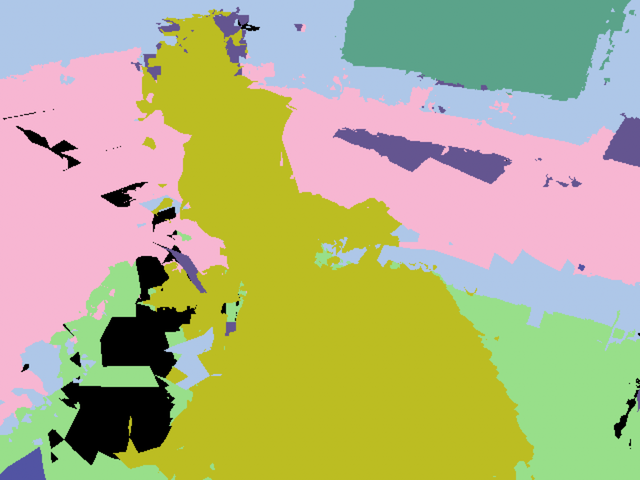} &
\includegraphics[width=0.155\linewidth]{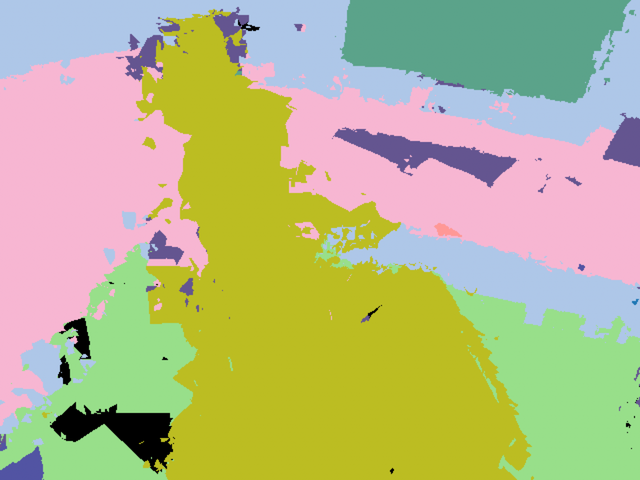} &
\includegraphics[width=0.155\linewidth]{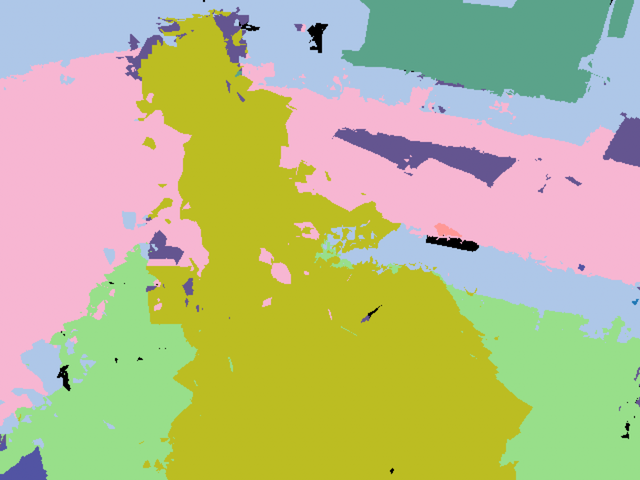} &
\includegraphics[width=0.155\linewidth]{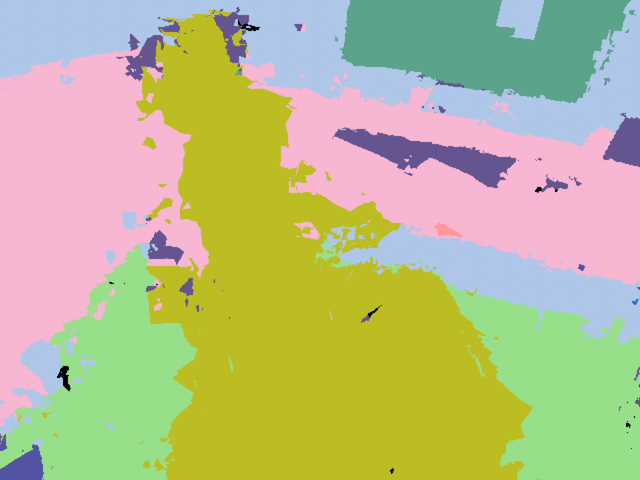} &
\includegraphics[width=0.155\linewidth]{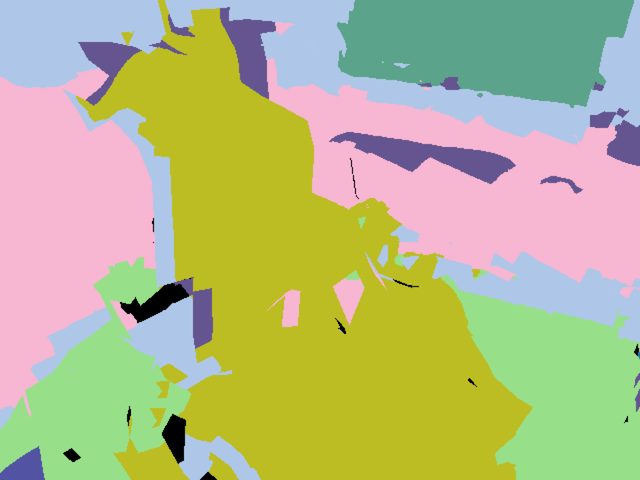} &
\includegraphics[width=0.155\linewidth]{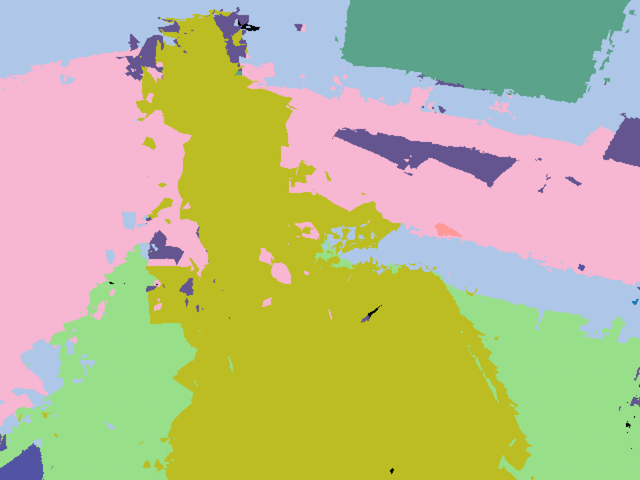} \\

\includegraphics[width=0.155\linewidth]{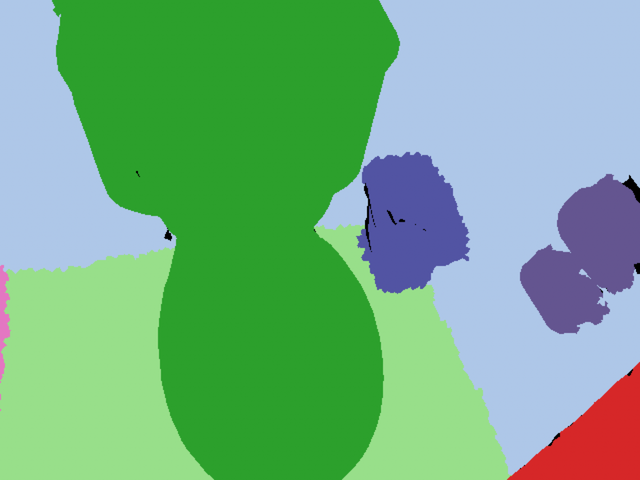} &
\includegraphics[width=0.155\linewidth]{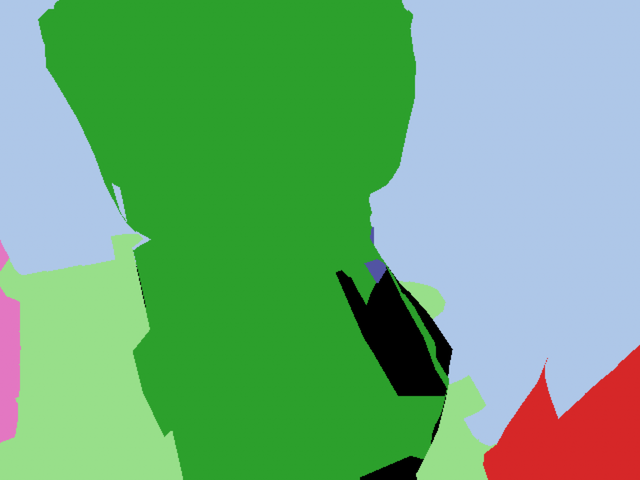} &
\includegraphics[width=0.155\linewidth]{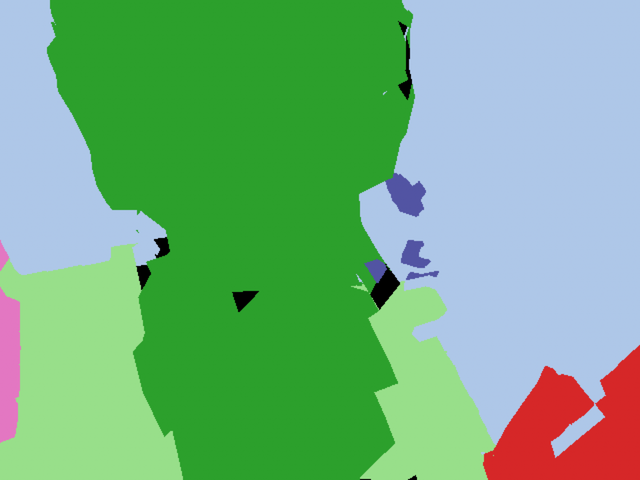} &
\includegraphics[width=0.155\linewidth]{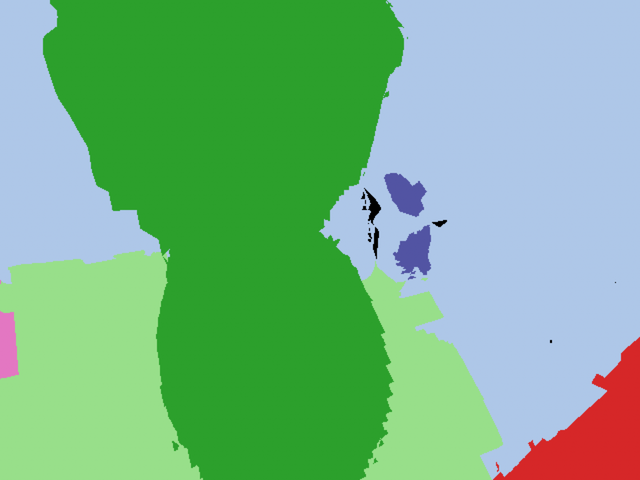} &
\includegraphics[width=0.155\linewidth]{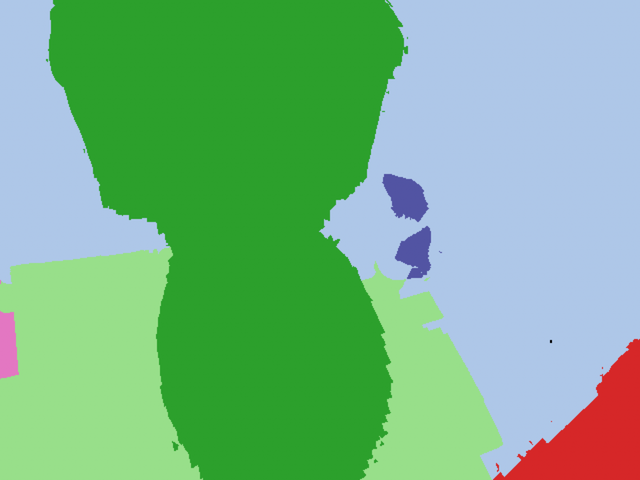} &
\includegraphics[width=0.155\linewidth]{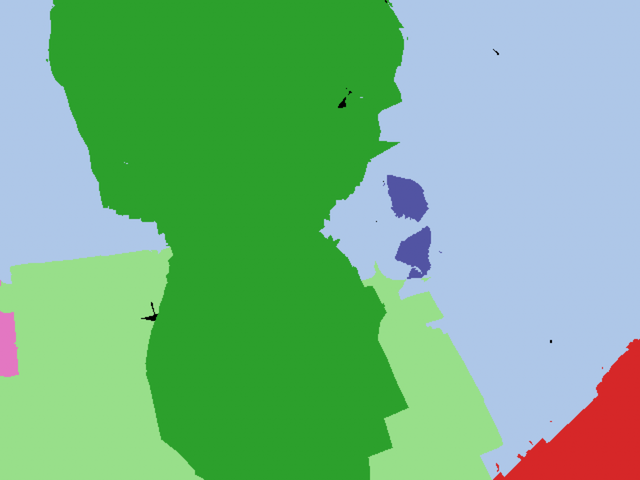} &
\includegraphics[width=0.155\linewidth]{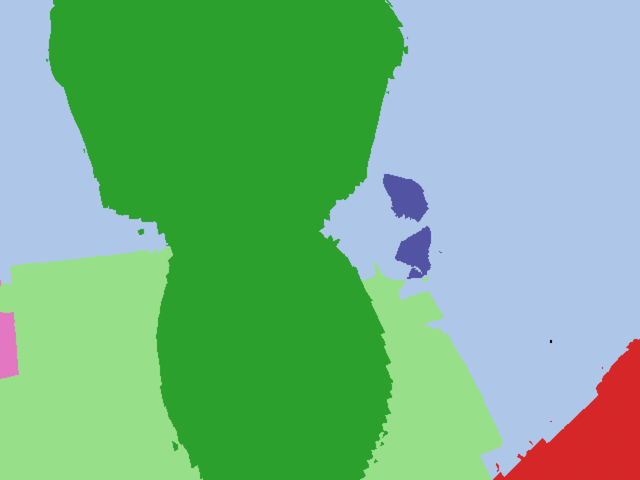} &
\includegraphics[width=0.155\linewidth]{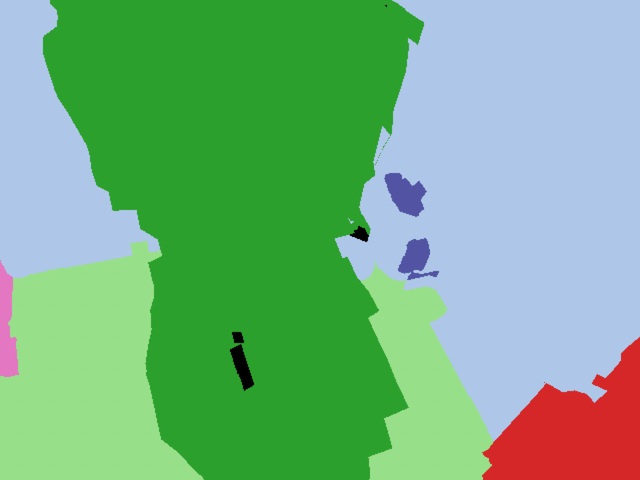} &
\includegraphics[width=0.155\linewidth]{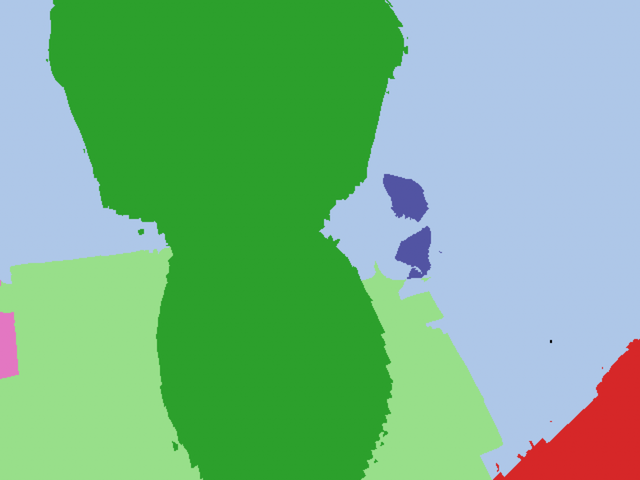} \\

\end{tabular}}
\caption{
\textbf{Semantic segmentations} on the ScanNet dataset rendered from reconstructed meshes using SegFormer~\cite{xie2021segformer}.
Columns show increasing reconstruction quality from coarse \textsc{Voxblox} (8\,cm), through adaptive methods and a mid-resolution \textsc{Voxblox} (4\,cm), to the fine-grained 1\,cm baseline.
Our deterministic entropy-based method (\textsc{Un-Map-Adapt}) and the budget-aware RL model (\textsc{UnRL}) recover cleaner object boundaries and more spatially consistent semantic regions than \textsc{Map-Adapt}, without relying on hand-crafted semantic class lists.
The RL variants (25\%, 50\%, 75\%) demonstrate controlled degradation by prioritizing high-uncertainty regions under increasing memory constraints.
}
\label{fig:qualitative_summary_sem_segformer}
\end{figure}

\Cref{fig:qualitative_summary_sem} provides semantic renderings for the same scenes and ablations as in Fig.~3 of the main paper, which presents qualitative texture reconstruction comparisons.
HSSD contains fewer object categories dominated by background classes, which reduces the impact of semantic refinement. Conversely, ScanNet contains a richer class distribution but is limited by RefineNet accuracy. SegFormer alleviates this limitation, producing sharper maps and making the benefits of our adaptive strategy more apparent. \Cref{fig:qualitative_summary_sem_segformer} presents the corresponding semantic renderings using SegFormer, where the improved backbone yields sharper class boundaries and a more visible separation between the adaptive methods.

\begin{table}[t]
\caption{\textbf{Accuracy and efficiency on ScanNet with SegFormer~\cite{xie2021segformer} providing the per-frame semantic segmentations.}
We report geometric reconstruction quality (completion error and completion ratio within 5\,cm), semantic accuracy, and texture fidelity (MS-SSIM, LPIPS), all obtained from mesh renderings in held-out test views.
Efficiency metrics include total voxel count and peak memory usage in gigabytes.
We also show MAP-ADAPT with random semantic-to-resolution assignments, demonstrating its sensitivity to user-defined class lists.
Our full entropy-driven method (\ablUncertaintyGeoTexture) outperforms MAP-ADAPT in \textit{all} accuracy metrics while avoiding reliance on manually specified semantic class lists.
The RL-based variants enable explicit control of memory through voxel-budget constraints, offering competitive accuracy with reduced memory.}
\label{tab:accuracy_scannet_segformer_full}
\centering
\setlength{\tabcolsep}{4pt}
\resizebox{\textwidth}{!}{
\begin{tabular}{lccccccc}
\toprule
\textbf{Method}
& \textbf{Compl. Err$\downarrow$}
& \textbf{Compl.$<$5cm (\%)$\uparrow$}
& \textbf{Sem. Acc.$\uparrow$}
& \textbf{MS-SSIM$\uparrow$}
& \textbf{LPIPS$\downarrow$}
& \textbf{Total Voxels (M)}
& \textbf{Memory (GB)$\downarrow$} \\
\midrule
\multicolumn{8}{l}{\textbf{Fixed-Resolution Baselines}} \\
\ablLarge~\cite{voxblox}  & 3.093 & 81.050 & 0.674 & 0.611 & 0.52 & \textbf{0.57} & \textbf{0.137}   \\
\ablMiddle~\cite{voxblox}   & 2.209 & 91.306 & 0.701 & 0.647 & 0.47 & 1.12 & 0.267 \\
\ablSmall~\cite{voxblox}    & \textbf{1.448} & \textbf{99.246} & \textbf{0.705} & \textbf{0.65} & \textbf{0.40} & 35.50 & 8.464 \\
\midrule
\multicolumn{8}{l}{\textbf{Adaptive Methods (w/o Budget Constraint)}} \\
\ablMapadaptGeo~\cite{zheng2024mapadapt} & 1.58 & 98.03 & \textbf{0.699} & \textbf{0.64} & 0.43 & 25.2 & 6.01 \\
\ablMapadaptGeo~\cite{zheng2024mapadapt} (Random) & 1.70 & 96.99 & 0.690 & 0.63 & 0.45 & \textbf{21.9} & \textbf{5.21} \\
\ablUncertaintyGeoTexture & \textbf{1.49} & \textbf{98.81} & \textbf{0.700} & \textbf{0.64} & \textbf{0.42} & 33.4 & 7.97 \\
\midrule
\multicolumn{8}{l}{\textbf{Adaptive Methods (w/ Budget Constraint)}} \\
\ablUncertaintyGeoTextureRL (75\%)
& \textbf{1.56} & \textbf{97.26} & \textbf{0.701} & \textbf{0.65} & \textbf{0.41} & 32.0 & 7.64 \\
\ablUncertaintyGeoTextureRL (50\%)
& 1.82 & 94.60 & 0.696 & 0.64 & 0.43 & 15.9 & 3.80 \\
\ablUncertaintyGeoTextureRL (25\%)
& 2.76 & 84.81 & 0.679 & 0.62 & 0.50 & \textbf{1.1} & \textbf{0.25} \\
\bottomrule
\end{tabular}
}
\end{table}

\section{Sensitivity Analysis}
\label[appendix]{sec:sensitivity}

We conduct a sensitivity analysis of the key hyperparameters for both the RL-based variant (UnRL) and our deterministic method (UN-MAP-ADAPT). All ablations are evaluated on the ScanNet validation set using SegFormer as the segmentation backbone. The UnRL ablations (\Cref{sec:unrl_sensitivity}) are averaged over 10 scenes, while the UN-MAP-ADAPT threshold sweep (\Cref{sec:unmapadapt_sensitivity}) is averaged over 5 scenes. We first ablate block size, number of blocks, and reward weight configuration for UnRL (\Cref{sec:unrl_sensitivity}), then sweep the subdivision threshold settings for UN-MAP-ADAPT (\Cref{sec:unmapadapt_sensitivity}). Our goal is to demonstrate that both methods are robust to reasonable hyperparameter choices, and that the RL agent's advantage lies precisely in eliminating the need for manual threshold tuning.

\subsection{UNRL Parameter Sensitivity}
\label[subappendix]{sec:unrl_sensitivity}

The \ablUncertaintyGeoTextureRL agent has three key design parameters: the spatial block size $s$, the number of sampled blocks $N$ per frame, and the reward weight configuration $(w_\text{geo}, w_\text{sem}, w_\text{tex})$ that aggregates the three loss modalities into a single scalar reward.
We ablate each parameter independently while keeping all others at their default values.

\paragraph{Block size.}
The block size $s$ controls the spatial granularity of the agent's decisions: all voxels within a block share the same thresholds on the complexity proxy.
\Cref{tab:ablation_block_size} reports results for $s \in \{0.1, 0.25, 0.5, 1.0\}$\,m.
Very small blocks ($s{=}0.1$\,m) underperform because each block contains too few voxels for reliable uncertainty and complexity statistics, leading to noisy observations and poor decisions.
Performance improves substantially from 0.1\,m to 0.5\,m and plateaus at 1.0\,m.
Although $s{=}1.0$\,m is marginally better on completion metrics, we select $s{=}0.5$\,m as the default because it provides finer spatial control: the agent can assign different thresholds to neighboring regions that a 1.0\,m block would merge.
The accuracy gap between the two ($<$0.01\,cm in completion error) confirms that the method is not sensitive to this choice within a reasonable range.

\begin{table}[H]
\centering
\caption{Ablation on block size $s$. All other hyperparameters are fixed to default. $\dagger$ denotes the default configuration. Averaged over 10 ScanNet validation scenes.}
\label{tab:ablation_block_size}
\setlength{\tabcolsep}{4pt}
\resizebox{\linewidth}{!}{%
\begin{tabular}{c ccc cc cc}
\toprule
Block Size & Compl.\ Err↓ & Compl.\ $<$5cm↑ & Sem.\ Acc↑ & MS-SSIM↑ & LPIPS↓ & Voxels (M)↑ & Mem.\ (GB) \\
$s$ (m) & (cm) & (\%) & (\%) & & & & \\
\midrule
0.1  & 1.880 & 93.90 & 70.45 & 0.611 & 0.434 & 12.17 & \textbf{2.90} \\
0.25 & 1.466 & 97.77 & 71.03 & 0.618 & 0.422 & 24.97 & 5.94 \\
0.5$^\dagger$ & 1.366 & 98.96 & 71.72 & \textbf{0.626} & 0.403 & 26.76 & 6.37 \\
1.0  & \textbf{1.360} & \textbf{99.12} & \textbf{71.78} & 0.623 & \textbf{0.402} & \textbf{27.21} & 6.48 \\
\bottomrule
\end{tabular}}
\end{table}

\paragraph{Number of blocks.}
The number of sampled blocks $N$ determines how many spatial regions the agent observes and acts on per frame.
Since the policy network uses a Perceiver-based attention mechanism, computational cost scales quadratically with $N$, which motivates keeping the observation compact.
\Cref{tab:ablation_num_blocks} shows that increasing $N$ from 10 to 50 yields only modest improvement, while increasing both training and inference cost.
We use $N{=}25$ as the default, which balances observation coverage with computational efficiency.
Notably, even $N{=}10$ achieves competitive results, suggesting that the proxy-weighted importance sampling effectively directs attention to the most informative blocks regardless of the total sample count.

\begin{table}[H]
\centering
\caption{Ablation on number of blocks $N$. All other hyperparameters are fixed to default. $\dagger$ denotes the default configuration. Averaged over 10 ScanNet validation scenes.}
\label{tab:ablation_num_blocks}
\setlength{\tabcolsep}{4pt}
\resizebox{\linewidth}{!}{%
\begin{tabular}{c ccc cc cc}
\toprule
Num.\ Blocks & Compl.\ Err↓ & Compl.\ $<$5cm↑ & Sem.\ Acc↑ & MS-SSIM↑ & LPIPS↓ & Voxels (M)↑ & Mem.\ (GB) \\
$N$ & (cm) & (\%) & (\%) & & & & \\
\midrule
10   & 1.472 & 97.71 & 71.44 & 0.625 & 0.408 & 25.92 & 6.17 \\
25$^\dagger$ & 1.366 & 98.96 & 71.72 & \textbf{0.626} & \textbf{0.403} & 26.76 & 6.37 \\
50   & \textbf{1.353} & \textbf{99.10} & \textbf{71.74} & 0.622 & 0.407 & \textbf{26.84} & 6.39 \\
\bottomrule
\end{tabular}}
\end{table}

\paragraph{Reward weights.}

The reward weights $(w_\text{geo}, w_\text{sem}, w_\text{tex})$ control how the agent aggregates the three loss modalities into a single scalar reward. The combined loss used as the RL training signal is:
\begin{equation}
\mathcal{L} = w_\text{geo} \cdot \mathcal{L}_\text{geo} + w_\text{sem} \cdot \mathcal{L}_\text{sem} + w_\text{tex} \cdot \mathcal{L}_\text{tex},
\label{eq:combined_loss_supp}
\end{equation}
where $\mathcal{L}_\text{geo}$, $\mathcal{L}_\text{sem}$, and $\mathcal{L}_\text{tex}$ denote the geometric, semantic, and texture reconstruction losses, respectively.

\Cref{tab:ablation_weights} shows that single-modality rewards (top three rows) perform noticeably worse than any balanced combination, confirming that all three signals carry complementary information about where subdivision is beneficial.
Among the balanced configurations (bottom four rows), performance differences are minimal: completion error varies by only 0.012\,cm and semantic accuracy by 0.15 percentage points.
This insensitivity is a desirable property; the weights serve primarily as an aggregation mechanism rather than a sensitive hyperparameter requiring careful tuning.
We adopt equal weights $(0.33, 0.33, 0.33)$ as the default for simplicity, noting that slightly emphasizing semantic loss $(0.25, 0.50, 0.25)$ yields virtually identical results.

\begin{table}[H]
\centering
\caption{Ablation on reward weight configuration $(w_\text{geo},\, w_\text{sem},\, w_\text{tex})$. $\dagger$ denotes the default configuration. Averaged over 10 ScanNet validation scenes.}
\label{tab:ablation_weights}
\setlength{\tabcolsep}{4pt}
\resizebox{\linewidth}{!}{%
\begin{tabular}{ccc ccc cc}
\toprule
$w_\text{geo}$ & $w_\text{sem}$ & $w_\text{tex}$ & Compl.\ Err↓ & Compl.\ $<$5cm↑ & Sem.\ Acc↑ & MS-SSIM↑ & LPIPS↓ \\
 & & & (cm) & (\%) & (\%) & & \\
\midrule
1.00 & 0.00 & 0.00 & 1.430 & 98.24 & 71.32 & 0.621 & 0.411 \\
0.00 & 1.00 & 0.00 & 1.513 & 97.83 & 71.31 & 0.620 & 0.413 \\
0.00 & 0.00 & 1.00 & 1.515 & 97.33 & 71.04 & 0.618 & 0.421 \\
\midrule
0.50 & 0.25 & 0.25 & 1.378 & 98.80 & 71.68 & 0.625 & 0.403 \\
0.25 & 0.25 & 0.50 & 1.376 & 98.74 & 71.57 & 0.625 & 0.404 \\
0.33 & 0.33 & 0.33$^\dagger$ & \textbf{1.366} & \textbf{98.96} & 71.71 & \textbf{0.626} & \textbf{0.403} \\
0.25 & 0.50 & 0.25 & \textbf{1.366} & \textbf{98.96} & \textbf{71.72} & \textbf{0.626} & \textbf{0.403} \\
\bottomrule
\end{tabular}}
\end{table}

\subsection{UN-MAP-ADAPT Parameter Sensitivity}
\label[subappendix]{sec:unmapadapt_sensitivity}

UN-MAP-ADAPT uses five hand-tuned thresholds that control when a voxel is subdivided based on its 
semantic entropy ($\tau^{\mathrm{sem}}$), texture complexity
($\tau^{\mathrm{tex}}_{\mathrm{lo}}$, $\tau^{\mathrm{tex}}_{\mathrm{hi}}$), and geometric complexity ($\tau^{\mathrm{geo}}_{\mathrm{lo}}$, $\tau^{\mathrm{geo}}_{\mathrm{hi}}$).
Unlike UnRL, which learns a subdivision strategy end-to-end, UN-MAP-ADAPT requires these thresholds to be specified manually and jointly determines the accuracy--memory operating point.

\Cref{tab:threshold_sensitivity} sweeps all five thresholds from loose (top rows, fewer subdivisions, low memory) to strict (bottom rows, more subdivisions, high memory).
Loose settings dramatically reduce memory but at significant accuracy cost: the loosest configuration degrades completion error to 2.22\,cm and semantic accuracy to 71.17\%.
As thresholds tighten, accuracy improves rapidly up to the default configuration (highlighted), after which returns diminish: below $\tau^{\mathrm{sem}}{=}0.10$, further tightening yields less than 0.03\,cm improvement in completion error while continuing to increase memory from 5.32\,GB toward 5.69\,GB.

This sweep illustrates the core motivation for our RL-based approach.
The optimal threshold configuration depends on the target memory budget, scene complexity, and downstream task requirements, and must be re-tuned whenever any of these factors change.
\ablUncertaintyGeoTextureRL replaces this manual search with a learned policy that automatically discovers the appropriate subdivision strategy for a given budget constraint, removing the need for per-deployment threshold engineering.

\begin{table}[H]
\centering
\caption{Threshold sensitivity analysis across configurations, averaged over 5 ScanNet validation scenes. The default configuration is highlighted in gray. Best values per metric are shown in \textbf{bold}.}
\label{tab:threshold_sensitivity}
\setlength{\tabcolsep}{4pt}
\resizebox{\linewidth}{!}{%
\begin{tabular}{ccccccccccc}
\toprule
$\tau^{\mathrm{sem}}$ &
$\tau^{\mathrm{tex}}_{\mathrm{lo}}$ &
$\tau^{\mathrm{tex}}_{\mathrm{hi}}$ &
$\tau^{\mathrm{geo}}_{\mathrm{lo}}$ &
$\tau^{\mathrm{geo}}_{\mathrm{hi}}$ &
\textbf{Compl. Err} $\downarrow$ &
\textbf{Compl. <5cm} $\uparrow$ &
\textbf{Sem. Acc} $\uparrow$ &
\textbf{MS-SSIM} $\uparrow$ &
\textbf{Voxels (M)} $\uparrow$ &
\textbf{Mem. (GB)} \\
\midrule
1.00 & 4.00 & 6.00 & 0.15 & 0.25 & 2.22 & 90.68 & 71.17 & 0.584 & 0.7 & 0.17 \\
0.50 & 2.00 & 4.00 & 0.10 & 0.20 & 1.66 & 96.16 & 72.86 & 0.604 & 6.5 & 1.54 \\
0.30 & 1.50 & 3.00 & 0.08 & 0.15 & 1.36 & 98.43 & 73.54 & 0.606 & 16.0 & 3.82 \\
0.20 & 1.25 & 2.50 & 0.06 & 0.12 & 1.24 & 99.07 & 74.02 & 0.612 & 20.0 & 4.78 \\
\rowcolor{gray!20}
0.10 & 1.00 & 2.00 & 0.05 & 0.10 & 1.20 & 99.37 & 74.55 & 0.617 & 22.3 & 5.32 \\
0.08 & 0.75 & 1.50 & 0.04 & 0.08 & 1.18 & 99.44 & 74.85 & \textbf{0.625} & 23.1 & 5.51 \\
0.06 & 0.50 & 1.00 & 0.04 & 0.07 & \textbf{1.17} & 99.53 & 74.86 & 0.624 & 23.6 & 5.62 \\
0.04 & 0.30 & 0.60 & 0.03 & 0.05 & \textbf{1.17} & 99.54 & 74.94 & \textbf{0.625} & 23.8 & 5.67 \\
0.02 & 0.15 & 0.30 & 0.02 & 0.04 & \textbf{1.17} & \textbf{99.55} & 74.98 & \textbf{0.626} & 23.8 & 5.68 \\
0.01 & 0.10 & 0.20 & 0.01 & 0.03 & \textbf{1.17} & \textbf{99.55} & \textbf{75.00} & \textbf{0.625} & \textbf{23.8} & 5.69 \\
\bottomrule
\end{tabular}%
}
\end{table}

\section{Training Hardware and Runtime}
\label[appendix]{sec:hardware}

All experiments, including both 3D semantic scene reconstruction and RL policy training, were conducted on a single workstation equipped with an Intel Core i7-14700K processor (20 cores, 28 threads) and an NVIDIA GeForce RTX 3080 GPU (10\,GB VRAM, CUDA 12.2).

For RL training, we used PPO with 4 parallel environments via \texttt{SubprocVecEnv}, training for 10M timesteps per experiment. Evaluation was performed every 50k steps on a held-out validation set of 5 episodes, with model checkpoints saved every 250k steps. Each training run took approximately 24 hours.

The 3D reconstruction pipeline, which generates the RL dataset by processing ScanNet scenes into per-frame voxel observations and reconstruction losses, was executed on the same machine.

\bibliography{main}

\end{document}